\documentclass[11pt]{article}

\usepackage{acl}

\usepackage{times}
\usepackage{latexsym}
\usepackage[T1]{fontenc}
\usepackage[utf8]{inputenc}
\usepackage{microtype}
\usepackage{inconsolata}
\usepackage{graphicx}
\usepackage{amsmath}
\usepackage{amssymb}
\usepackage{mathtools}
\usepackage{amsthm}
\usepackage{booktabs}
\usepackage{subcaption}
\usepackage{multirow}
\usepackage{array}
\usepackage{caption}
\usepackage{algorithm}
\usepackage{algorithmic}
\usepackage{xcolor}
\usepackage{tabularx}
\usepackage{caption}
\usepackage{bm}
\usepackage{placeins}

\theoremstyle{plain}
\newtheorem{theorem}{Theorem}
\newtheorem{proposition}[theorem]{Proposition}
\newtheorem{lemma}[theorem]{Lemma}
\theoremstyle{definition}

\title{ESSA: Evolutionary Strategies for Scalable Alignment}

\author{%
\begin{tabular}{@{}c@{\hspace{1.4em}}c@{\hspace{1.4em}}c@{\hspace{1.4em}}c@{}}
\textbf{Daria Korotyshova} &
\textbf{Boris Shaposhnikov} &
\textbf{Alexey Malakhov} &
\textbf{Alexey Khokhulin\textsuperscript{\normalfont\textdagger}} \\
\normalfont T-Tech &
\normalfont T-Tech &
\normalfont T-Tech &
\normalfont Yandex \\[1.2em]
\textbf{Nikita Surnachev} &
\textbf{Kirill Ovcharenko\textsuperscript{\normalfont\textdagger}} &
\textbf{George Bredis} &
\textbf{Alexey Gorbatovski} \\
\normalfont T-Tech &
\normalfont Yandex &
\normalfont T-Tech &
\normalfont T-Tech \\[1.2em]
& \textbf{Viacheslav Sinii} & \textbf{Daniil Gavrilov} & \\
& \normalfont T-Tech & \normalfont T-Tech &
\end{tabular}%
}

\begin{document}
\maketitle
\begingroup
\renewcommand{\thefootnote}{\textdagger}
\footnotetext{Work performed while at T-Tech.}
\endgroup

\begin{abstract}
Online alignment of large language models (LLMs) is dominated by reinforcement learning from human feedback (RLHF) with gradient-based optimizers such as PPO or GRPO. While effective, these pipelines require backpropagation through long rollouts, gradient synchronization across devices, and careful hyperparameter tuning, all of which become increasingly costly at scale. We present \textbf{ESSA} (Evolutionary Strategies for Scalable Alignment), a gradient-free online alignment stage that follows supervised fine-tuning (SFT) and replaces the gradient loop with inference-only black-box search. ESSA optimizes only the singular values of low-rank adaptation (LoRA) factors after a short SFT warm-start, restricting the search to a compact, task-aligned subspace where evolutionary search is practical even for 72B-parameter models. Because the loop is inference-only, ESSA is compatible with INT4/INT8 weight quantization and reduces inter-GPU communication to a few bytes per iteration. Across instruction following (IFEval), preference-based assistant tuning (HelpSteer2, HH-RLHF), and mathematical reasoning (GSM8K, MATH500), ESSA matches or exceeds LoRA-GRPO in the reported LoRA comparisons; on GSM8K it also outperforms Online DPO and PPO, while remaining competitive with both methods on IFEval. At scale, ESSA reaches a fixed MATH500 accuracy on Qwen2.5-32B/PRM800K up to 7.8$\times$ faster than LoRA-GRPO on 128 GPUs.
\end{abstract}

\section{Introduction}
\label{sec:intro}

Aligning large language models (LLMs) to follow instructions, respect safety constraints, and produce preferred answers is now a standard step in modern NLP pipelines~\citep{rlhf,llama,lambert2024tulu}. The dominant tool for online alignment is reinforcement learning from human feedback (RLHF), typically with PPO~\citep{ppo} or GRPO~\citep{grpo}. Full-parameter PPO/GRPO implementations train an actor (and often a critic) through backpropagation over long rollouts, require all-reduce of model-sized gradient tensors at every step, and rely on careful tuning of learning rate, KL penalty, clipping, and batch size. As model size grows, these mechanics become a dominant practical cost: communication, memory, and engineering overhead all scale with the parameter count rather than with task difficulty~\citep{moss1,verl}.

We revisit a much simpler primitive: evolutionary search with only forward passes. Evolutionary strategies (ES) are gradient-free, embarrassingly parallel, communicate only random seeds and scalar rewards, and place no requirements on the inference engine beyond the ability to sample from the model~\citep{es,cma-es}. Their classical limitation is the curse of dimensionality: applied naively to billions of parameters, ES is sample-inefficient and unstable. The question is whether a sufficiently compact, task-aligned parameterization can make ES a competitive alternative to the gradient loop in modern alignment pipelines after supervised fine-tuning.

\begin{figure*}[t]
    \centering
    \includegraphics[width=0.92\textwidth]{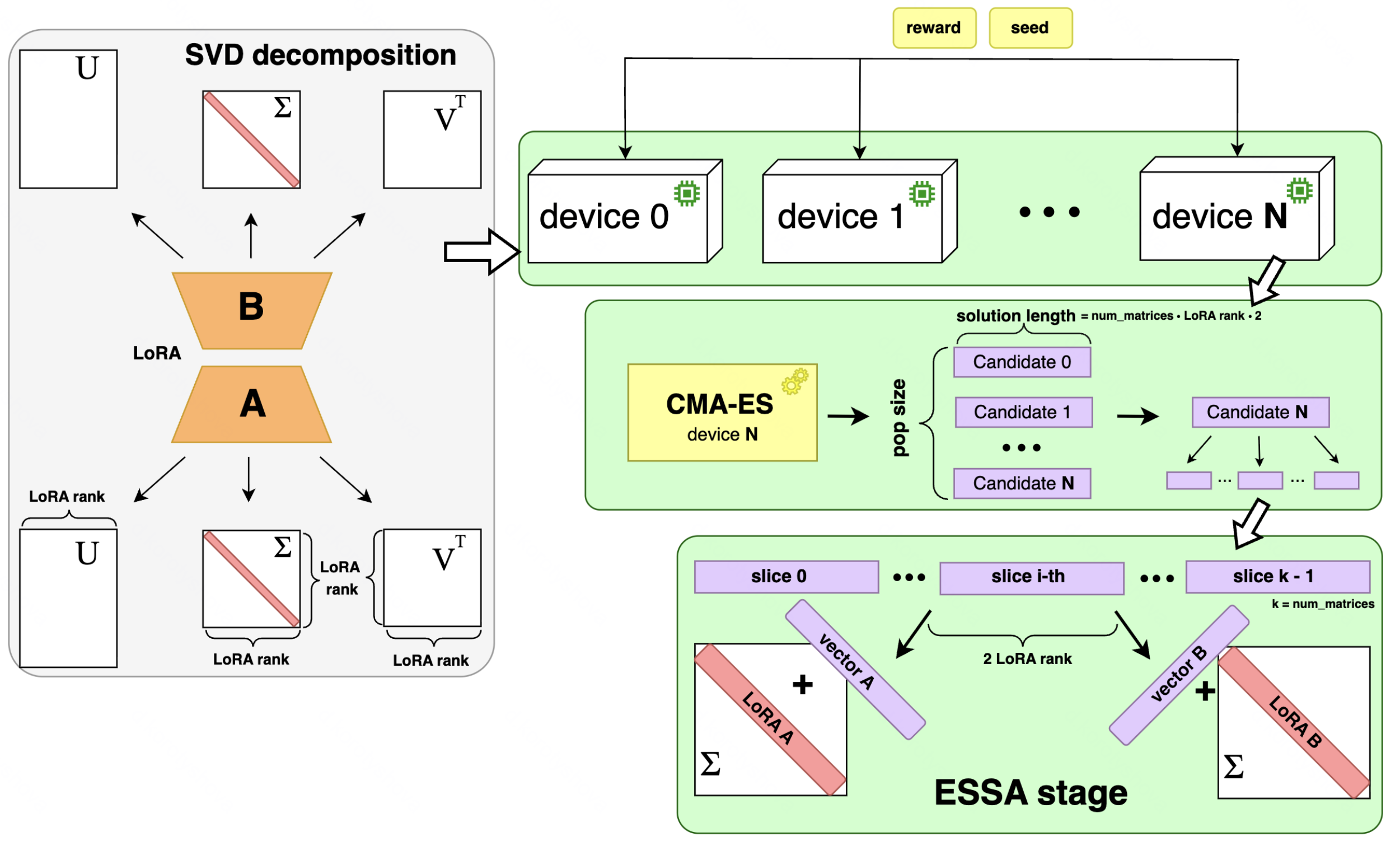}
    \caption{ESSA framework. After SFT, each LoRA factor $A,B$ is decomposed by SVD; orthogonal bases $U,V$ are frozen, and only the top-$\alpha$ fraction of the singular values is optimized by CMA-ES. Each device evaluates one candidate by reconstructing the LoRA factors and running an inference pass; workers exchange only random seeds and scalar rewards. \textbf{Slice~$n$} denotes the contiguous block of singular values corresponding to one LoRA matrix (e.g.\ $W_Q,W_K,W_V,W_O$ at a given transformer layer), and \textbf{device~$N$} denotes the GPU worker.}
    \label{fig:essa}
\end{figure*}

We answer this question affirmatively with \textbf{ESSA} (Evolutionary Strategies for Scalable Alignment). ESSA targets this post-SFT online alignment stage\footnote{Standard alignment pipelines, such as InstructGPT~\citep{NEURIPS2022_b1efde53}, Tulu~3~\citep{lambert2024tulu}, and Llama~3~\citep{llama}, apply an SFT warm-start before the RL stage, and we follow this convention. ESSA replaces only the RL stage; the SFT cost is shared with all baselines and reported separately.} and combines three ingredients: (i) restricting updates to low-rank adaptation (LoRA) modules in the attention matrices~\citep{lora}; (ii) decomposing each LoRA factor via singular value decomposition (SVD) and freezing the singular vectors, leaving only $\mathcal{O}(r)$ singular values per factor as trainable parameters~\citep{transformer2}; and (iii) optimizing this compact vector with CMA-ES~\citep{cma-es}, evaluating each population member with a single inference pass. The loop has three consequences (Figure~\ref{fig:essa}):

\begin{itemize}
  \setlength\itemsep{1pt}
\item \textbf{Inference-only.} Because no backward pass is performed, ESSA can run natively with INT8/INT4-quantized weights, enabling one-candidate-per-GPU evaluation for models with up to 72B parameters on a single GPU.
\item \textbf{Communication-minimal.} Workers exchange a 32-bit seed and a 32-bit scalar reward per evaluation, eliminating the all-reduce of model-sized tensors that bottlenecks full-parameter gradient RLHF.
\item \textbf{Hyperparameter-light.} ESSA has three principal swept algorithmic knobs (LoRA rank, population size, and the fraction $\alpha$ of singular values optimized) and we observe broad plateaus of stable accuracy for all three (Section~\ref{sec:sensitivity}).
\end{itemize}

\noindent\textbf{Key insight:} In our controlled comparison, population-based gradient-free optimization substantially outperforms SVD-GRPO in the same SFT-seeded singular-value space (Section~\ref{sec:optimizer_controls}).

\noindent\textbf{Contributions.} (1) We introduce ESSA, a gradient-free method for post-SFT online alignment that applies CMA-ES to the singular values of SVD-decomposed LoRA factors, targeting parameter-efficient settings with inexpensive reward evaluation. (2) We give a fair comparison against Online DPO, PPO, LoRA-GRPO, full-parameter GRPO, and SVD-GRPO across instruction following (IFEval), preference-based assistant tuning (HelpSteer2, Anthropic HH-RLHF) and verifiable-reward math (GSM8K, MATH500); ESSA matches or exceeds LoRA-GRPO in the reported LoRA comparisons, outperforms all LoRA baselines on GSM8K, and remains competitive with PPO and Online DPO on IFEval. ESSA also matches full-parameter GRPO on math while training $\sim$10\textsuperscript{6}$\times$ fewer parameters. (3) We characterize ESSA's scaling from 16 to 128 GPUs using wall-clock measurements and a theoretical iteration-time bound, and show that INT4 quantization reduces accuracy by approximately one percentage point relative to BF16.

\section{Related Work}
\label{sec:related}

\paragraph{Online vs.\ offline alignment.}
Online RLHF~\citep{NEURIPS2022_b1efde53} with PPO~\citep{ppo}, REINFORCE~\citep{reinforce}, RLOO~\citep{rloo}, GRPO~\citep{grpo}, REINFORCE++~\citep{reinforce++}, or DAPO~\citep{dapo} remains the standard recipe for verifiable-reward and preference-based tasks. Offline preference learning (DPO~\citep{dpo}, ORPO~\citep{orpo}, SimPO~\citep{simpo}) removes online rollouts but is bounded by the static preference dataset and underperforms online methods~\citep{gap,dpovsppo}. Accordingly, we compare ESSA against LoRA-based Online DPO, PPO, and LoRA-GRPO, which share its online setting; offline methods represent an orthogonal design choice.

\paragraph{Parameter-efficient and singular-value adaptation.}
LoRA~\citep{lora}, adapters~\citep{adapters}, and prefix-tuning~\citep{prefix_tuning} all reduce the number of trainable parameters. A separate line of work, including Transformer\textsuperscript{2}~\citep{transformer2}, SVFT~\citep{svft}, SVFit~\citep{svfit}, and Singular Value Fine-Tuning~\citep{svf2022}, freezes singular vectors and optimizes only singular values using gradient-based objectives in supervised fine-tuning settings. ESSA shares the singular-value parameterization but differs in two crucial ways. (i) \emph{Online alignment setting:} prior singular-value methods are evaluated for supervised fine-tuning rather than post-SFT online policy optimization from scalar rewards. (ii) \emph{Optimization regime:} our controlled comparisons using the same SFT initialization and singular-value coordinates show that population-based gradient-free optimization is substantially more effective than SVD-GRPO in this space (Section~\ref{sec:optimizer_controls}). Among the evaluated methods, CMA-ES offers the strongest practical trade-off between convergence speed and tuning requirements, reaching strong performance rapidly without task-specific step-size tuning~\citep{cma-es}. ESSA's contribution is the demonstration that this parameterization supports practical gradient-free online alignment at LLM scale.

\paragraph{Evolutionary strategies for LLMs.}
Classical ES variants (CMA-ES, NES~\citep{nes}, ARS~\citep{ars}, GES~\citep{ges}) offer gradient-free, highly parallel optimization with linear scaling in the number of evaluators. Zero-order methods~\citep{zo-opt} approximate gradients from loss differences but underperform on complex reasoning. Recent ES-for-LLM work (GENOME/GENOME+~\citep{genome}, LoRAHub~\citep{lorahub}, DFO~\citep{dfo}, ES-FT~\citep{esft}, and EGGROLL~\citep{eggroll}) either targets hyperscale full-parameter search or restricts ES to model merging (GENOME/GENOME+, LoRAHub), i.e.\ searching only for coefficients that combine pretrained checkpoints rather than learning new behaviour. Among ES-based methods for LLM fine-tuning (Appendix Table~\ref{tab:es_comparison}), ESSA uses the lowest-dimensional search space among these methods, which
makes population-based search practical at LLM scale~\citep{es}, and is the first ES-based method we are aware of to be applied to LLMs at the 72B parameter scale on a single 80\,GB GPU. We also outperform EGGROLL and match ES-FT on Countdown while using a much smaller search space (Appendix~\ref{app:other}).

\section{Method}
\label{sec:method}

\paragraph{Setting and scope.}
We consider post-SFT online alignment, where the model has already learned the task's input/output format and is further optimized using a scalar reward provided by either a rule-based verifier or a reward model. ESSA is neither a replacement for SFT nor a method for pretraining from scratch: its singular-value updates assume that the SVD basis of the SFT-trained LoRA factors already contains task-relevant directions. We validate this assumption using a controlled MNIST analogue of the full pipeline (Appendix Figure~\ref{fig:essa_mnist}). Our pipeline therefore consists of two stages: (1) standard SFT on a small subset of the data, followed by (2) ESSA optimization on a disjoint online split.

\paragraph{LoRA + SVD parameterization.}
For each attention matrix $W_0\!\in\!\mathbb{R}^{m\times n}$ we introduce a LoRA update $\Delta W\!=\!BA$ with $B\!\in\!\mathbb{R}^{m\times r}$, $A\!\in\!\mathbb{R}^{r\times n}$, $r\!\ll\!\min(m,n)$~\citep{lora}. After SFT, we SVD-decompose each factor:
\[
A = U_A \Sigma_A V_A^{\top}, \qquad B = U_B \Sigma_B V_B^{\top}.
\]
The orthogonal bases $U_A,V_A,U_B,V_B$ are frozen; only the top-$\alpha$ fraction of the singular values in $\Sigma_A,\Sigma_B$ is exposed to the optimizer. This shrinks the trainable space per factor from $\mathcal{O}(r(m{+}n))$ to $\mathcal{O}(r)$.

\paragraph{Evolutionary optimization.}
Let $\theta \in \mathbb{R}^d$ concatenate all trainable singular-value offsets ($d \approx 1{,}800$ for Qwen2.5-7B with $r=8$ and four attention matrices per layer). At iteration $t$, CMA-ES~\citep{cma-es} samples a population $\{x_t^{(k)}\}_{k=1}^{\lambda}$ from $\mathcal{N}(m_t,\sigma_t^2 C_t)$, where $m_t$ is the distribution mean, $\sigma_t$ is the global step size, and $C_t$ is the covariance matrix. It evaluates the scalar reward $F(x_t^{(k)})$ for each candidate and updates $(m_{t+1},\sigma_{t+1},C_{t+1})$ based on the reward rankings. Crucially, the workers share a single random-number stream, allowing each worker to locally regenerate any population member from a seed; communication is therefore reduced to $\langle\text{seed},\text{reward}\rangle$ pairs. The full procedure is given in Algorithm~\ref{alg:essa} (Appendix~\ref{app:algorithm}).

\paragraph{Iteration-time bound.}
Let $G$ be the number of GPUs, $M_{\text{params}}$ the model size in bytes, and $\eta_{\text{gen}},\eta_{\text{fb}}$ the per-sample generation and forward-backward times. Under mild assumptions on tensor-parallel sizes (Appendix~\ref{app:theory}), the ESSA iteration is strictly faster than the idealized full-parameter gradient pipeline whenever
\begin{equation}
\label{eq:essa_bound}
\begin{aligned}
N_{\text{pop}}B^{\text{essa}}
&< B^{\text{grad}}\!\left(1 +
\tfrac{\eta_{\text{fb}}}{G\,\eta_{\text{gen}}}\right) \\
&\quad + \tfrac{(G{-}1)M_{\text{params}}}
{\mathrm{peak\_bw}\,G\,\eta_{\text{gen}}}.
\end{aligned}
\end{equation}
Equation~\ref{eq:essa_bound} models the full-parameter pipeline and does not describe the LoRA-GRPO baseline evaluated in Figure~\ref{fig:scaling}. The right-hand side grows linearly in $M_{\text{params}}$, so the ESSA-feasible population scales with model size (Section~\ref{sec:scaling}).

\section{Experiments}
\label{sec:exp}

\paragraph{Tasks and backbones.}
We evaluate on four task families that span the modern alignment spectrum: \emph{(i) Instruction following:} train on an IFEval-like dataset~\citep{ifevallike}, evaluate on IFEval~\citep{ifeval}; backbone Llama-3.1-8B~\citep{llama}. \emph{(ii) Preference-based assistant tuning:} train and evaluate on HelpSteer2~\citep{hs2} with the ArmoRM reward model~\citep{armorm}, backbone Llama-3.1-8B; safety/helpfulness on Anthropic HH-RLHF~\citep{anthropic_hhrlhf} with the OpenAssistant DeBERTa-v3 reward model, backbone Mistral-7B~\citep{mistral7b}. \emph{(iii) Mathematical reasoning:} GSM8K~\citep{gsm8k} and PRM800K$\to$MATH500~\citep{math500} plus AIME'24/'25~\citep{aime24,aime25}, MinervaMath~\citep{minervamath}, OlympiadBench~\citep{olympiadbench}, AMC'23~\citep{amc_aime}; backbones Qwen2.5-7B/-Math-7B/-32B/-72B~\citep{qwen25,qwen25math}. \emph{(iv) Symbolic puzzle:} Countdown (Appendix~\ref{app:other}).

\paragraph{Baselines.}
Throughout the paper, we use \emph{LoRA-GRPO} to denote GRPO
updates restricted to LoRA parameters; full-parameter GRPO and
SVD-GRPO are named explicitly.
We compare against four online LoRA baselines: Online DPO, PPO,
LoRA-GRPO, and SVD-GRPO (GRPO restricted to singular-value updates),
as well as full-parameter GRPO. For the gradient-baseline
experiments reported in the main text, we train with the verl
library~\citep{verl} and select the learning rate from
$\{10^{-6}, 5\!\times\!10^{-6}, 10^{-5},
5\!\times\!10^{-5}, 10^{-4}, 5\!\times\!10^{-4},
10^{-3}, 5\!\times\!10^{-3}, 10^{-2}\}$.
Run-level LoRA-GRPO and SVD-GRPO sweep results are reported in
Appendix Tables~\ref{tab:grpo-run-level-gsm-ranks},
\ref{tab:grpo-run-level-ifeval}, and
\ref{tab:grpo-run-level-prm}. SFT checkpoints are shared
between ESSA and all baselines; the SFT cost is reported
separately (Appendix~\ref{app:implementation}) and contributes
an identical constant to every method. Unless noted, all
methods run in BFLOAT16; the 72B experiments additionally
report ESSA under INT4 (see Section~\ref{sec:sensitivity} for
the precision study).

\subsection{LoRA-budget alignment}
\label{sec:main_comparison}

Table~\ref{tab:main_compare} reports test accuracy at the
checkpoint from the final validation evaluation on GSM8K (verifiable rewards,
Qwen2.5-7B) and IFEval (rule-based rewards,
Llama-3.1-8B). The corresponding training configurations and
policy-generation and reward-query counts are reported in Appendix
Table~\ref{tab:main-reported-configs}. Among the
parameter-efficient methods, ESSA achieves the highest accuracy
on GSM8K while optimizing only approximately 2K--4K
singular-value coordinates and requiring no backpropagation. On
GSM8K, ESSA improves over the strongest LoRA baseline, LoRA-GRPO, by
4.4 percentage points (0.867 vs.\ 0.823). On IFEval, ESSA
outperforms LoRA-GRPO by 7.2 percentage points (0.555 vs.\ 0.483),
and is within 0.4 percentage points of Online DPO and PPO (both 0.559).

Full-parameter GRPO, which updates billions of model weights
rather than the approximately 2K--4K coordinates optimized by
ESSA, reaches higher accuracy on IFEval (0.692 vs.\ 0.555),
whereas the two methods achieve comparable accuracy on GSM8K
(0.870 vs.\ 0.867).

\begin{figure}[t]
    \centering
    \includegraphics[width=0.95\linewidth]
        {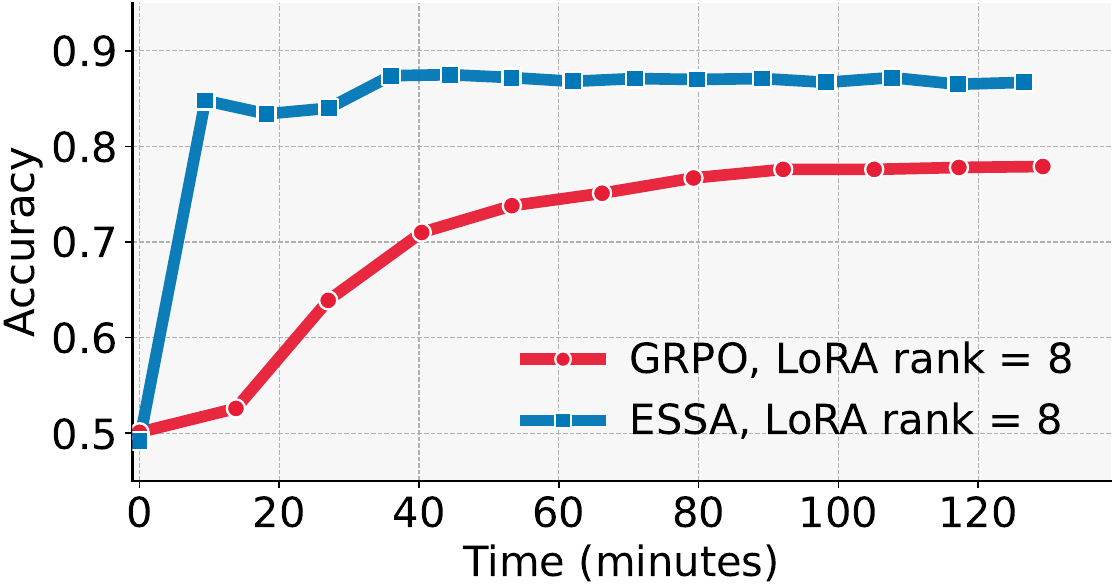}
    \caption{
        Validation accuracy on \textbf{GSM8K} with
        \textbf{Qwen2.5-7B} in a separate LoRA-rank-8 comparison.
        ESSA (blue): batch size 100; final test accuracy over five
        seeds $0.853\pm0.008$. LoRA-GRPO (red): learning rate
        $1\!\times\!10^{-4}$, global batch size 512, minibatch size
        64; final test accuracy over five seeds $0.847\pm0.014$.
        ESSA reaches its plateau rapidly, while LoRA-GRPO improves more
        gradually. Full configurations and policy-generation and
        reward-query counts are reported in Appendix
        Table~\ref{tab:main-reported-configs}; complete run-level
        LoRA-GRPO learning-rate sweep results are reported in Appendix
        Table~\ref{tab:grpo-run-level-gsm-ranks}.
    }
    \label{fig:gsm8k}
\end{figure}

Figure~\ref{fig:gsm8k} provides a separate comparison for ESSA and LoRA-GRPO on GSM8K using
Qwen2.5-7B at LoRA rank 8. Additional ESSA--LoRA-GRPO comparisons
across backbones, datasets, and LoRA ranks are reported in
Appendix~\ref{app:full_grpo}.

\begin{table}[t]
    \centering
    \small
    \setlength{\tabcolsep}{2pt}

    \begin{tabular}{@{}lrcc@{}}
        \toprule
        \textbf{Method}
        & \begin{tabular}[c]{@{}c@{}}
            \textbf{Optimized} \\
            \textbf{dim.}
          \end{tabular}
        & \textbf{GSM8K}
        & \textbf{IFEval} \\
        \midrule

        SFT only
            & ---
            & 0.723
            & 0.451 \\

        Online DPO
            & 13/17 M
            & $0.818\pm0.001$
            & $0.559\pm0.008$ \\

        PPO
            & 26/34 M
            & $0.791\pm0.010$
            & \underline{$0.559\pm0.007$} \\

        LoRA-GRPO
            & 13/17 M
            & $0.823\pm0.012$
            & $0.483\pm0.002$ \\

        \midrule

        \begin{tabular}[t]{@{}l@{}}
            Full-parameter\\
            GRPO (no SFT)
        \end{tabular}
            & 7/8 B
            & $\mathbf{0.870\pm0.009}$
            & $\mathbf{0.692\pm0.008}$ \\

        \midrule

        \textbf{ESSA (ours)}
            & 2/4 K
            & \underline{$0.867\pm0.010$}
            & $0.555\pm0.009$ \\

        \bottomrule
    \end{tabular}

    \caption{
        Test accuracy at the checkpoint from the final validation evaluation on
        \textbf{GSM8K} (\textbf{Qwen2.5-7B}) at LoRA rank 16 and \textbf{IFEval}
        (\textbf{Llama-3.1-8B}) at LoRA rank 8. Results for the online
        alignment methods are reported as mean $\pm$ standard
        deviation over five independent seeds; SFT only denotes
        the shared initialization checkpoint. Optimized
        dim.\ denotes the number of optimized scalar parameters for
        \textbf{Qwen2.5-7B} on \textbf{GSM8K} / \textbf{Llama-3.1-8B} on \textbf{IFEval}. For PPO,
        the reported dimension includes the LoRA parameters optimized
        in both the policy and value models. For these main
        comparisons, the gradient-based LoRA baselines use the best
        learning rate selected from a sweep of nine values. Complete
        run-level LoRA-GRPO learning-rate sweep results are reported in
        Appendix Tables~\ref{tab:grpo-run-level-gsm-ranks}
        and~\ref{tab:grpo-run-level-ifeval}. Full-parameter GRPO is
        trained without SFT and is included as an unconstrained
        reference. Training configurations and policy-generation and
        reward-query counts are reported in Appendix
        Table~\ref{tab:main-reported-configs}. Underlining denotes the
        best parameter-efficient result for each task.
    }
    \label{tab:main_compare}
\end{table}

\subsection{Optimizer and subspace controls}
\label{sec:optimizer_controls}

To identify which components of ESSA drive its performance, we
compare alternative optimization methods in the same
singular-value space. All methods start from the same SFT
checkpoint.

\begin{figure}[t]
    \centering
    \includegraphics[width=\columnwidth]
    {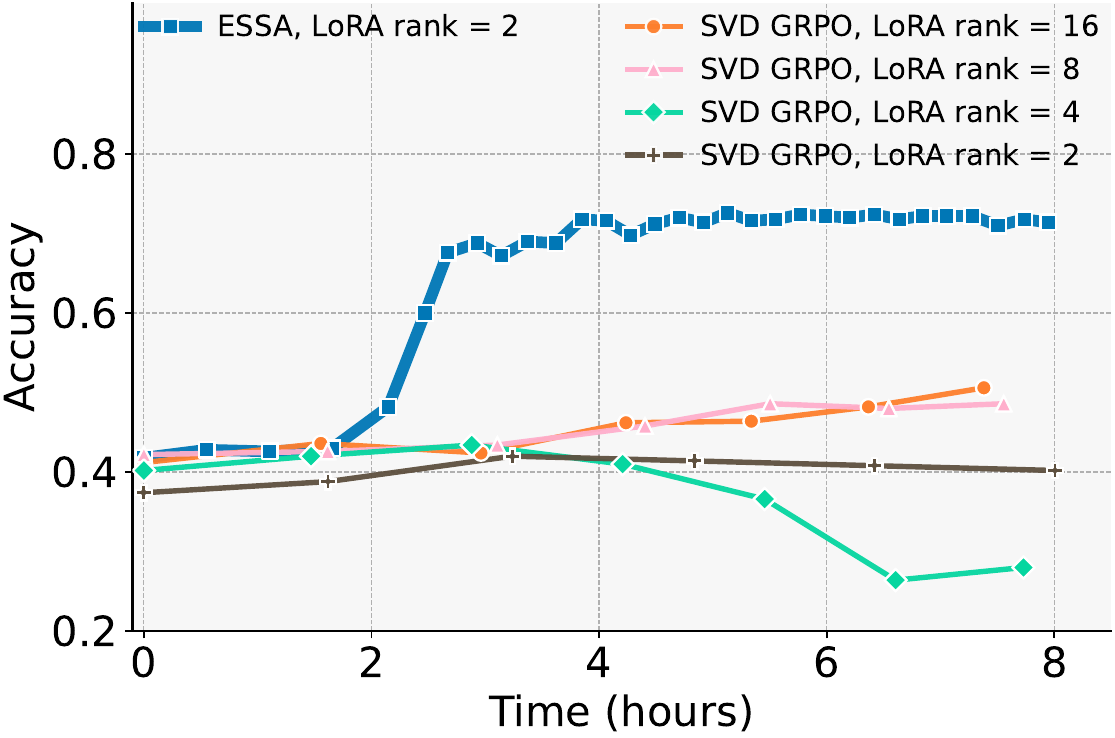}
    \caption{
        Validation accuracy over time on \textbf{PRM800K} with
        \textbf{Qwen2.5-7B}. ESSA (blue) uses LoRA rank 2,
        population size 24, batch size 300, and
        $\alpha=1.0$. SVD-GRPO (singular-value updates only)
        uses LoRA ranks 2/4/8/16, learning rate
        $1\!\times\!10^{-2}$, global batch size 512, and
        minibatch size 64. At the checkpoints from the final
        validation evaluation, ESSA achieves a MATH500 test accuracy of
        $0.723\pm0.005$ even with rank 2, whereas SVD-GRPO
        achieves $0.422\pm0.008$, $0.322\pm0.012$,
        $0.488\pm0.001$, and $0.510\pm0.002$ for ranks 2, 4,
        8, and 16, respectively. Configurations and generation
        counts are reported in Appendix
        Table~\ref{tab:main-reported-configs}; complete
        SVD-GRPO learning-rate sweeps are reported in Appendix
        Table~\ref{tab:grpo-run-level-prm}.
    }
    \label{fig:svd_grpo_ablation}
\end{figure}

\paragraph{SVD-GRPO ablation.}
We isolate the optimization regime from the parameterization by
applying SVD-GRPO to the same SFT-initialized SVD-LoRA coordinates
used by ESSA. For each LoRA rank $r\in\{2,4,8,16\}$, we sweep
nine learning rates and select the checkpoint using validation
accuracy. At the checkpoint from the final validation evaluation, SVD-GRPO
achieves MATH500 test accuracies of $0.422\pm0.008$,
$0.322\pm0.012$, $0.488\pm0.001$, and $0.510\pm0.002$ for
ranks 2, 4, 8, and 16, respectively. In contrast, ESSA achieves
$0.723\pm0.005$ even with LoRA rank 2
(Figure~\ref{fig:svd_grpo_ablation}). Thus, the parameterization
alone does not explain ESSA's performance. Configurations and
generation counts are reported in Appendix
Table~\ref{tab:main-reported-configs}; complete SVD-GRPO
learning-rate sweeps are reported in Appendix
Table~\ref{tab:grpo-run-level-prm}.

\paragraph{Alternative gradient-free optimizers.}
We next compare CMA-ES with plain isotropic
$(\mu,\lambda)$-ES, ARS, random search, and MeZO. The
methods use the same SFT-seeded search space and a per-step
budget of 96 candidate evaluations. CMA-ES uses the same fixed
initial global step size in all experiments; step sizes for the
alternative optimizers are tuned separately. Results are
reported in Table~\ref{tab:optimizer_controls}, with convergence
details in Appendix Table~\ref{tab:optimizer-convergence}. Full
configurations and generation counts are reported in Appendix
Table~\ref{tab:other-es-configs}.

\begin{table}[t]
\centering
\small
\setlength{\tabcolsep}{3pt}
\resizebox{\columnwidth}{!}{%
\begin{tabular}{lcc}
\toprule
\textbf{Optimizer}
& \textbf{GSM8K}
& \textbf{MATH500} \\
\midrule
CMA-ES (ESSA)
& $0.867\pm0.010$ & $0.706\pm0.009$ \\
Plain isotropic $(\mu,\lambda)$-ES
& $\mathbf{0.893\pm0.006}$ & $0.736\pm0.008$ \\
ARS
& $0.884\pm0.004$ & $\mathbf{0.754\pm0.006}$ \\
Random search
& $0.842\pm0.005$ & $0.636\pm0.004$ \\
MeZO (SPSA)
& $0.826\pm0.008$ & $0.472\pm0.010$ \\
\bottomrule
\end{tabular}%
}
\caption{
Test accuracy at the checkpoint from the final validation evaluation on \textbf{GSM8K}
and \textbf{MATH500}, where the \textbf{MATH500} models are trained on \textbf{PRM800K}.
Both experiments use \textbf{Qwen2.5-7B}, LoRA rank 16, population size
96, with per-candidate batch size 100 on \textbf{GSM8K} and 300 on
\textbf{PRM800K}$\rightarrow$\textbf{MATH500}. All methods operate on the same
SFT-seeded singular-value parameterization. Results are reported
as mean $\pm$ standard deviation over five independent seeds.
Full configurations and generation counts are reported in
Table~\ref{tab:other-es-configs}; optimizer convergence
is reported in Appendix Table~\ref{tab:optimizer-convergence}.
}
\label{tab:optimizer_controls}
\end{table}

Plain isotropic ES achieves the highest test accuracy at the
checkpoint from the final validation evaluation on GSM8K, while ARS performs best
on MATH500. We retain CMA-ES because it makes stronger
early progress: on MATH500 it reaches 0.732 validation accuracy
after approximately 2K candidate evaluations, earlier than
plain ES and ARS (Appendix
Table~\ref{tab:optimizer-convergence}).

The SFT-derived basis provides an additional but smaller benefit.
Under a matched 37-step GSM8K budget, it reaches 0.838 test
accuracy, compared with 0.823 for a random orthogonal basis.
ESSA is also stable across SFT-data fractions from 5\% to 100\%,
with test accuracy between 0.699 and 0.728. These controls are
reported in Appendix~\ref{app:optimizer-controls}.

\subsection{Parallel scaling}
\label{sec:scaling}

\begin{figure}[t]
    \centering
    \includegraphics[width=0.95\linewidth]{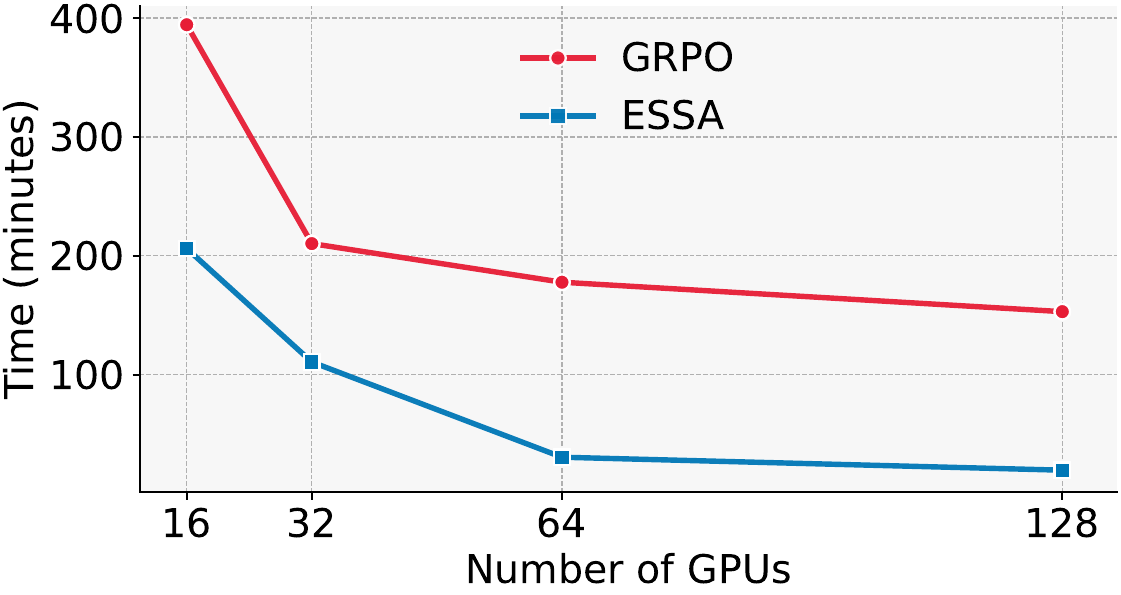}
    \caption{
        LoRA-GRPO and ESSA scaling on \textbf{PRM800K} with
        \textbf{Qwen2.5-32B}: wall-clock time to reach 0.835
        MATH500 test accuracy as a function of GPU count. ESSA
        (red) uses LoRA rank 16, population size 128, batch
        size 256, and $\alpha=1.0$. LoRA-GRPO (blue) uses
        LoRA rank 16, learning rate $1\times10^{-5}$, global
        prompt batch 512, actor mini-batch size 64, and 8
        responses per prompt. Per-GPU-count wall-clock times and
        policy-generation/reward-query counts are also listed
        in Appendix Table~\ref{tab:main-reported-configs}.
    }
    \label{fig:scaling}
\end{figure}

Figure~\ref{fig:scaling} compares the wall-clock time required
by LoRA-GRPO and ESSA to reach a fixed MATH500 accuracy of
0.835 on Qwen2.5-32B/PRM800K as the number of GPUs increases
from 16 to 128. Across this range, ESSA reduces its time to the
target by approximately an order of magnitude, whereas
LoRA-GRPO improves by roughly $2.6\times$. Consequently,
ESSA's relative wall-clock advantage increases with the number
of workers and is largest in the 128-GPU configuration, where
ESSA reaches the target approximately $7.8\times$ faster.
Per-GPU-count wall-clock times and the corresponding policy-generation
and reward-query counts are reported in Appendix
Table~\ref{tab:main-reported-configs}.

This speedup is not a claim that ESSA always uses fewer policy
generations. Time to accuracy depends jointly on the number of
generated responses, generation throughput, and work performed
outside policy generation.

\paragraph{Measured step-time breakdown.}
At the largest evaluated scale, we profile both methods on 128
NVIDIA H100-equivalent GPUs using the configurations from
Figure~\ref{fig:scaling}, the same SFT checkpoint, and BF16
precision. For both ESSA and LoRA-GRPO, we first average each
component time across all 128 ranks within each step. We then
compute the reported mean and standard deviation across 20
pre-specified steady-state training steps.

\begin{table*}[t]
    \centering
    \small
    \renewcommand{\arraystretch}{1.08}
    \begin{tabular*}{\textwidth}{
        @{\extracolsep{\fill}}
        l
        rr
        rr
        @{}
    }
        \toprule
        & \multicolumn{2}{c}{\textbf{ESSA}}
        & \multicolumn{2}{c}{\textbf{LoRA-GRPO}} \\
        \cmidrule(lr){2-3}
        \cmidrule(lr){4-5}

        \textbf{Step component}
        & \textbf{s/step}
        & \textbf{\%}
        & \textbf{s/step}
        & \textbf{\%} \\
        \midrule

        Candidate reconstruction
            & $5.38\pm0.12$
            & $2.28\pm0.05$
            & -- & -- \\

        Adapter dispatch and loading
            & $0.18\pm0.01$
            & $0.08\pm0.004$
            & -- & -- \\

        Policy rollout generation
            & $189.34\pm25.67$
            & $80.17\pm10.87$
            & $187.82\pm23.75$
            & $51.91\pm3.02$ \\

        Reward evaluation
            & $0.71\pm0.09$
            & $0.30\pm0.04$
            & $1.71\pm0.27$
            & $0.47\pm0.04$ \\

        Old-policy and reference forwards
            & -- & --
            & $46.15\pm0.10$
            & $12.81\pm0.91$ \\

        Actor forward
            & -- & --
            & $23.38\pm0.03$
            & $6.49\pm0.44$ \\

        Backward compute
            & -- & --
            & $36.92\pm0.06$
            & $10.25\pm0.72$ \\

        Exposed gradient all-reduce
            & -- & --
            & $30.54\pm0.38$
            & $8.47\pm0.47$ \\

        Optimizer step
            & -- & --
            & $0.339\pm0.002$
            & $0.094\pm0.007$ \\

        Policy-to-rollout weight sync
            & -- & --
            & $0.236\pm0.045$
            & $0.066\pm0.017$ \\

        Search-distribution update
            & $7.83\pm0.21$
            & $3.32\pm0.09$
            & -- & -- \\

        ESSA residual orchestration and idle
            & $32.74\pm0.78$
            & $13.86\pm0.33$
            & -- & -- \\

        LoRA-GRPO barriers, idle, and stragglers
            & -- & --
            & $34.03\pm0.51$
            & $9.44\pm0.51$ \\

        \midrule

        \textbf{Total step}
            & $\mathbf{236.18\pm25.68}$
            & \textbf{100}
            & $\mathbf{361.11\pm24.72}$
            & \textbf{100} \\

        \bottomrule
    \end{tabular*}

    \caption{
        Step-time breakdown on 128 NVIDIA H100-equivalent GPUs for \textbf{Qwen2.5-32B} on \textbf{PRM800K}.
        ESSA uses LoRA rank 16, population size 128, and 256
        prompts per candidate. LoRA-GRPO uses LoRA rank 16,
        a global prompt batch of 512, and eight responses per
        prompt. For each step and component, times are first
        averaged across all 128 ranks. The displayed mean and
        sample standard deviation are then computed across the
        same 20 pre-specified steady-state training steps. The training configurations are
        listed in Appendix
        Table~\ref{tab:main-reported-configs}.
    }
    \label{tab:step_profile}
\end{table*}

\paragraph{Why ESSA's steps are faster.}
The mean rollout-generation times are nearly identical for ESSA
($189.34\pm25.67$\,s) and LoRA-GRPO
($187.82\pm23.75$\,s), although an ESSA step generates 32{,}768
responses and a LoRA-GRPO step generates 4{,}096. ESSA achieves
this eightfold response throughput because its 128 population
members are independent and can be distributed across the
cluster. LoRA-GRPO instead divides a fixed global batch of 512
prompts among the same workers, leaving little per-device work
at 128 GPUs.

LoRA-GRPO is slower because each step also serializes old-policy and
reference-model forwards ($12.81\pm0.91\%$), the actor update
(25.3\%), and barriers, idle time, and stragglers
($9.44\pm0.51\%$). ESSA performs none of the reference-model,
backward, gradient-synchronization, optimizer, or policy-weight
resynchronization stages; candidate reconstruction, adapter
loading, and the search-distribution update together occupy less
than 6\% of its mean step time. Consequently, the mean complete
ESSA step takes $236.18\pm25.68$\,s, compared with
$361.11\pm24.72$\,s for LoRA-GRPO, corresponding to a
$1.53\times$ speedup based on the mean step times.

Gradient all-reduce accounts for $8.47\pm0.47\%$ of the complete
LoRA-GRPO step and is therefore not its sole bottleneck. The
measured slowdown is explained by the combination of an
underfilled fixed global batch, additional policy and actor
computations, and synchronization and idle time. The larger
$7.8\times$ time-to-accuracy advantage in
Figure~\ref{fig:scaling} combines this lower mean per-step cost
with the number of optimization steps required to reach the
target.

\paragraph{Generation and reward-query accounting.}
Appendix Table~\ref{tab:main-reported-configs} reports training
policy-generation and reward-query counts ($G=Q$) for the main
experiments. The counts vary in both directions: in the rank-8
GSM8K comparison, ESSA uses 201{,}600 generations versus 204{,}800
for LoRA-GRPO, whereas the main GSM8K comparison uses 489{,}600 versus
184{,}320 and IFEval uses 1{,}764{,}000 versus 122{,}880. These
counts correspond to the displayed trajectories rather than a
common accuracy threshold, so they do not constitute a uniform
sample-efficiency comparison.

Each generated response is scored once, making reward-query and
policy-generation counts equal. ESSA's measured speed therefore
comes from parallel generation and the absence of LoRA-GRPO's
additional forward, backward, optimizer, and synchronization
stages, not from consistently using fewer samples. With expensive
learned reward models, the additional reward evaluations may
reduce or eliminate this advantage.

\paragraph{Batch-size and theoretical scope.}
We keep the LoRA-GRPO global prompt batch fixed at 512 and do not
sweep batch size at 128 GPUs. Figure~\ref{fig:scaling} therefore
characterizes this measured configuration, not the best possible
LoRA-GRPO scaling over all batch sizes. Equation~\eqref{eq:essa_bound}
describes an idealized full-parameter gradient pipeline rather
than this LoRA-GRPO baseline. Accordingly, the scaling claim rests
on the wall-clock measurements in Figure~\ref{fig:scaling}, and
the explanation of the 128-GPU result rests on the measured
decomposition in Table~\ref{tab:step_profile}.

\subsection{Instruction following, preference, and safety}
\label{sec:nlp}

\begin{figure}[t]
    \centering
    \includegraphics[width=0.95\linewidth]
        {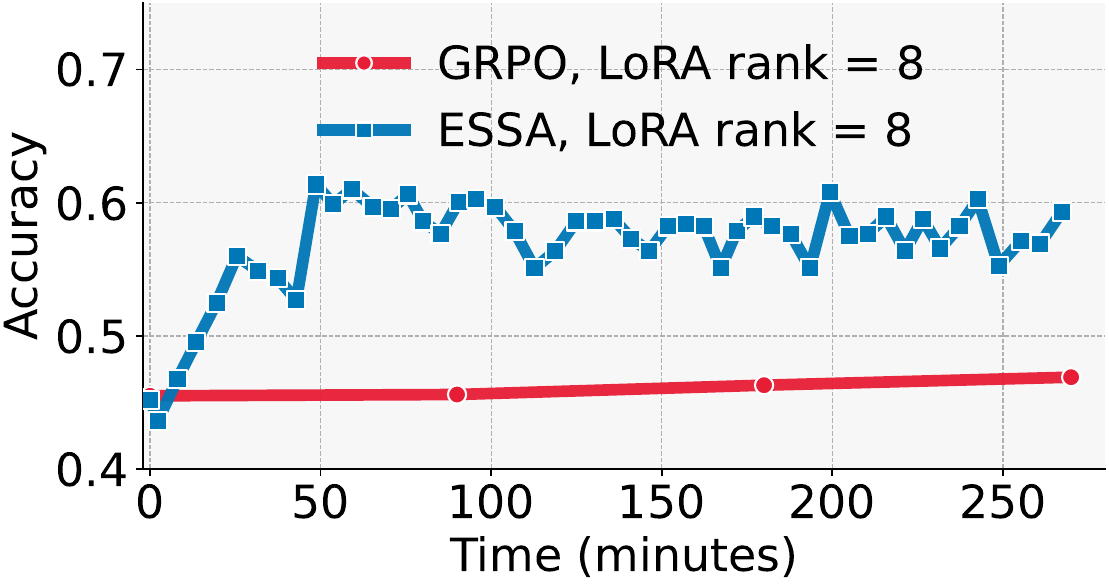}
    \caption{
        Validation strict accuracy over time on \textbf{IFEval}
        with \textbf{Llama-3.1-8B}. ESSA (blue) uses LoRA rank 8,
        population size 24, batch size 500, and $\alpha=1.0$.
        LoRA-GRPO (red) uses LoRA rank 8, learning rate
        $5\!\times\!10^{-5}$, global batch size 512, and
        minibatch size 64. At the checkpoint from the final validation evaluation,
        final strict test accuracy over five independent seeds is
        $0.555\pm0.009$ for ESSA and $0.483\pm0.002$ for
        LoRA-GRPO. Full configurations and policy-generation and
        reward-query counts are reported in Appendix
        Table~\ref{tab:main-reported-configs}; complete run-level
        LoRA-GRPO learning-rate sweep results are reported in
        Appendix Table~\ref{tab:grpo-run-level-ifeval}.
    }
    \label{fig:ifeval}
\end{figure}

\paragraph{Instruction following (IFEval).}
Figure~\ref{fig:ifeval} compares ESSA and LoRA-GRPO on
Llama-3.1-8B/IFEval using strict accuracy. ESSA reaches
approximately 0.6 validation strict accuracy within 60 minutes
and remains near this level, whereas LoRA-GRPO improves more
gradually. At the checkpoint from the final validation evaluation, the final
strict test accuracy is $0.555\pm0.009$ for ESSA and
$0.483\pm0.002$ for LoRA-GRPO, reported over five independent
seeds. Thus, ESSA's advantage over LoRA-GRPO is not specific to
verifiable-reward math and also holds for rule-based instruction
following. Full configurations and policy-generation and
reward-query counts are reported in Appendix
Table~\ref{tab:main-reported-configs}; complete run-level
LoRA-GRPO learning-rate sweep results are reported in Appendix
Table~\ref{tab:grpo-run-level-ifeval}.

\paragraph{Preference-based assistant tuning (HelpSteer2).}
On Llama-3.1-8B/HelpSteer2 with the ArmoRM reward model, which
provides a noisy learned reward signal, both ESSA and LoRA-GRPO
improve over the shared SFT initialization
(Appendix Figure~\ref{fig:helpsteer}). Additional diagnostics (Appendix~\ref{app:helpsteer}, Table~\ref{tab:helpsteer_diagnostics}) 
show that mean response length
increases by 19\% for ESSA and 15\% for LoRA-GRPO. ESSA also
remains substantially closer to the SFT policy in generation
entropy: 0.735 for ESSA versus 1.973 for LoRA-GRPO, compared
with 0.642 for the SFT checkpoint. Distinct-2 rises from 0.201
at SFT to 0.613 for ESSA and 0.733 for LoRA-GRPO; the relative
parameter drift from the SFT LoRA update is 2.403 for ESSA and
4.233 for LoRA-GRPO. Full configurations and
policy-generation and reward-query counts are reported in
Appendix Table~\ref{tab:main-reported-configs}. Cross-domain experiments, in which the SFT initialization and the target alignment domain differ (IFEval $\leftrightarrow$ HelpSteer2),
are reported in Appendix~\ref{app:ood}.

\paragraph{Safety / helpfulness (HH-RLHF).}
We further evaluate ESSA in the safety domain by training
Mistral-7B on Anthropic HH-RLHF~\citep{anthropic_hhrlhf} with
the OpenAssistant DeBERTa-v3 reward model. ESSA reaches a higher
in-domain reward than LoRA-GRPO
(Appendix Figure~\ref{fig:hh_rlhf}). On a held-out set of
safety/helpfulness prompts, an independent GPT-4o side-by-side
judge reports 12\% preference for ESSA, 76\% ties, and 12\%
preference for LoRA-GRPO, indicating parity under a judge that
was not used to produce the training reward. Full configurations
and policy-generation and reward-query counts are reported in
Appendix Table~\ref{tab:main-reported-configs}.

\subsection{Scaling to 32B/72B on advanced math}
\label{sec:scale_math}

Tables~\ref{tab:benchs_32}~and~\ref{tab:benchs_72} report avg@$k$ and pass@8 on six advanced-math benchmarks with Qwen2.5-32B and Qwen2.5-72B. ESSA matches or exceeds LoRA-GRPO on the aggregate scores, including on AIME'24/'25 where backbone capacity matters most. At 72B we run ESSA under INT4 with one candidate per GPU; LoRA-GRPO requires tensor parallelism $(TP=4)$ for the backward pass.

\begin{table}[t]
\centering
\small
\setlength{\tabcolsep}{2pt}
\resizebox{\columnwidth}{!}{%
\begin{tabular}{lcccccc|c}
\toprule
\textbf{Method} & \textbf{MATH} & \textbf{Min.} &
\textbf{Olym.} & \textbf{AI24} & \textbf{AI25} &
\textbf{AMC} & \textbf{Avg} \\
\midrule
\multicolumn{8}{c}{\textbf{avg@k}}\\
LoRA-GRPO
& $81.8\pm0.1$
& $41.2\pm0.1$
& $45.7\pm0.1$
& $14.6\pm0.1$
& $10.0\pm0.1$
& $61.7\pm0.1$
& $42.5\pm0.1$ \\
ESSA
& $\mathbf{82.1\pm0.1}$
& $\mathbf{41.8\pm0.1}$
& $\mathbf{47.6\pm0.1}$
& $\mathbf{17.3\pm0.1}$
& $\mathbf{12.1\pm0.1}$
& $\mathbf{63.3\pm0.1}$
& $\mathbf{44.0\pm0.1}$ \\
\midrule
\multicolumn{8}{c}{\textbf{pass@8}}\\
LoRA-GRPO
& $\mathbf{93.6\pm0.1}$
& $\mathbf{53.3\pm0.1}$
& $65.3\pm0.1$
& $32.9\pm0.1$
& $\mathbf{29.2\pm0.1}$
& $\mathbf{87.3\pm0.1}$
& $\mathbf{60.2\pm0.1}$ \\
ESSA
& $92.6\pm0.1$
& $52.2\pm0.1$
& $\mathbf{66.4\pm0.1}$
& $\mathbf{35.3\pm0.1}$
& $28.0\pm0.1$
& $86.9\pm0.1$
& $\mathbf{60.2\pm0.1}$ \\
\bottomrule
\end{tabular}}
\caption{Results on advanced-math benchmarks with
\textbf{Qwen2.5-32B}. ESSA uses LoRA rank 8; LoRA-GRPO uses
LoRA rank 16. Results are reported as mean $\pm$ standard
deviation over five independent seeds. Rows are grouped by metric:
avg@k and pass@8. For \textbf{AIME'24}, \textbf{AIME'25}, and
\textbf{AMC'23}, avg@k is reported at $k=16$; for the other
benchmarks, $k=8$. The final \textbf{Avg} averages across all six
benchmarks. Full configurations and policy-generation and
reward-query counts are reported in Appendix
Table~\ref{tab:main-reported-configs}.}
\label{tab:benchs_32}
\end{table}

\begin{table}[t]
\centering
\small
\setlength{\tabcolsep}{2pt}
\resizebox{\columnwidth}{!}{%
\begin{tabular}{lcccccc|c}
\toprule
\textbf{Method} & \textbf{MATH} & \textbf{Min.} &
\textbf{Olym.} & \textbf{AI24} & \textbf{AI25} &
\textbf{AMC} & \textbf{Avg} \\
\midrule
\multicolumn{8}{c}{\textbf{avg@k}}\\
LoRA-GRPO
& $\mathbf{83.0\pm0.1}$
& $43.0\pm0.1$
& $\mathbf{49.1\pm0.3}$
& $\mathbf{18.8\pm0.1}$
& $11.0\pm0.5$
& $62.2\pm0.2$
& $44.5\pm0.2$ \\
ESSA
& $82.8\pm0.1$
& $\mathbf{43.5\pm0.5}$
& $48.4\pm0.3$
& $17.9\pm0.3$
& $\mathbf{13.3\pm0.6}$
& $\mathbf{62.5\pm0.3}$
& $\mathbf{44.7\pm0.4}$ \\
\midrule
\multicolumn{8}{c}{\textbf{pass@8}}\\
LoRA-GRPO
& $\mathbf{94.2\pm0.2}$
& $51.8\pm0.2$
& $66.7\pm0.3$
& $\mathbf{37.5\pm0.2}$
& $23.5\pm0.2$
& $85.6\pm0.2$
& $59.9\pm0.2$ \\
ESSA
& $93.0\pm0.2$
& $\mathbf{52.9\pm0.4}$
& $\mathbf{66.8\pm0.4}$
& $37.4\pm0.3$
& $\mathbf{26.8\pm0.2}$
& $\mathbf{85.8\pm0.1}$
& $\mathbf{60.5\pm0.3}$ \\
\bottomrule
\end{tabular}}
\caption{Results on advanced-math benchmarks with
\textbf{Qwen2.5-72B}. ESSA uses LoRA rank 4; LoRA-GRPO uses
LoRA rank 16. Results are reported as mean $\pm$ standard
deviation over five independent seeds. Rows are grouped by metric:
avg@k and pass@8. For \textbf{AIME'24}, \textbf{AIME'25}, and
\textbf{AMC'23}, avg@k is reported at $k=16$; for the other
benchmarks, $k=8$. The final \textbf{Avg} averages across all six
benchmarks. Full configurations and policy-generation and
reward-query counts are reported in Appendix
Table~\ref{tab:main-reported-configs}.}
\label{tab:benchs_72}
\end{table}

\subsection{Sensitivity and precision}
\label{sec:sensitivity}

\paragraph{Hyperparameters.}
A grid sweep over LoRA rank, population size, and $\alpha$ for five \mbox{(backbone, task)} pairs (full plots in Appendix~\ref{app:hyperparams}; training-dynamics curves in Appendix~\ref{app:dynamics}) reveals three consistent trends: (i) accuracy plateaus for $r\!\ge\!2$ in most settings; (ii) populations of 24--96 are sufficient; (iii) accuracy is stable for $\alpha\!\ge\!0.4$. ESSA therefore avoids the multi-knob sensitivity (KL coefficient, clipping, learning rate, mini-batch size) of PPO/GRPO.

\paragraph{Precision.}
Because its optimization loop is inference-only, ESSA runs
natively with INT8- or INT4-quantized weights. On Qwen2.5-32B/PRM800K, reducing precision from BF16 to INT4
decreases MATH500 test accuracy by 1.1 percentage points
(Table~\ref{tab:precision}). Full training curves under each precision are shown in
Appendix~\ref{app:precision}. INT4 execution also enables the
72B configuration to run with one candidate per NVIDIA
H100-equivalent GPU. Full configurations and policy-generation
and reward-query counts are reported in Appendix
Table~\ref{tab:main-reported-configs}.

\begin{table}[t]
\centering
\small
\setlength{\tabcolsep}{8pt}
\begin{tabular}{lc}
\toprule
\textbf{Weight precision}
& \textbf{Test MATH500 accuracy} \\
\midrule
BF16 & $0.845\pm0.004$ \\
INT8 & $0.841\pm0.006$ \\
INT4 & $0.834\pm0.006$ \\
\bottomrule
\end{tabular}
\caption{
    ESSA test accuracy at the checkpoint from the final validation evaluation as
    a function of \textbf{Qwen2.5-32B} weight precision on \textbf{MATH500} after
    training on \textbf{PRM800K}. All experiments use LoRA rank 8,
    population size 64, batch size 256, and $\alpha=1.0$.
    Results are reported as mean $\pm$ standard deviation over
    five independent seeds. Full configurations and
    policy-generation and reward-query counts are reported in
    Appendix Table~\ref{tab:main-reported-configs}.
}
\label{tab:precision}
\end{table}

\section{Discussion}

ESSA shows that the modern post-SFT alignment stage admits a
gradient-free realization that is competitive in quality,
friendly to quantization, and exhibits stronger measured
multi-GPU scaling than the evaluated LoRA-GRPO configuration.
The three design choices (restricting to LoRA, parameterizing by
singular values, and optimizing with CMA-ES) are individually
old, but combined they turn online alignment into a problem where
the per-iteration cost is dominated by inference rather than by
gradient-specific training and communication stages. Our
comparison against Online DPO, PPO, LoRA-GRPO, full-parameter
GRPO, and SVD-GRPO shows that ESSA achieves the strongest
parameter-efficient result on verifiable-reward GSM8K. On
instruction following, ESSA outperforms LoRA-GRPO and remains
competitive with Online DPO and PPO; on the preference-based safety task, it
reaches parity with LoRA-GRPO as judged by GPT-4o. The optimizer
and subspace controls refine the interpretation of these results.
Plain isotropic ES and ARS match or exceed CMA-ES at convergence.
SVD-GRPO and MeZO remain substantially below the population-based
methods on the harder MATH500 task, indicating that merely
restricting a gradient method to the same coordinates does not
reproduce the result. A random orthogonal basis also supports
substantial improvement and comes within 1.5 percentage points of
the SFT-SVD basis under a matched budget, although the SFT-SVD
basis converges faster and reaches a higher plateau. Taken
together, the results support a broader account: ESSA succeeds by
combining a compact SFT-seeded parameter space with
population-based gradient-free search. CMA-ES is a practical
default because of its early progress and limited tuning
requirements, rather than a uniquely necessary component.

\section*{Limitations}

\paragraph{SFT initialization is required.}
ESSA is positioned as a post-SFT alignment stage and relies on a
strong SFT initialization; without any SFT, ESSA still improves
over the base model but does not reach the SFT$\to$ESSA accuracy
(Appendix Figure~\ref{fig:train_qwen_gsm_base}). This dependence
is shared by standard modern alignment pipelines
\citep{NEURIPS2022_b1efde53,lambert2024tulu,llama}, but it does
mean that ESSA does not replace SFT.

\paragraph{Reward-evaluation cost.}
ESSA can use substantially more reward evaluations than
LoRA-GRPO, although the ratio varies across experiments
(Appendix Table~\ref{tab:main-reported-configs}). With cheap
rewards (rule-based verifiers or small reward models), this
additional cost can be offset by the savings from removing the
backward pass and associated training stages; with very expensive
reward models, the higher evaluation count could erode the
wall-clock advantage. The break-even point depends on the ratio
between reward-model cost and policy-model cost and is a useful
subject for future work.

\paragraph{Low-rank assumption.}
The singular-value parameterization assumes that useful alignment
updates can be expressed in a low-dimensional space initialized
from SFT. Our results across eight (backbone, task) pairs support
this assumption in the evaluated settings, but do not establish
that it holds universally. Tasks that genuinely require a
high-rank update may not be reachable by ESSA at the LoRA ranks
we explored.

\paragraph{Mechanistic scope.}
Our controls identify an effective optimization regime but do
not establish a unique explanation for why population-based
search performs well in this space. In particular, covariance
adaptation is not necessary: plain isotropic ES and ARS match or
exceed CMA-ES at convergence. The relative ranking of these
optimizers may also change across tasks, reward-noise levels, and
evaluation budgets. We therefore treat CMA-ES as a practical
default rather than as an essential component of the method.

\paragraph{Backbone diversity.}
We covered Qwen2.5 (7B/Math-7B/32B/72B), Llama-3.1-8B, and
Mistral-7B, but a broader study across additional families
(Gemma, Phi, OLMo) would further strengthen the generality claim.

\section*{Ethical Considerations}

ESSA is a generic alignment method; it inherits the well-known dual-use considerations of any RLHF-style technique. Because ESSA reduces the engineering cost of online alignment, it may lower the barrier to both beneficial and harmful applications. Our HH-RLHF and HelpSteer2 experiments use only publicly available, license-compatible datasets and reward models, and our reward signals correspond to helpfulness/harmlessness preferences released by their original authors. We do not introduce any new dataset or release any new model weights with this work. We did not perform any human-subjects evaluation. The GPT-4o side-by-side comparison in Section~\ref{sec:nlp} was performed via API access in compliance with OpenAI's usage policies. All datasets (GSM8K, PRM800K, IFEval, HelpSteer2, HH-RLHF, AIME, AMC, MinervaMath, OlympiadBench), backbone models (Qwen2.5, Llama-3.1, Mistral-7B), reward models (ArmoRM, OpenAssistant DeBERTa-v3), and software (verl, vLLM, EvoTorch) are released under permissive licenses (MIT, Apache~2.0, CC-BY-4.0) or their respective model community licenses (the Llama~3 and Llama~3.1 community licenses), and our academic-research use complies with each artifact's terms.

\paragraph{Use of AI assistants.}
We used AI assistants (general-purpose LLMs) for grammar checks and short-form text polishing throughout the manuscript, and as coding aides for plotting and table formatting. All scientific claims, experiments, analyses, and design decisions were authored by the human authors, and we take full responsibility for the content.

\bibliography{essa}

\clearpage
\appendix

\begin{algorithm}[H]
\caption{ESSA: Distributed Evolutionary Search over LoRA Singular Values}
\label{alg:essa}
\begin{algorithmic}[1]
\REQUIRE Training set $\mathcal{D}$; SFT-initialized LoRA factors $\{A_i,B_i\}$; fraction $\alpha$; population $P$; GPUs $N$; generations $E$.
\STATE \textbf{SVD init:} for each $A_i,B_i$ compute $A_i = U_{A_i}\Sigma_{A_i}V_{A_i}^{\top}$, $B_i = U_{B_i}\Sigma_{B_i}V_{B_i}^{\top}$; freeze $U,V$; concatenate top-$\alpha$ singular values into $\sigma\!\in\!\mathbb{R}^d$.
\STATE Initialize $N$ workers with identical seeds and a common CMA-ES state $(m_0,\sigma_0,C_0)$.
\FOR{$t = 0,\dots,E-1$}
    \FOR{worker $k=1,\dots,N$ \textbf{in parallel}}
        \STATE Sample candidate offsets $\{\delta^{(k,j)}\}$ from shared seeds.
        \FOR{each local candidate $j$}
           \STATE Reconstruct $A_i'\!=\!U_{A_i}\mathrm{diag}(\sigma_{A_i}\!+\!\delta_{A_i})V_{A_i}^{\top}$ and similarly $B_i'$; set $W_i'\!=\!W_i + B_i'A_i'$.
           \STATE Sample mini-batch $\mathcal{D}_{\text{mini}}\!\subset\!\mathcal{D}$ and compute reward $F$.
        \ENDFOR
    \ENDFOR
    \STATE All workers exchange $\langle\text{seed},F\rangle$ pairs.
    \STATE Each worker reconstructs the same population and performs one CMA-ES update of $(m,\sigma,C)$.
\ENDFOR
\STATE Return aligned model with final singular values.
\end{algorithmic}
\end{algorithm}

\section{ESSA Algorithm}
\label{app:algorithm}

Algorithm~\ref{alg:essa} formalises the full ESSA training loop sketched in Section~\ref{sec:method}: SVD initialisation of the LoRA factors from the SFT checkpoint, parallel candidate sampling across $N$ workers, fitness evaluation by a single forward pass per candidate, and a CMA-ES update per generation. Workers exchange only $\langle\text{seed}, F\rangle$ pairs, keeping per-step communication constant in model size.

\section{Detailed Iteration-Time Analysis}
\label{app:theory}

This appendix expands the iteration-latency analysis sketched in
Section~\ref{sec:scaling}, comparing one optimization step of an
idealized full-parameter gradient-based RLHF pipeline against one
ESSA step. The gradient pipeline incurs an all-reduce of a
model-sized tensor, whereas ESSA exchanges only small per-candidate
records, such as random seeds or identifiers and scalar rewards.
This analysis does not model the LoRA-GRPO baseline measured in
Section~\ref{sec:scaling}. The proposition below isolates a
sufficient condition under which ESSA is strictly faster per
iteration under the idealized timing model.

\begin{proposition}[ESSA iteration-time advantage]
\label{prop:essa_scaling}
Assume $m_{\mathrm{fb}}\geq1$, $m_{\mathrm{gen}}\leq G$, and that
the total ESSA synchronization payload satisfies
$M_{\mathrm{essa}}\leq M_{\mathrm{params}}$. If the ESSA population
size $N_{\mathrm{pop}}$ and per-candidate batch size
$B^{\mathrm{essa}}$ satisfy
\begin{equation}
\label{eq:essa_bound_app}
\begin{aligned}
N_{\mathrm{pop}} B^{\mathrm{essa}}
&< B^{\mathrm{grad}}\!\left(
1 + \frac{\eta_{\mathrm{fb}}}
         {G\,\eta_{\mathrm{gen}}}\right) \\
&\quad + \frac{(G-1)M_{\mathrm{params}}}
     {\mathrm{peak\_bw}\,G\,\eta_{\mathrm{gen}}},
\end{aligned}
\end{equation}
then the modeled ESSA iteration time is strictly smaller than that
of the idealized full-parameter gradient pipeline:
$T^{\mathrm{essa}}<T^{\mathrm{grad}}$.
\end{proposition}

To prove the proposition, we (i) introduce notation, (ii) bound
per-iteration computation and communication, (iii) optimize the
device split between training and generation, and (iv) derive a
conservative population-size bound under which ESSA is guaranteed
to be faster within the timing model.

\paragraph{Notation.}
Let $B^{\mathrm{grad}}$ be the global batch size used in gradient
methods, $B^{\mathrm{essa}}$ the batch size processed by a single
population instance, $b_{\mathrm{fb}},b_{\mathrm{gen}}$ the
microbatch sizes, and $m_{\mathrm{fb}},m_{\mathrm{gen}}$ the numbers
of GPUs per model instance for training and generation. Let $G$ be
the total number of GPUs, with $G_{\mathrm{fb}}$ and
$G_{\mathrm{gen}}$ allocated to training and generation,
respectively. Let $\tau_{\mathrm{fb}}(b)$ and
$\tau_{\mathrm{gen}}(b)$ be the forward-backward and generation
microbatch times, and define
$\eta_{\mathrm{fb}}=\tau_{\mathrm{fb}}(b_{\mathrm{fb}})
/b_{\mathrm{fb}}$ and
$\eta_{\mathrm{gen}}=\tau_{\mathrm{gen}}(b_{\mathrm{gen}})
/b_{\mathrm{gen}}$ as the corresponding per-sample times. Finally,
let
$k^{\mathrm{par}}_{\mathrm{fb}}=G_{\mathrm{fb}}/m_{\mathrm{fb}}$
and
$k^{\mathrm{par}}_{\mathrm{gen}}
=G_{\mathrm{gen}}/m_{\mathrm{gen}}$
be the numbers of microbatches processed in parallel.

Then
$T^{\mathrm{grad}}
=T^{\mathrm{grad}}_{\mathrm{fb\mbox{-}gen}}
+T^{\mathrm{grad}}_{\mathrm{comm}}$
and
$T^{\mathrm{essa}}
=T^{\mathrm{essa}}_{\mathrm{gen}}
+T^{\mathrm{essa}}_{\mathrm{comm}}$.
Under ideal asynchronous training and generation,
\[
T^{\mathrm{grad}}_{\mathrm{fb\mbox{-}gen}}
=
\max\!\left(
T^{\mathrm{grad}}_{\mathrm{fb}},
T^{\mathrm{grad}}_{\mathrm{gen}}
\right).
\]

\paragraph{Computation.}
Processing $B^{\mathrm{grad}}$ samples requires
$B^{\mathrm{grad}}/b_{\mathrm{fb}}$ microbatches, of which
$k^{\mathrm{par}}_{\mathrm{fb}}$ run in parallel:
\begin{align*}
T^{\mathrm{grad}}_{\mathrm{fb}}
&=
\frac{B^{\mathrm{grad}}m_{\mathrm{fb}}}
     {G_{\mathrm{fb}}}\eta_{\mathrm{fb}},\\
T^{\mathrm{grad}}_{\mathrm{gen}}
&=
\frac{B^{\mathrm{grad}}m_{\mathrm{gen}}}
     {G_{\mathrm{gen}}}\eta_{\mathrm{gen}}.
\end{align*}
For ESSA, the entire population is evaluated by generation only:
\[
T^{\mathrm{essa}}_{\mathrm{gen}}
=
\frac{N_{\mathrm{pop}}B^{\mathrm{essa}}m_{\mathrm{gen}}}
     {G}\eta_{\mathrm{gen}}.
\]

\paragraph{Communication.}
Let $M_{\mathrm{params}}$ be the model-parameter size in bytes.
For the idealized full-parameter gradient pipeline, a ring
all-reduce consists of a reduce-scatter and an all-gather:
\[
T^{\mathrm{grad}}_{\mathrm{comm}}
=
2\cdot
\frac{M_{\mathrm{params}}(G-1)}
     {G\,\mathrm{peak\_bw}}.
\]

Let $M_{\mathrm{essa}}$ denote the total payload synchronized by
ESSA in one iteration, including the scalar rewards and any random
seeds or candidate identifiers required to reconstruct the
population. This payload grows with the population size but remains
negligible relative to a model-sized tensor for the configurations
considered here. Under the same bandwidth model,
\[
T^{\mathrm{essa}}_{\mathrm{comm}}
=
\frac{M_{\mathrm{essa}}(G-1)}
     {G\,\mathrm{peak\_bw}}.
\]

\paragraph{Optimal device split.}
Let $\theta\in(0,1)$ be the fraction of devices used for training,
so that $G_{\mathrm{fb}}=\theta G$ and
$G_{\mathrm{gen}}=(1-\theta)G$. Then
\[
T^{\mathrm{grad}}_{\mathrm{fb\mbox{-}gen}}(\theta)
=
\frac{B^{\mathrm{grad}}}{G}
\max\!\left(
\frac{m_{\mathrm{fb}}\eta_{\mathrm{fb}}}{\theta},
\frac{m_{\mathrm{gen}}\eta_{\mathrm{gen}}}{1-\theta}
\right).
\]

\begin{lemma}[Optimal split]
\label{lem:theta}
The minimum of
$T^{\mathrm{grad}}_{\mathrm{fb\mbox{-}gen}}(\theta)$ over
$\theta\in(0,1)$ is attained at
\begin{align*}
\theta^\star
&=
\frac{m_{\mathrm{fb}}\eta_{\mathrm{fb}}}
     {m_{\mathrm{fb}}\eta_{\mathrm{fb}}
      +m_{\mathrm{gen}}\eta_{\mathrm{gen}}},\\
T^{\mathrm{grad}}_{\mathrm{fb\mbox{-}gen}}(\theta^\star)
&=
\frac{B^{\mathrm{grad}}}{G}
\left(
m_{\mathrm{fb}}\eta_{\mathrm{fb}}
+m_{\mathrm{gen}}\eta_{\mathrm{gen}}
\right).
\end{align*}
\end{lemma}

\begin{proof}
The function
$\max(a/\theta,b/(1-\theta))$ is minimized when its two arguments
are equal:
$a/\theta=b/(1-\theta)$, where
$a=m_{\mathrm{fb}}\eta_{\mathrm{fb}}$ and
$b=m_{\mathrm{gen}}\eta_{\mathrm{gen}}$.
\end{proof}

Define the ideal gradient iteration time under perfect scheduling:
\begin{align*}
T_\star^{\mathrm{grad}}
&=
\frac{B^{\mathrm{grad}}}{G}
\left(
m_{\mathrm{fb}}\eta_{\mathrm{fb}}
+m_{\mathrm{gen}}\eta_{\mathrm{gen}}
\right)\\
&\quad+
2\cdot
\frac{M_{\mathrm{params}}(G-1)}
     {G\,\mathrm{peak\_bw}}.
\end{align*}
Under the timing model,
$T^{\mathrm{grad}}\geq T_\star^{\mathrm{grad}}$.

\begin{theorem}[ESSA iteration is faster under a conservative bound]
\label{thm:essa_faster}
Assume $m_{\mathrm{fb}}\geq1$, $m_{\mathrm{gen}}\leq G$, and
$M_{\mathrm{essa}}\leq M_{\mathrm{params}}$. If
Equation~\ref{eq:essa_bound_app} holds, then
\[
T^{\mathrm{essa}}
<
T_\star^{\mathrm{grad}}
\leq
T^{\mathrm{grad}}
\]
under the timing model defined above.
\end{theorem}

\begin{proof}
Substituting the computation and communication terms into
$T^{\mathrm{essa}}<T_\star^{\mathrm{grad}}$ gives
\begin{align}
N_{\mathrm{pop}}B^{\mathrm{essa}}
<
&\;
B^{\mathrm{grad}}
\left(
1+
\frac{m_{\mathrm{fb}}\eta_{\mathrm{fb}}}
     {m_{\mathrm{gen}}\eta_{\mathrm{gen}}}
\right)
\nonumber\\
&+
\frac{
(2M_{\mathrm{params}}-M_{\mathrm{essa}})(G-1)
}{
\mathrm{peak\_bw}\,
m_{\mathrm{gen}}\eta_{\mathrm{gen}}
}.
\label{eq:essa_exact_bound}
\end{align}
Because $m_{\mathrm{fb}}\geq1$ and
$m_{\mathrm{gen}}\leq G$,
\[
\frac{m_{\mathrm{fb}}}{m_{\mathrm{gen}}}
\geq\frac{1}{G}.
\]
Moreover, $M_{\mathrm{essa}}\leq M_{\mathrm{params}}$ implies
$2M_{\mathrm{params}}-M_{\mathrm{essa}}
\geq M_{\mathrm{params}}$, and
$m_{\mathrm{gen}}\leq G$ implies
$1/m_{\mathrm{gen}}\geq1/G$.
Therefore, Equation~\ref{eq:essa_bound_app} is a sufficient
condition for Equation~\ref{eq:essa_exact_bound}, and hence for
$T^{\mathrm{essa}}<T_\star^{\mathrm{grad}}$.
The optimal-split construction gives
$T_\star^{\mathrm{grad}}\leq T^{\mathrm{grad}}$.
\end{proof}

\paragraph{Discussion.}
Theorem~\ref{thm:essa_faster} gives a conservative threshold on the
ESSA population size below which its modeled iteration time is
smaller than that of the idealized full-parameter gradient pipeline.
The right-hand side of Equation~\ref{eq:essa_bound_app} grows
linearly with $M_{\mathrm{params}}$ when the other quantities are
fixed. Its dependence on $G$ occurs through $1/G$ and $(G-1)/G$ and
therefore saturates as the cluster size increases. Accordingly, the
bound supports the model-size argument, while the empirical
cluster-scaling behavior is established separately by the wall-clock
measurements in Section~\ref{sec:scaling}.

\begin{figure}[h]
    \centering
    \begin{subfigure}{0.9\linewidth}
        \includegraphics[width=\linewidth]{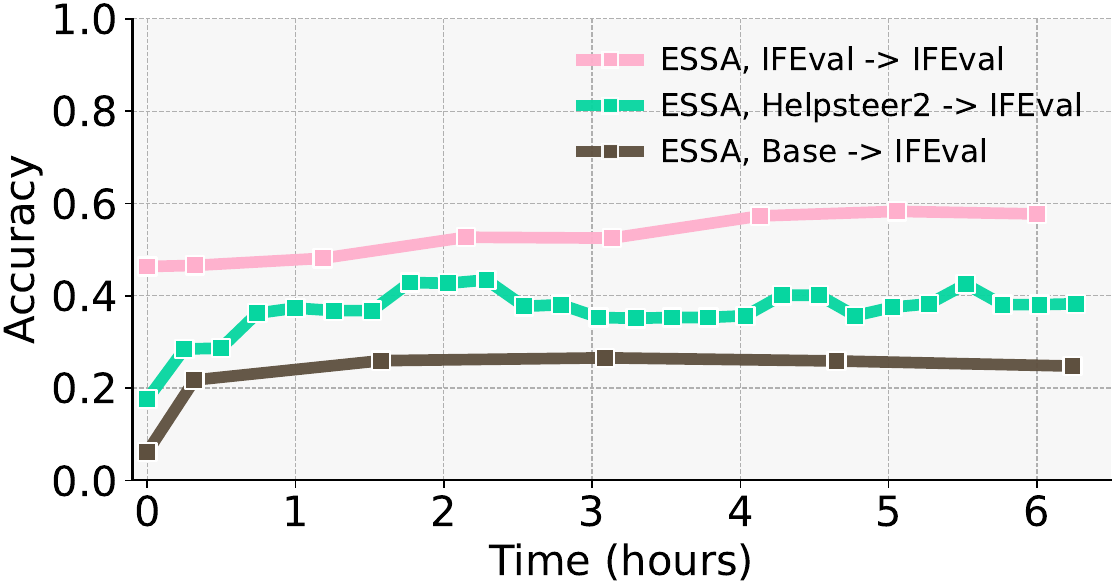}
        \caption{LoRA rank 8, population size 48, $\alpha\!=\!1.0$}
    \end{subfigure}\\[4pt]
    \begin{subfigure}{0.9\linewidth}
        \includegraphics[width=\linewidth]{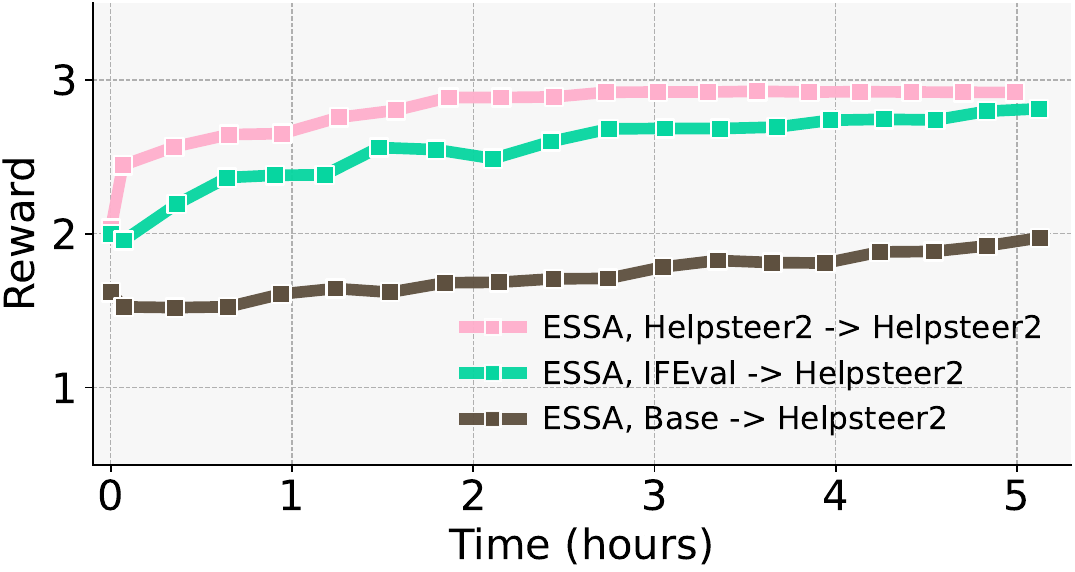}
        \caption{LoRA rank 8, population size 36, $\alpha\!=\!1.0$}
    \end{subfigure}
    \caption{Validation metrics over time on \textbf{IFEval} (a) and \textbf{HelpSteer2} (b) with \textbf{Llama-3.1-8B}. (a) IFEval $\to$ IFEval (pink), HelpSteer2 $\to$ IFEval (green), base model $\to$ IFEval (brown) experiments, batch size 500. (b): HelpSteer2 $\to$ HelpSteer2 (pink), IFEval $\to$ HelpSteer2 (green), base model $\to$ HelpSteer2 (brown) experiments, batch size 100.}
    \label{fig:ood_drift}
\end{figure}

\section{Cross-Domain Robustness}
\label{app:ood}

This section extends the IFEval and HelpSteer2 experiments in
Section~\ref{sec:nlp} and the LoRA-budget results in
Section~\ref{sec:main_comparison} by evaluating ESSA when the SFT
initialization and the target alignment task come from different
domains.

To assess robustness under distribution shift, we conduct additional
cross-domain experiments using two datasets with substantially
different reward structures: HelpSteer2 and IFEval. HelpSteer2
encourages general-purpose response quality through a learned reward
model, whereas IFEval evaluates strict constraint satisfaction using
rule-based criteria. Moving between these domains therefore tests
whether ESSA remains effective when the SFT warm-start and the target
alignment domain differ. The HelpSteer2 experiments use the same
RLHFlow/ArmoRM-Llama3-8B-v0.1 reward model as the corresponding
main-text experiment. The IFEval experiments train on the same
IFEval-like dataset used in the main text.

For each target dataset $B$, we evaluate three configurations. The
first performs SFT on dataset $A$ and subsequently applies ESSA to
dataset $B$ ($A\rightarrow B$). The second performs both SFT and ESSA
on dataset $B$ ($B\rightarrow B$). The third applies ESSA directly
to dataset $B$ without an SFT warm-start
(Base model $\rightarrow B$); LoRA factor $A$ is initialized using
Kaiming-uniform initialization~\citep{he}, while factor $B$ is initialized to zero.

In both evaluated directions (Figure~\ref{fig:ood_drift}), ESSA improves performance on the target
dataset even when the SFT checkpoint originates from the other domain.
In the IFEval-to-HelpSteer2 condition, the gap between cross-domain
and in-domain SFT is modest, indicating that the initialization learned
from the IFEval-like dataset provides useful directions for subsequent
optimization on HelpSteer2. In the opposite direction, the gap is
larger, consistent with HelpSteer2 not explicitly teaching the strict
formatting and constraint-following behavior required by IFEval.
Nevertheless, ESSA improves over the
Base model $\rightarrow$ IFEval configuration initialized with random
LoRA parameters.

In both directions, the SFT-initialized configurations outperform
optimization from the base model, indicating that an out-of-domain SFT
checkpoint can still provide useful structure for ESSA's evolutionary
search. These results show that ESSA remains effective under the two
evaluated distribution shifts, although in-domain SFT remains the
strongest initialization.

\section{Toy Example on MNIST}
\label{app:mnist}

This section provides a minimal illustration of the
singular-value-only parameterization introduced in
Section~\ref{sec:method} and complements the subspace controls in
Section~\ref{sec:optimizer_controls}.

A natural concern with a singular-value-only parameterization is that
it may be too restrictive in a simple linear or logistic-regression
setting. When the singular directions of a fixed low-rank update are
held constant, only their magnitudes can change, which may appear to
provide insufficient flexibility for meaningful adaptation.

ESSA's effectiveness for LLM alignment relies on the structured
representations already present in the pretrained backbone and refined
by the downstream SFT initialization. To illustrate this distinction,
we design an MNIST experiment that mirrors the essential initialization
and parameterization stages of the LLM experiments within a minimal
linear model. The MNIST experiment proceeds as follows:

\begin{figure}[!t]
    \centering

    \begin{subfigure}{0.7\linewidth}
        \centering
        \includegraphics[width=\linewidth]{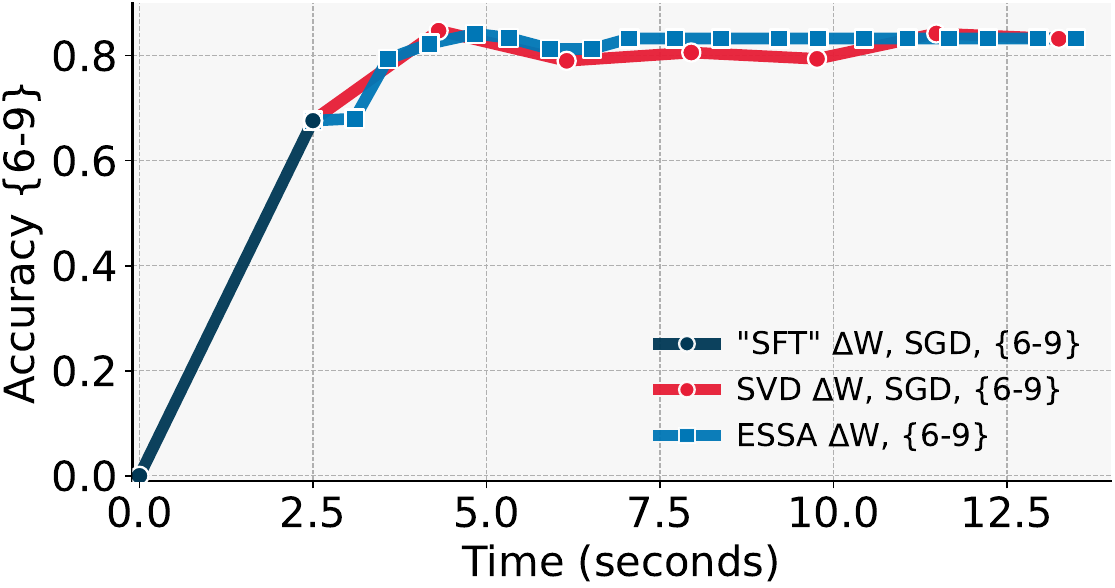}
        \caption{$W = W' + \Delta W$}
        \label{fig:mnist_pretrain_sft}
    \end{subfigure}

    \vspace{0.5em}

    \begin{subfigure}{0.7\linewidth}
        \centering
        \includegraphics[width=\linewidth]{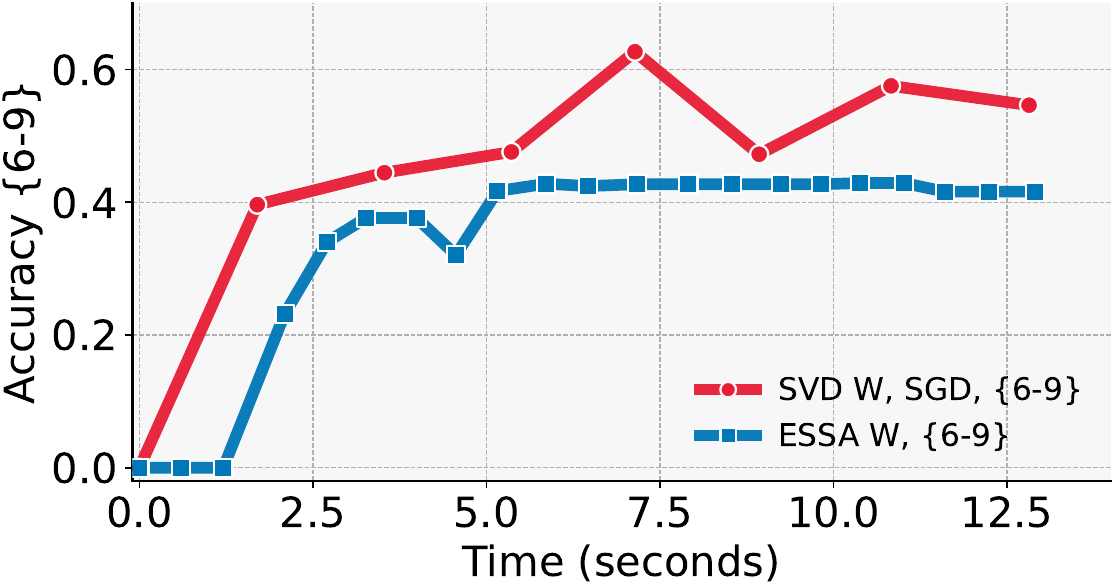}
        \caption{$W = W'$}
        \label{fig:mnist_pretrain_only}
    \end{subfigure}

    \vspace{0.5em}

    \begin{subfigure}{0.7\linewidth}
        \centering
        \includegraphics[width=\linewidth]{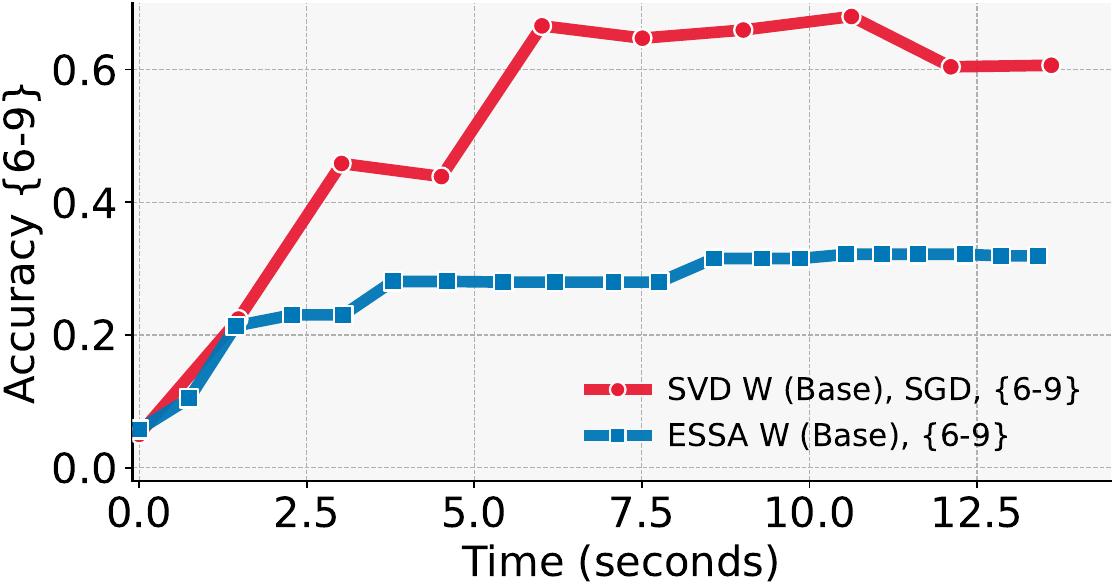}
        \caption{$W \sim \mathcal{N}(0, 2/n)$}
        \label{fig:mnist_random_init}
    \end{subfigure}

    \caption{
        Validation accuracy over time on \textbf{MNIST}, evaluated on the
        test set containing digits 6--9, using a single matrix
        $W \in \mathbb{R}^{n \times m}$ with $n=784$ and $m=10$.
        Panels (a)--(c) correspond to the setups
        $W = W' + \Delta W$, $W = W'$, and
        $W \sim \mathcal{N}(0, 2/n)$, respectively, where $W'$ denotes
        the pretrained base weights. The training set contains 400 examples.
        ESSA (blue): solution length 10, population size 24, and batch size 64.
        SGD (red): learning rate 0.1 and batch size 64.
    }
    \label{fig:essa_mnist}
\end{figure}

\begin{enumerate}
    \item At the pretrain stage (analog of LLM pretraining) we train a logistic regression model on MNIST digits 0-5 (train size 5000) using SGD optimizer. We obtain a "base model" $W = W'$, where $W'$ is the single matrix of size 784 by 10, that knows something about the data distribution, similar to LLM pretraining.
    \item At the "SFT" stage (analogue of LoRA SFT) we take 400 samples of digits 6-9, initialize a matrix update $\Delta W$ with Kaiming init, and train only $\Delta W$ using SGD: $W = W' + \Delta W$, where $\Delta W$ is the single matrix of size 784 by 10. 
    \item At the last stage we decomposed $\Delta W = B S A$ and train only the singular values $S$ using ES or SGD optimizer. ESSA optimizes singular values directly using train accuracy, while the SGD baseline optimized them through the standard cross-entropy loss.
\end{enumerate}

In Figure \ref{fig:mnist_pretrain_sft}, ESSA follows the behavior of SGD on the MNIST task with high fidelity. This demonstrates that once the singular vectors encode useful directions, modulating only the associated singular amplitudes is already sufficient to achieve effective adaptation. In Figure \ref{fig:mnist_pretrain_only}, we remove the additive update entirely. We decompose the pretrained weights $W = W'= B'S'A'$, and train only the singular values $S'$ using ESSA and SGD. Even in this more constrained setup, ESSA achieved performance comparable to SGD. The pretrained matrix contained enough structure that rebalancing its singular directions remained a viable strategy for improving accuracy. Finally, in Figure \ref{fig:mnist_random_init}, we examine the case without pretraining or SFT, where the model is initialized with Kaiming-normal weights $W \sim \mathcal{N}(0, 2/n)$ and the singular vectors carry no semantic information. In this setup ESSA improves only slowly, while SGD makes substantially faster progress. Without a structured subspace, changing only singular values is insufficient to emulate full learning dynamics.

These experiments demonstrate that the singular-value parameterization is effective precisely in the settings where LLMs operate: pretrained models with downstream SFT produce meaningful low-rank directions, and ESSA leverages this structure by adjusting their relative scaling. This behavior is consistent with the results observed in our large-scale LLM experiments.

\newpage 

\begin{figure}[h]
  \centering
  \includegraphics[width=0.7\linewidth]{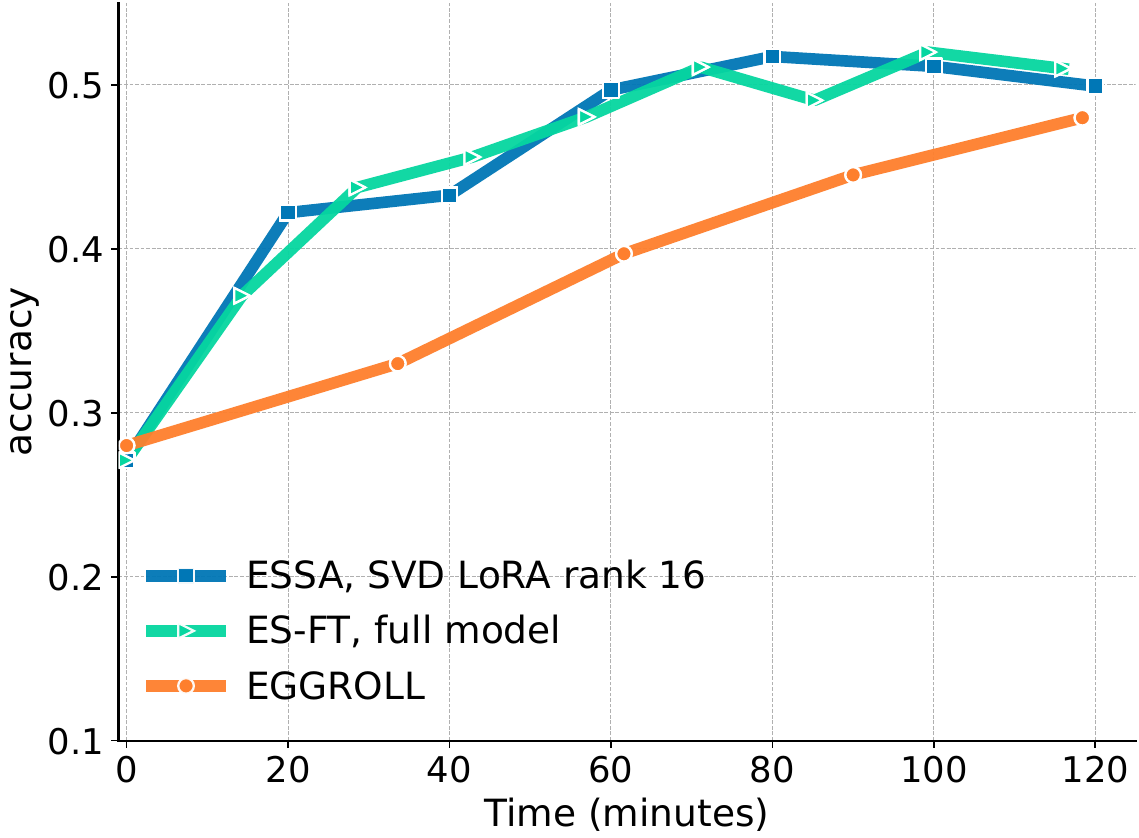}
  \caption{Validation accuracy over time on \textbf{Countdown} with
  \textbf{Qwen2.5-7B} using eight GPUs. ESSA (blue): population size
  48, $\alpha=1.0$, and batch size 200. ES-FT (green):
  $\sigma=1\times10^{-3}$, $\texttt{lr}=1\times10^{-4}$, population size
  48, and batch size 200. EGGROLL (orange):
  $\sigma=1\times10^{-3}$, rank 1, and batch size 200.}
  \label{fig:other}
\end{figure}

\begin{table}[h]
\centering
\small
\resizebox{\columnwidth}{!}{%
\begin{tabular}{lcl}
\toprule
\textbf{Method} & \textbf{Search dim.} & \textbf{Tasks} \\
\midrule
ESSA (ours) & $\mathcal{O}(r)$ &
\begin{tabular}[c]{@{}l@{}}
GSM8K, PRM800K, IFEval,\\
HelpSteer2, HH-RLHF, Countdown
\end{tabular} \\
\midrule
ES-FT~\citep{esft} & $\mathcal{O}(nm)$ &
\begin{tabular}[c]{@{}l@{}}
Countdown, MATH,\\
ARC-AGI, Sudoku
\end{tabular} \\
EGGROLL~\citep{eggroll} & $\mathcal{O}(r(n{+}m))$ &
\begin{tabular}[c]{@{}l@{}}
Countdown, GSM8K, MATH500,\\
AIME, OlympiadBench, AMC
\end{tabular} \\
\bottomrule
\end{tabular}}
\caption{ES-based methods for LLM fine-tuning. ESSA's search
dimension is $\mathcal{O}(r)$ per LoRA factor, substantially smaller
than that of full-parameter ES
(ES-FT, $\mathcal{O}(M_{\mathrm{params}})$) and LoRA-coefficient ES
(EGGROLL, $\mathcal{O}(r(m{+}n))$ per matrix). For Qwen2.5-7B with
28 layers, four adapted attention matrices per layer, and $r=8$,
ESSA exposes
$28\times4\times2r\approx1.8\mathrm{K}$ singular-value coordinates:
the two LoRA factors $A$ and $B$ each contribute $r$ coordinates per
adapted matrix. ES-FT searches the full ${\sim}7$B model parameters,
whereas EGGROLL searches
$28\times4\times r(m{+}n)\approx5\mathrm{M}$ LoRA coefficients.}
\label{tab:es_comparison}
\end{table}

\section{Other ES Methods on Countdown}
\label{app:other}

This section provides a cross-task extension of the alternative
gradient-free optimizer comparison in
Section~\ref{sec:optimizer_controls}. Whereas the main-text
experiments compare optimizers on GSM8K and
PRM800K$\rightarrow$MATH500, the following experiment compares ESSA
with two existing ES-based LLM fine-tuning methods on Countdown.

Figure~\ref{fig:other} compares ESSA, ES-FT, and EGGROLL for
optimizing Qwen2.5-7B on Countdown using eight GPUs.
Table~\ref{tab:es_comparison} summarizes their search dimensions
and task coverage. ESSA substantially outperforms EGGROLL while
matching the convergence of ES-FT, which fine-tunes the full model.

\clearpage

\section{Hyperparameter Sensitivity}
\label{app:hyperparams}

This section provides the complete results supporting the
hyperparameter-sensitivity analysis in
Section~\ref{sec:sensitivity}. The GSM8K experiments additionally
complement the LoRA-budget results in
Section~\ref{sec:main_comparison}; the PRM800K experiments complement
the optimizer and subspace analysis in
Section~\ref{sec:optimizer_controls}; and the IFEval experiments
complement the instruction-following results in
Section~\ref{sec:nlp}.

Figures~\ref{fig:essa_hypers_qwen_math_gsm}--\ref{fig:essa_hypers_llama_ifeval} present five sensitivity studies:
Qwen2.5-Math-7B and Qwen2.5-7B on GSM8K, Qwen2.5-Math-7B and
Qwen2.5-7B on PRM800K, and Llama-3.1-8B on IFEval.

The five sweeps support three trends summarized in the main text:
(i) accuracy peaks at low LoRA ranks
($r\in\{2,4,8\}$) in four of the five settings, while
Qwen2.5-Math-7B/GSM8K favors a larger rank;
(ii) populations of 24--96 are generally sufficient, with diminishing
returns beyond 96; and
(iii) accuracy is broadly stable for $\alpha\geq0.4$, indicating
limited sensitivity to the fraction of singular-value coordinates
optimized.

\subsubsection{Qwen2.5-Math-7B on GSM8K}

This experiment supports Section~\ref{sec:sensitivity} and extends
the GSM8K results in Section~\ref{sec:main_comparison} to
Qwen2.5-Math-7B. Figure~\ref{fig:essa_hypers_qwen_math_gsm}
shows that accuracy peaks at 0.896 for LoRA rank 32 and population
size 96. Moderate ranks also perform strongly: ranks 4, 8, and 16
reach accuracies of 0.886, 0.886, and 0.882, respectively, at
population size 48. Population size 8 generally underperforms,
whereas performance remains relatively stable across population
sizes 24--192.

When the population size is fixed to the best value for each rank,
ESSA maintains high accuracy across nearly all evaluated values of
$\alpha$. Rank 32 reaches 0.899 at $\alpha=0.1$, rank 16 reaches
0.891 at $\alpha=0.6$, and rank 8 reaches 0.886 at $\alpha=1.0$.
Smaller fractions can therefore match or outperform
$\alpha=1.0$, indicating that optimizing all singular-value
coordinates is not required for strong performance.

\subsubsection{Qwen2.5-7B on GSM8K}

This experiment provides the sensitivity analysis for the
Qwen2.5-7B/GSM8K setting used in
Section~\ref{sec:main_comparison}. Figure~\ref{fig:essa_hypers_qwen_gsm}
shows stable performance across a broad range of configurations,
with accuracy peaking at 0.893 for LoRA rank 2 and population size
96. LoRA ranks 2--4 achieve the strongest performance, while larger
ranks provide diminishing or slightly degraded results.

With population size fixed, ESSA maintains accuracy of approximately
0.87--0.89 across the evaluated values of $\alpha$, indicating
robustness to the fraction of singular-value coordinates optimized.

\subsubsection{Qwen2.5-Math-7B on PRM800K}

This experiment supports Section~\ref{sec:sensitivity} and
complements the PRM800K$\rightarrow$MATH500 analysis in
Section~\ref{sec:optimizer_controls}.
Figure~\ref{fig:essa_hypers_qwen_math_prm} shows that accuracy peaks
at 0.784 for LoRA rank 2 and population size 24. Higher ranks or
very large populations gradually reduce performance, indicating
that increasing the search dimension does not improve the result.

With population size fixed, the highest score in the corresponding
$\alpha$ sweep is 0.616 at LoRA rank 8 and $\alpha=1.0$, while
intermediate values of $\alpha$ remain competitive.

\subsubsection{Qwen2.5-7B on PRM800K}

This experiment provides additional sensitivity results for the
Qwen2.5-7B/PRM800K setting analyzed in
Section~\ref{sec:optimizer_controls}.
Figure~\ref{fig:essa_hypers_qwen_prm} shows that accuracy peaks at
0.748 for LoRA rank 2 and population size 48. Performance decreases
for larger ranks or oversized populations, indicating diminishing
returns beyond a compact search space.

With population size fixed, intermediate values
$\alpha\approx0.4$--$0.8$ perform best, reaching up to 0.696
accuracy at LoRA rank 8, while very small fractions and
$\alpha=1.0$ perform slightly worse.

\subsubsection{Llama-3.1-8B on IFEval}

This experiment provides the sensitivity analysis for the
Llama-3.1-8B/IFEval setting used in
Sections~\ref{sec:main_comparison} and~\ref{sec:nlp}.
Figure~\ref{fig:essa_hypers_llama_ifeval} shows stable performance
across configurations, with the highest accuracy of 0.614 achieved
at LoRA rank 8 and population size 24. Ranks 4--16 yield comparable
results, while rank 32 performs slightly worse, indicating that
low or moderate ranks and population sizes are sufficient in this
setting.

\clearpage

\begin{figure*}[p]
    \centering
    \begin{subfigure}{0.48\textwidth}
        \centering
        \includegraphics[width=\linewidth]
            {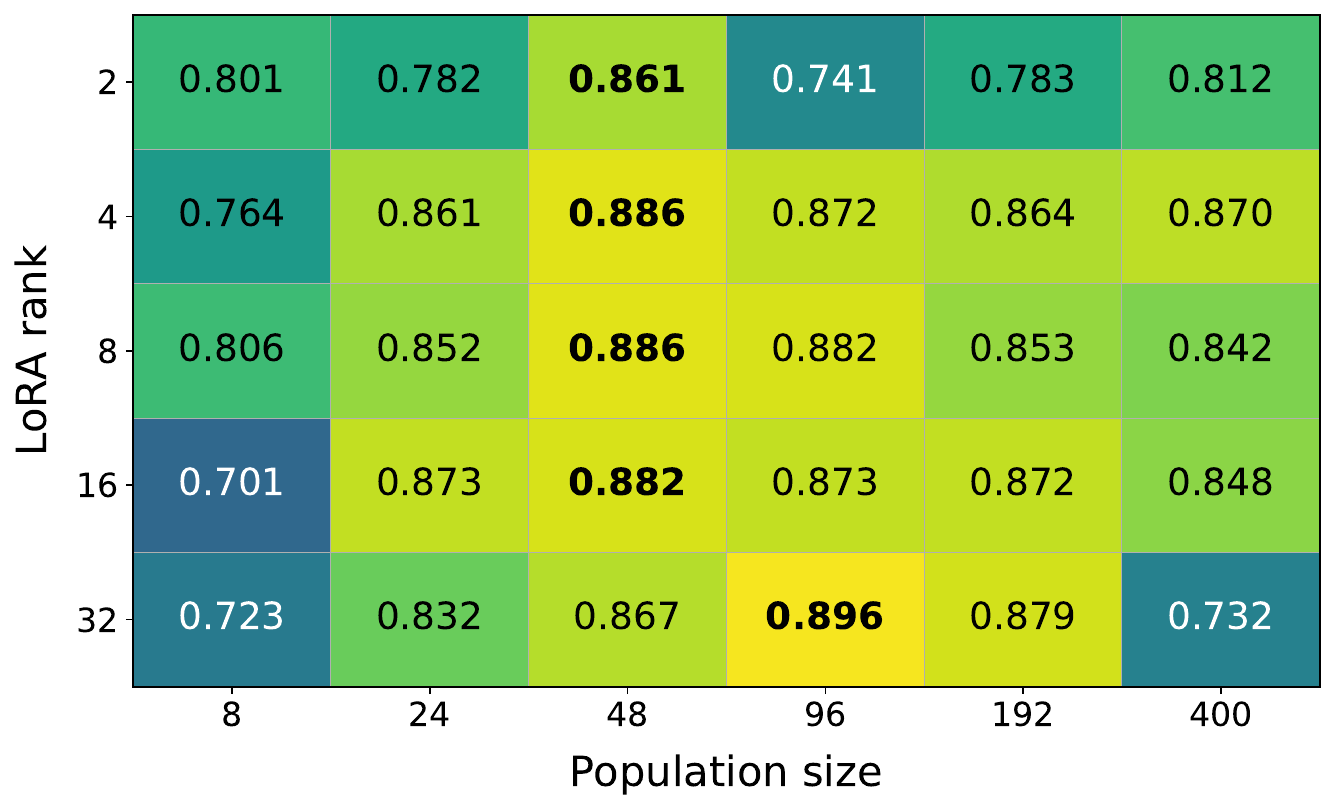}
        \caption{LoRA rank vs.\ population size}
    \end{subfigure}
    \hfill
    \begin{subfigure}{0.48\textwidth}
        \centering
        \includegraphics[width=\linewidth]
            {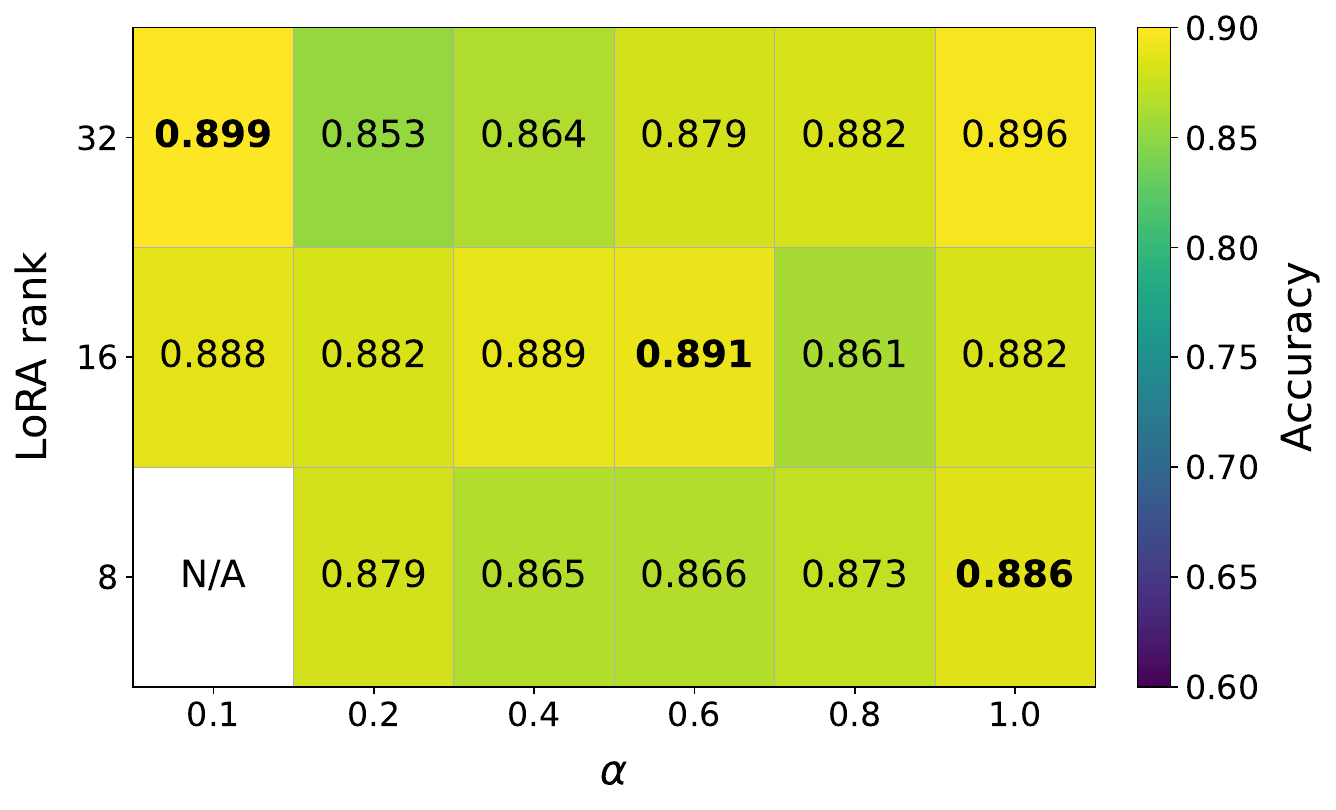}
        \caption{Fraction $\alpha$ at the best (rank, pop.) cell}
    \end{subfigure}
    \caption{Hyperparameter sensitivity of ESSA on \textbf{Qwen2.5-Math-7B} for \textbf{GSM8K}. Batch size 100. 
\textbf{(a)} Accuracy when varying LoRA rank and population size.
\textbf{(b)} For each LoRA rank, the population size is fixed to the best value found in (a),
while the percentage $\alpha$ of trainable singular values is varied.
This illustrates how ESSA performance depends jointly on adapter rank
and the fraction of singular values optimized. The single white cell occurs because for LoRA rank~8 and $\alpha\!=\!0.1$,
rounding down yields zero trainable singular values, so no valid accuracy is reported.}
    \label{fig:essa_hypers_qwen_math_gsm}
\end{figure*}

\begin{figure*}[p]
    \centering
    \begin{subfigure}{0.48\textwidth}
        \centering
        \includegraphics[width=\linewidth]
            {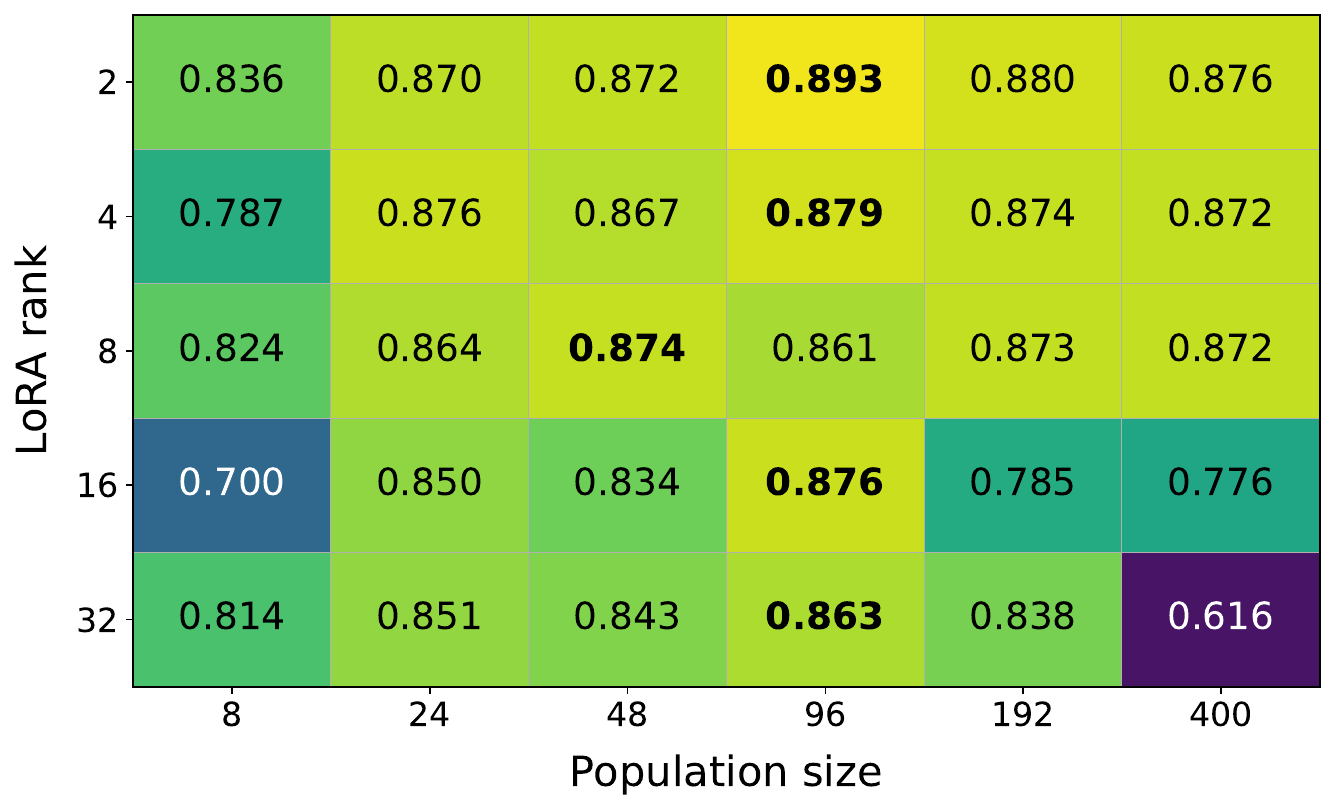}
        \caption{LoRA rank vs.\ population size}
    \end{subfigure}
    \hfill
    \begin{subfigure}{0.48\textwidth}
        \centering
        \includegraphics[width=\linewidth]
            {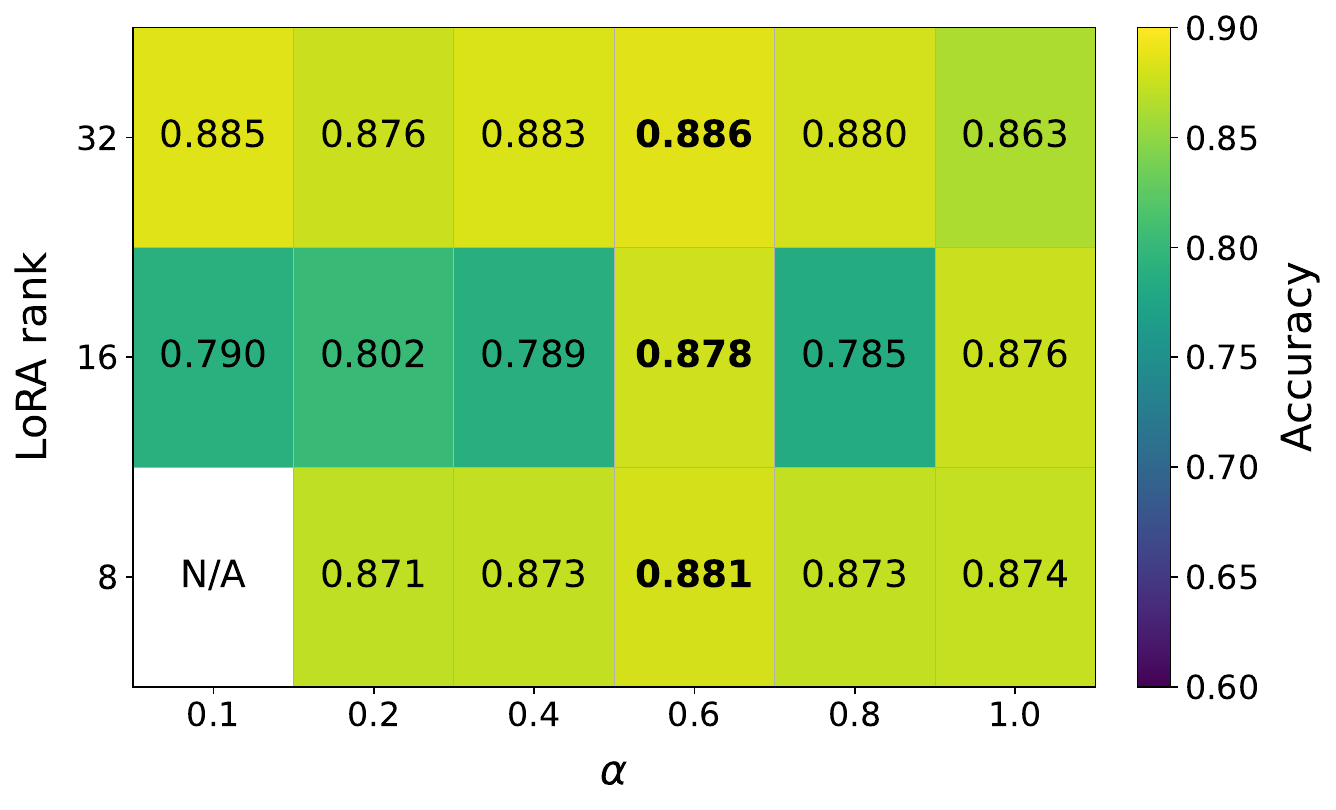}
        \caption{Fraction $\alpha$ at the best cell}
    \end{subfigure}
    \caption{Hyperparameter sensitivity of ESSA on \textbf{Qwen2.5-7B} for \textbf{GSM8K}. Batch
size 100.
\textbf{(a)} Accuracy when varying LoRA rank and population size.
\textbf{(b)} For each LoRA rank, the population size is fixed to the best value found in (a),
while the percentage $\alpha$ of trainable singular values is varied.
This illustrates how ESSA performance depends jointly on adapter rank
and the fraction of singular values optimized. The single white cell occurs because for LoRA rank~8 and $\alpha\!=\!0.1$,
rounding down yields zero trainable singular values, so no valid accuracy is reported.}
    \label{fig:essa_hypers_qwen_gsm}
\end{figure*}

\begin{figure*}[p]
    \centering
    \begin{subfigure}{0.48\textwidth}
        \centering
        \includegraphics[width=\linewidth]
            {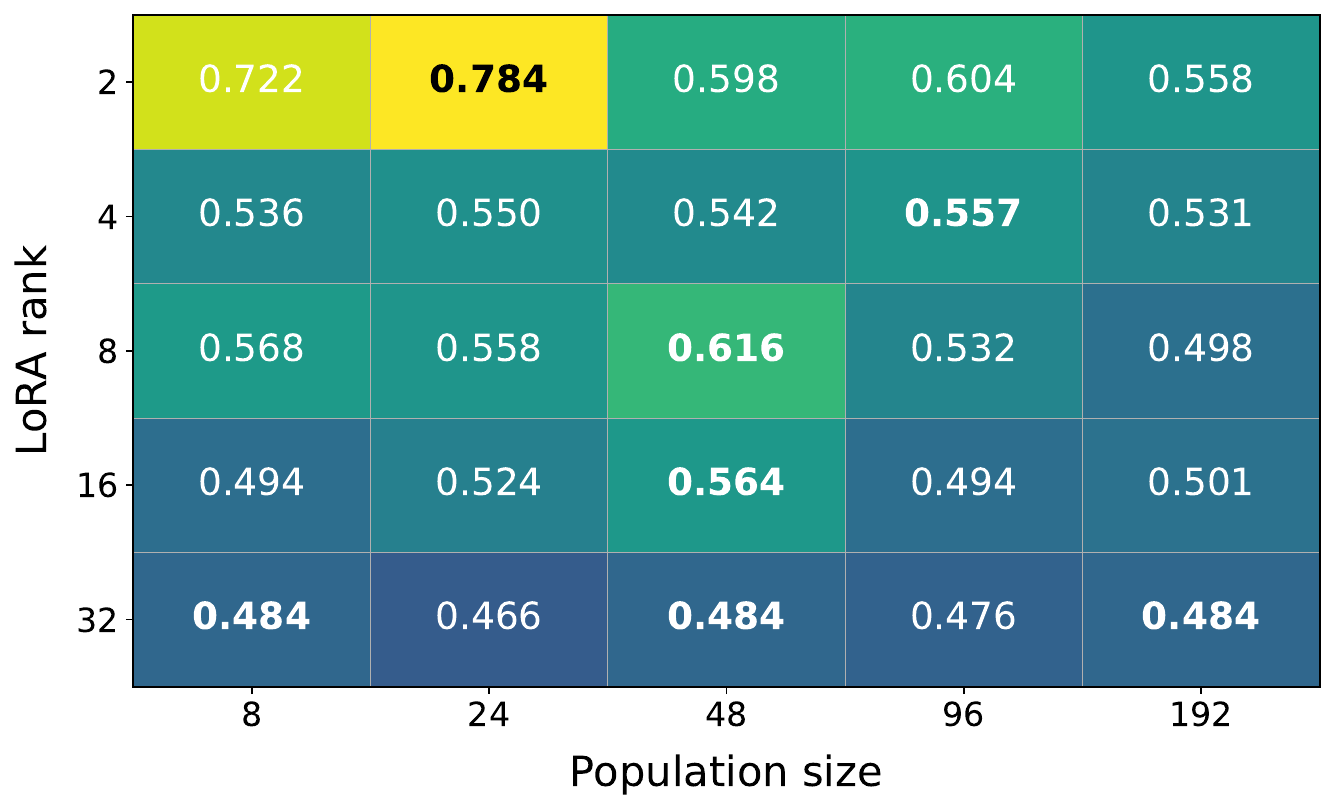}
        \caption{LoRA rank vs.\ population size}
    \end{subfigure}
    \hfill
    \begin{subfigure}{0.48\textwidth}
        \centering
        \includegraphics[width=\linewidth]
            {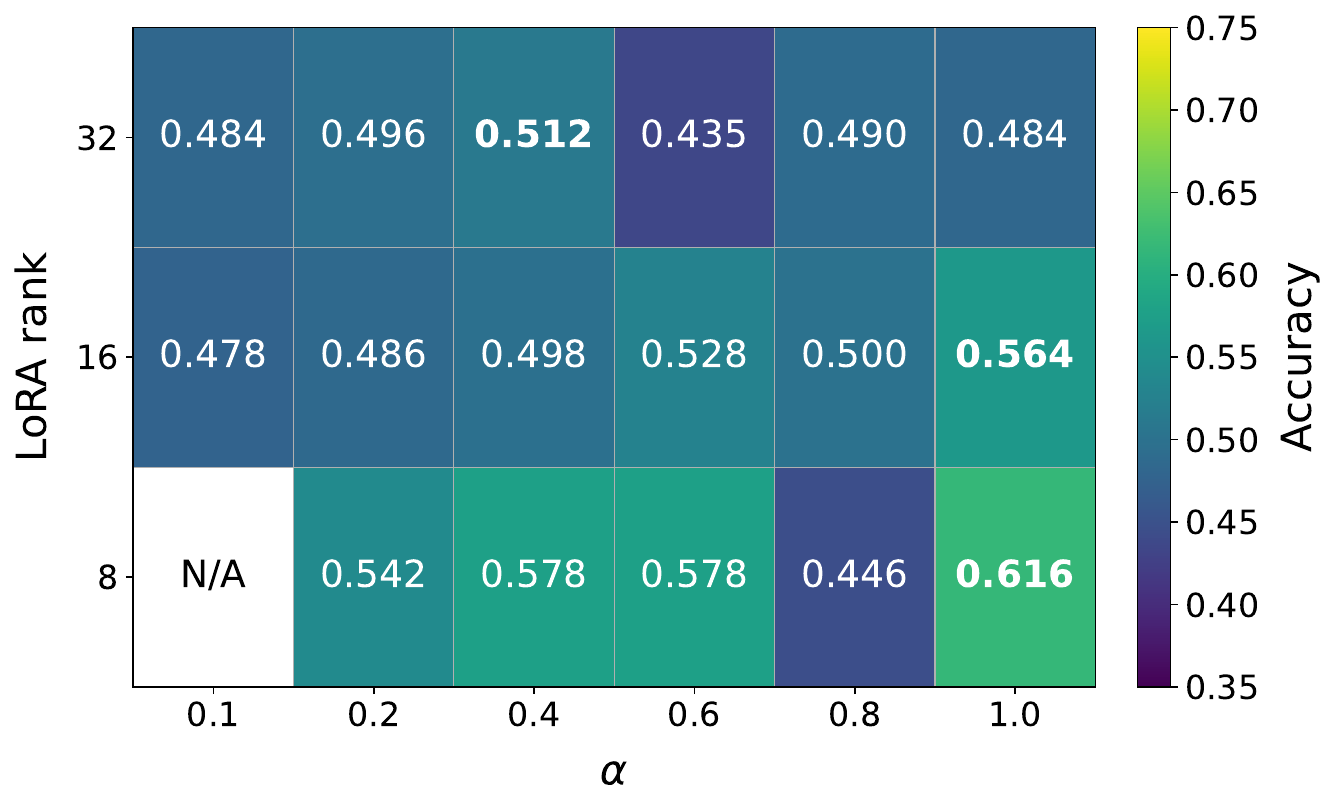}
        \caption{Fraction $\alpha$ at the best cell}
    \end{subfigure}
    \caption{Hyperparameter sensitivity of ESSA on \textbf{Qwen2.5-Math-7B} for \textbf{PRM800K}. Batch size 300.
\textbf{(a)} Accuracy when varying LoRA rank and population size.
\textbf{(b)} For each LoRA rank, the population size is fixed to the best value found in (a),
while the percentage $\alpha$ of trainable singular values is varied.
This illustrates how ESSA performance depends jointly on adapter rank
and the fraction of singular values optimized. The single white cell occurs because for LoRA rank~8 and $\alpha\!=\!0.1$,
rounding down yields zero trainable singular values, so no valid accuracy is reported.
    }
    \label{fig:essa_hypers_qwen_math_prm}
\end{figure*}

\clearpage

\begingroup
\makeatletter
\setlength{\@dblfptop}{0pt}
\setlength{\@dblfpsep}{16pt}
\setlength{\@dblfpbot}{0pt plus 1fil}
\makeatother

\begin{figure*}[p]
    \centering
    \begin{subfigure}{0.48\textwidth}
        \centering
        \includegraphics[width=\linewidth]
            {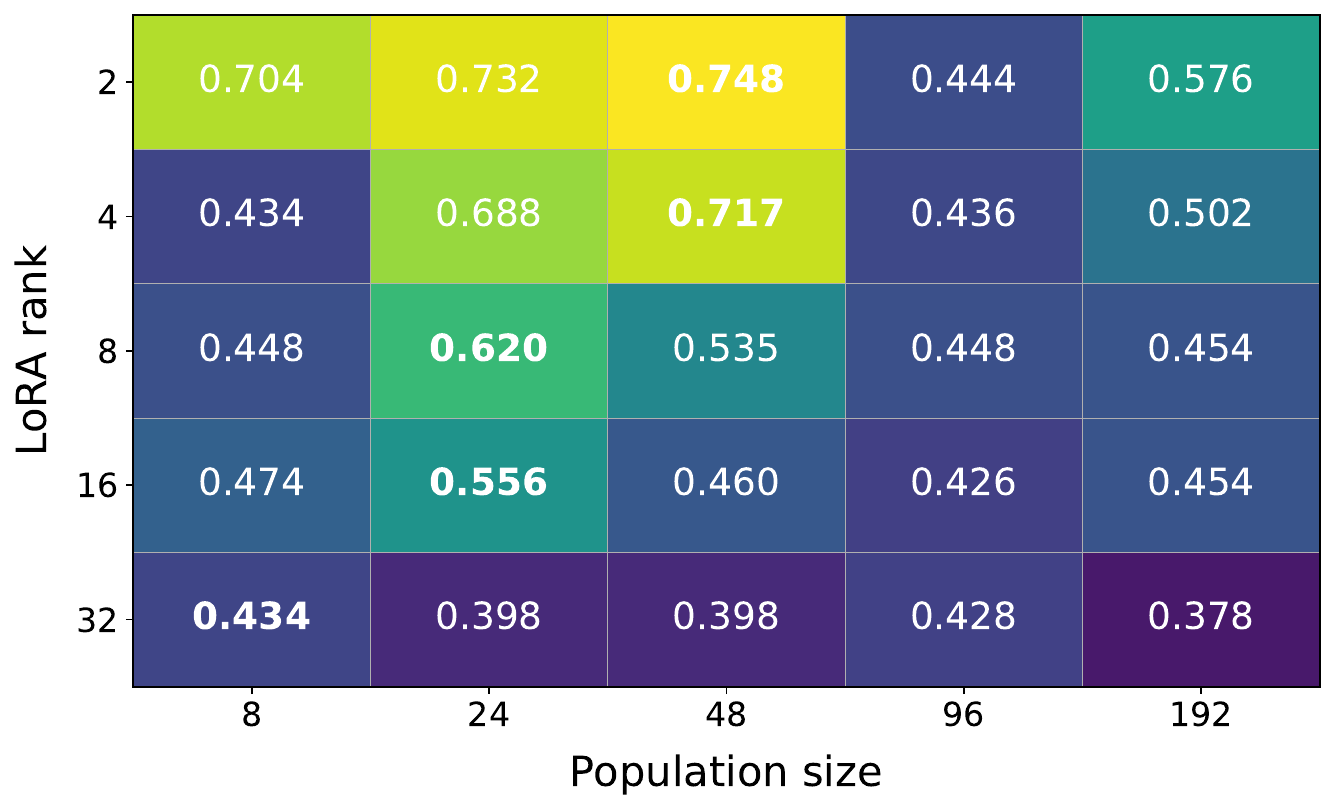}
        \caption{LoRA rank vs.\ population size}
    \end{subfigure}
    \hfill
    \begin{subfigure}{0.48\textwidth}
        \centering
        \includegraphics[width=\linewidth]
            {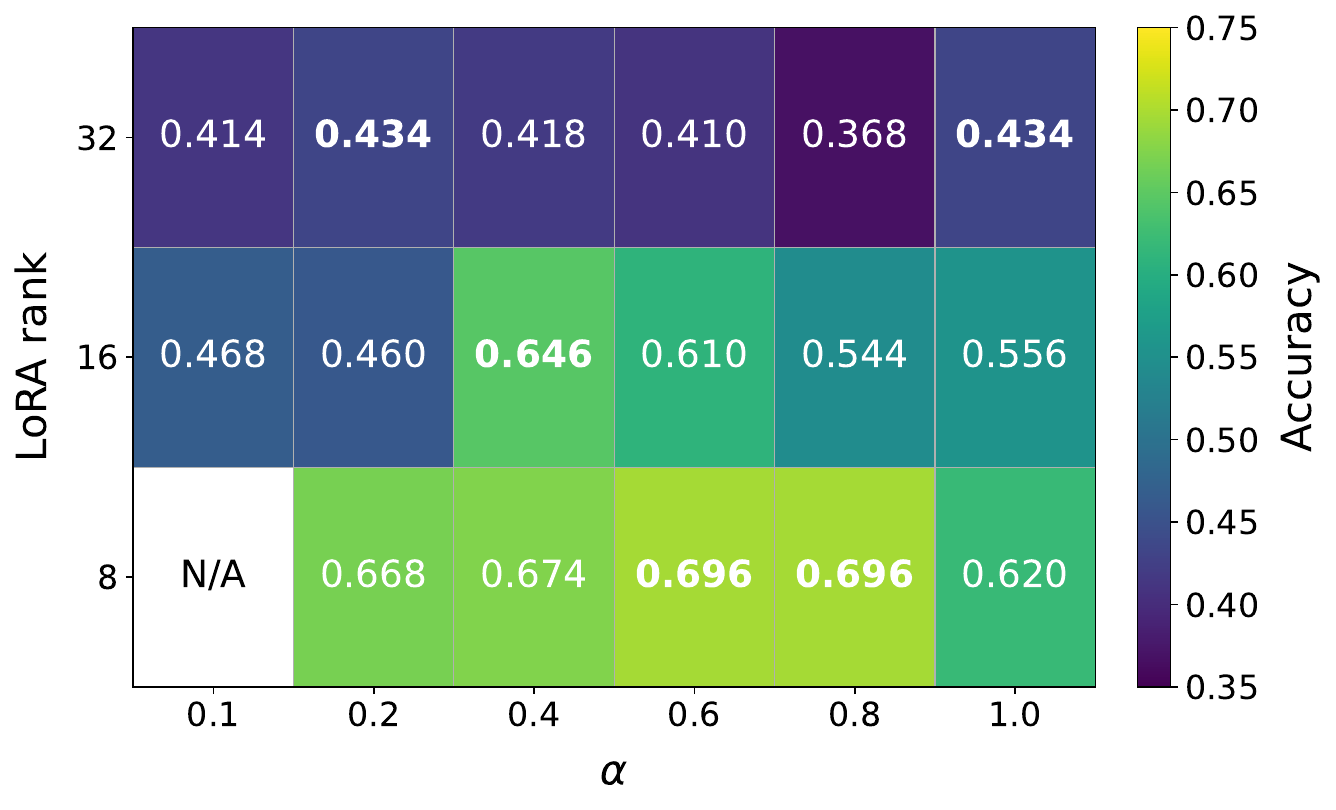}
        \caption{Fraction $\alpha$ at the best cell}
    \end{subfigure}
    \caption{Hyperparameter sensitivity of ESSA on \textbf{Qwen2.5-7B} for \textbf{PRM800K}. Batch
size 300.
\textbf{(a)} Accuracy when varying LoRA rank and population size.
\textbf{(b)} For each LoRA rank, the population size is fixed to the best value found in (a),
while the percentage $\alpha$ of trainable singular values is varied.
This illustrates how ESSA performance depends jointly on adapter rank
and the fraction of singular values optimized. The single white cell occurs because for LoRA rank~8 and $\alpha\!=\!0.1$,
rounding down yields zero trainable singular values, so no valid accuracy is reported.}
    \label{fig:essa_hypers_qwen_prm}
\end{figure*}

\begin{figure*}[p]
    \centering
    \includegraphics[width=0.5\linewidth]
        {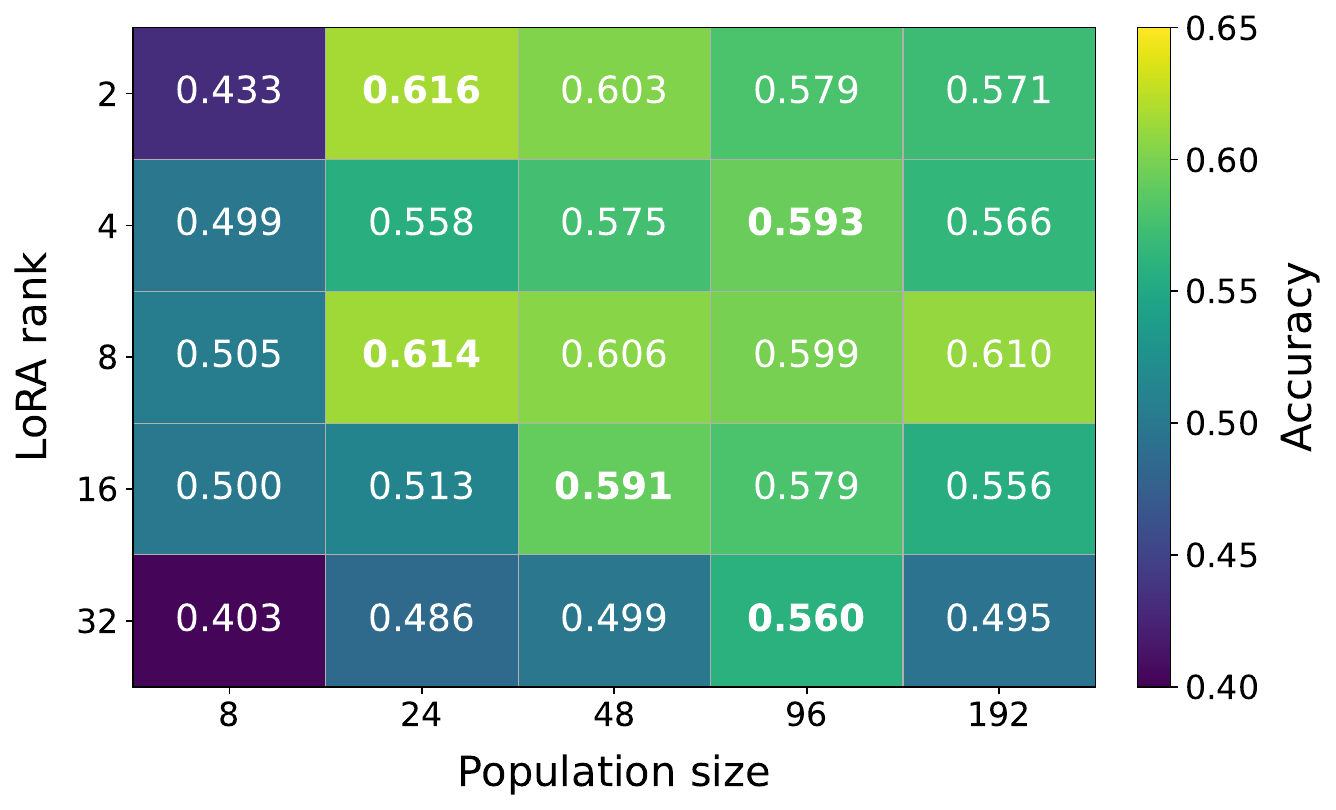}
    \caption{Hyperparameter sensitivity of ESSA on \textbf{Llama-3.1-8B} for \textbf{IFEval}. Accuracy when varying LoRA rank and population size.
    }
    \label{fig:essa_hypers_llama_ifeval}
\end{figure*}

\clearpage

\section{Training Dynamics}
\label{app:dynamics}

This section provides the full training-dynamics results supporting
the hyperparameter-sensitivity analysis in
Section~\ref{sec:sensitivity}. The GSM8K curves additionally
complement the LoRA-budget comparison in
Section~\ref{sec:main_comparison}; the PRM800K curves complement
the optimizer and subspace analysis in
Section~\ref{sec:optimizer_controls}; and the IFEval curves
complement the instruction-following results in
Section~\ref{sec:nlp}.

Figures~\ref{fig:train_qwen_math_gsm_loras}--
\ref{fig:train_llama_ifeval_pops} show how ESSA validation
accuracy evolves when varying the population size at fixed LoRA
ranks, the LoRA rank at fixed population sizes, and, where
applicable, the fraction $\alpha$ of singular-value coordinates
optimized. These curves complement the static sensitivity heatmaps
reported in Appendix~\ref{app:hyperparams}.

\subsubsection{Qwen2.5-Math-7B on GSM8K}

These experiments support the sensitivity analysis in
Section~\ref{sec:sensitivity} and extend the GSM8K results in
Section~\ref{sec:main_comparison} to Qwen2.5-Math-7B.
Figures~\ref{fig:train_qwen_math_gsm_loras},
\ref{fig:train_qwen_math_gsm_pops}, and
\ref{fig:train_qwen_math_gsm_alphas} present the complete
training-dynamics study corresponding to the static results in
Figure~\ref{fig:essa_hypers_qwen_math_gsm}.
Figure~\ref{fig:train_qwen_math_gsm_loras} varies the population
size at fixed LoRA ranks,
Figure~\ref{fig:train_qwen_math_gsm_pops} varies the LoRA rank at
fixed population sizes, and
Figure~\ref{fig:train_qwen_math_gsm_alphas} varies $\alpha$.

\subsubsection{Qwen2.5-7B on GSM8K}

These experiments provide the full sensitivity results for the
Qwen2.5-7B/GSM8K setting used in
Section~\ref{sec:main_comparison} and support the analysis in
Section~\ref{sec:sensitivity}.
Figures~\ref{fig:train_qwen_gsm_loras},
\ref{fig:train_qwen_gsm_pops}, and
\ref{fig:train_qwen_gsm_alphas} correspond to the static results
in Figure~\ref{fig:essa_hypers_qwen_gsm}.
Figure~\ref{fig:train_qwen_gsm_loras} varies the population size
at fixed LoRA ranks,
Figure~\ref{fig:train_qwen_gsm_pops} varies the LoRA rank at fixed
population sizes, and
Figure~\ref{fig:train_qwen_gsm_alphas} varies $\alpha$.

\subsubsection{Qwen2.5-Math-7B on PRM800K}

These experiments support the sensitivity analysis in
Section~\ref{sec:sensitivity} and complement the
PRM800K$\rightarrow$MATH500 analysis in
Section~\ref{sec:optimizer_controls}.
Figures~\ref{fig:train_qwen_math_prm_loras},
\ref{fig:train_qwen_math_prm_pops}, and
\ref{fig:train_qwen_math_prm_alphas} present the complete
training-dynamics study corresponding to the static results in
Figure~\ref{fig:essa_hypers_qwen_math_prm}.
Figure~\ref{fig:train_qwen_math_prm_loras} varies the population
size at fixed LoRA ranks,
Figure~\ref{fig:train_qwen_math_prm_pops} varies the LoRA rank at
fixed population sizes, and
Figure~\ref{fig:train_qwen_math_prm_alphas} varies $\alpha$.

\subsubsection{Qwen2.5-7B on PRM800K}

These experiments provide additional training dynamics for the
Qwen2.5-7B/PRM800K setting analyzed in
Section~\ref{sec:optimizer_controls} and support the sensitivity
analysis in Section~\ref{sec:sensitivity}.
Figures~\ref{fig:train_qwen_prm_loras},
\ref{fig:train_qwen_prm_pops}, and
\ref{fig:train_qwen_prm_alphas} correspond to the static results
in Figure~\ref{fig:essa_hypers_qwen_prm}.
Figure~\ref{fig:train_qwen_prm_loras} varies the population size
at fixed LoRA ranks,
Figure~\ref{fig:train_qwen_prm_pops} varies the LoRA rank at fixed
population sizes, and
Figure~\ref{fig:train_qwen_prm_alphas} varies $\alpha$.

\subsubsection{Llama-3.1-8B on IFEval}

These experiments provide the full sensitivity results for the
Llama-3.1-8B/IFEval setting used in the LoRA-budget comparison in
Section~\ref{sec:main_comparison} and the instruction-following
analysis in Section~\ref{sec:nlp}. They also support the
hyperparameter analysis in Section~\ref{sec:sensitivity}.
Figures~\ref{fig:train_llama_ifeval_loras} and
\ref{fig:train_llama_ifeval_pops} correspond to the static results
in Figure~\ref{fig:essa_hypers_llama_ifeval}.
Figure~\ref{fig:train_llama_ifeval_loras} varies the population
size at fixed LoRA ranks, while
Figure~\ref{fig:train_llama_ifeval_pops} varies the LoRA rank at
fixed population sizes.

\clearpage
\begingroup
\makeatletter
\setlength{\@dblfptop}{0pt}
\setlength{\@dblfpsep}{56pt}
\setlength{\@dblfpbot}{0pt plus 1fil}
\makeatother

\begin{figure*}[p]
  \centering
  \begin{subfigure}{0.32\textwidth}\includegraphics[width=\linewidth]{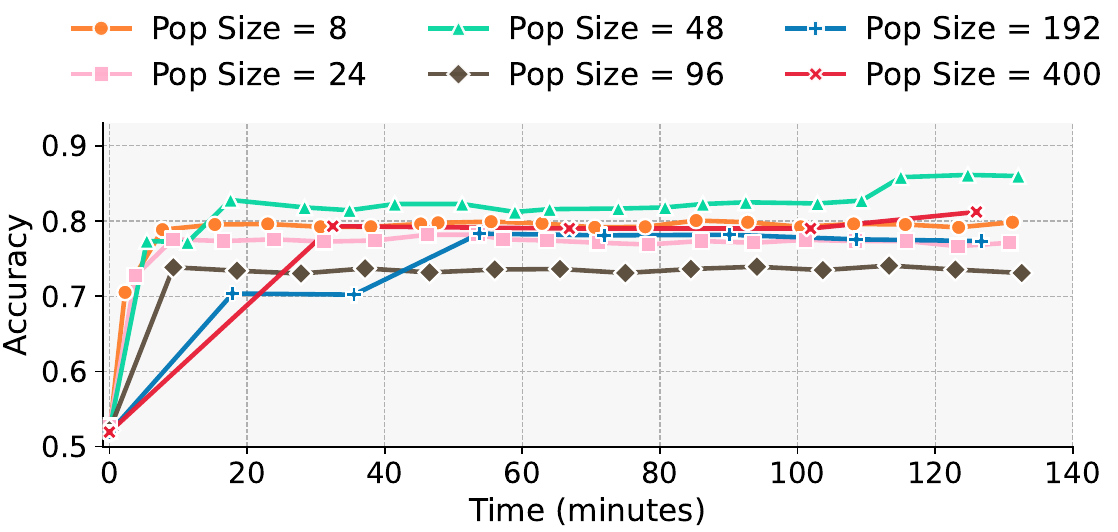}\caption{LoRA rank 2}\end{subfigure}\hfill
  \begin{subfigure}{0.32\textwidth}\includegraphics[width=\linewidth]{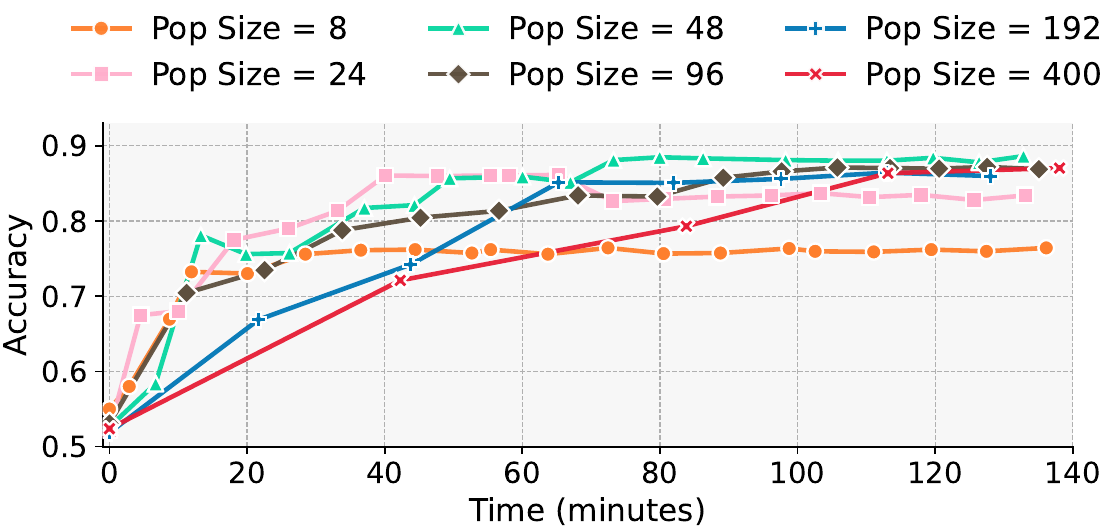}\caption{LoRA rank 4}\end{subfigure}\hfill
  \begin{subfigure}{0.32\textwidth}\includegraphics[width=\linewidth]{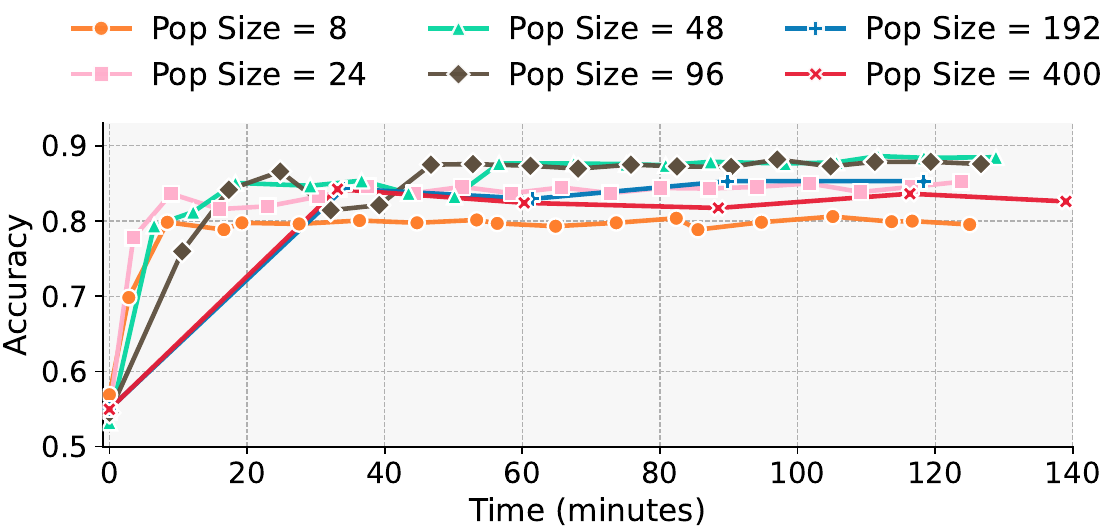}\caption{LoRA rank 8}\end{subfigure}\\[6pt]
  \hspace*{\fill}%
  \begin{subfigure}{0.32\textwidth}\includegraphics[width=\linewidth]{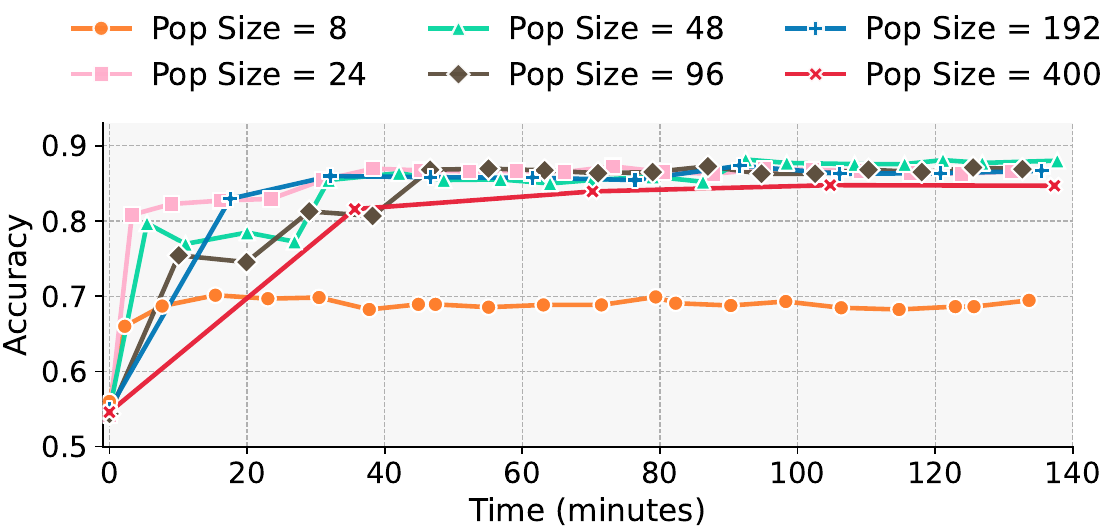}\caption{LoRA rank 16}\end{subfigure}\hfill
  \begin{subfigure}{0.32\textwidth}\includegraphics[width=\linewidth]{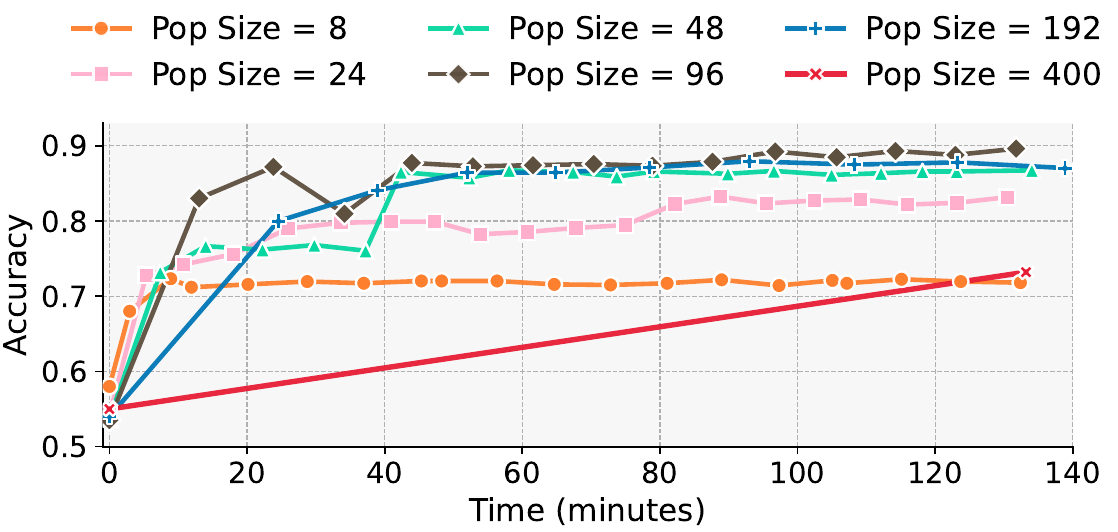}\caption{LoRA rank 32}\end{subfigure}%
  \hspace*{\fill}
  \caption{Validation accuracy over time on \textbf{GSM8K} with \textbf{Qwen2.5-Math-7B}
    when varying the \textsc{ESSA} population size
    (8, 24, 48, 96, 192, 400) at fixed LoRA ranks
    (2, 4, 8, 16, 32). Batch size 100.}
  \label{fig:train_qwen_math_gsm_loras}
\end{figure*}

\begin{figure*}[p]
  \centering
  \begin{subfigure}{0.32\textwidth}\includegraphics[width=\linewidth]{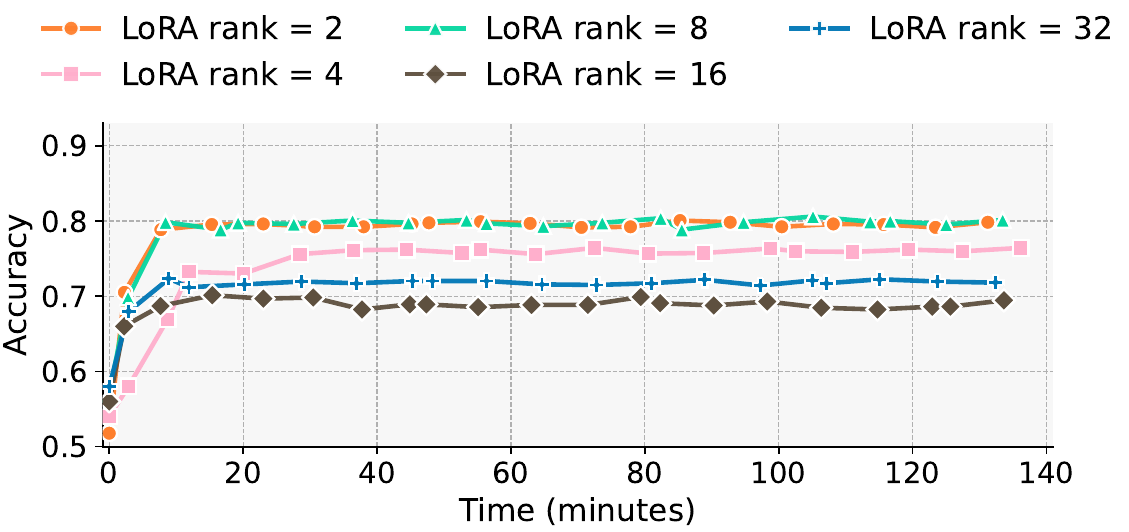}\caption{Population size 8}\end{subfigure}\hfill
  \begin{subfigure}{0.32\textwidth}\includegraphics[width=\linewidth]{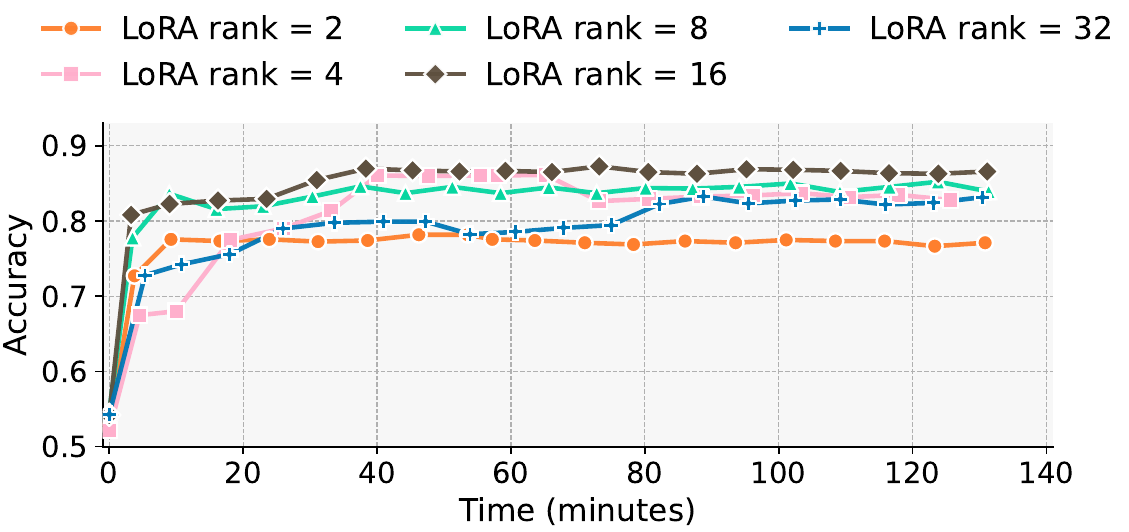}\caption{Population size 24}\end{subfigure}\hfill
  \begin{subfigure}{0.32\textwidth}\includegraphics[width=\linewidth]{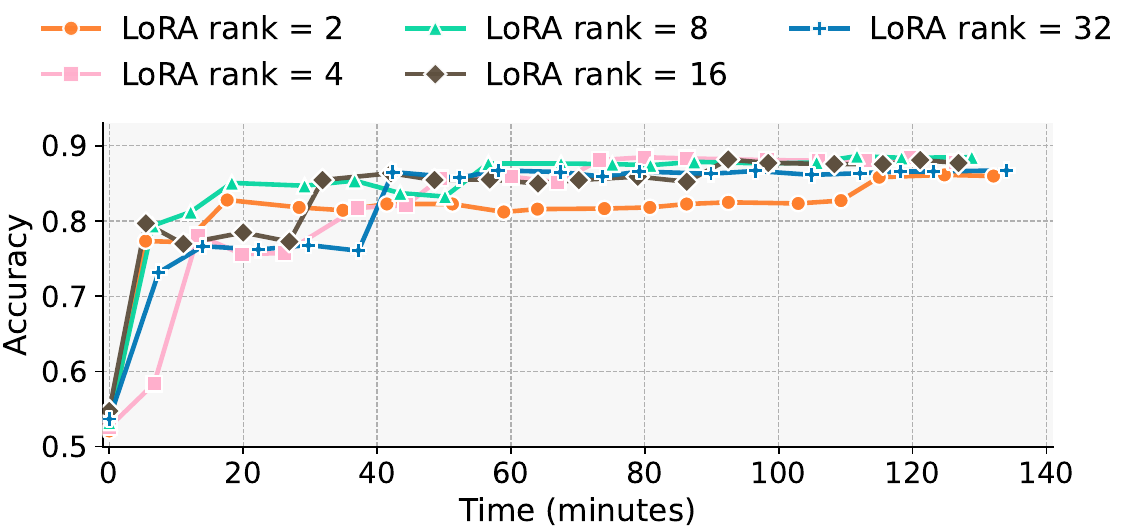}\caption{Population size 48}\end{subfigure}\\[6pt]
  \begin{subfigure}{0.32\textwidth}\includegraphics[width=\linewidth]{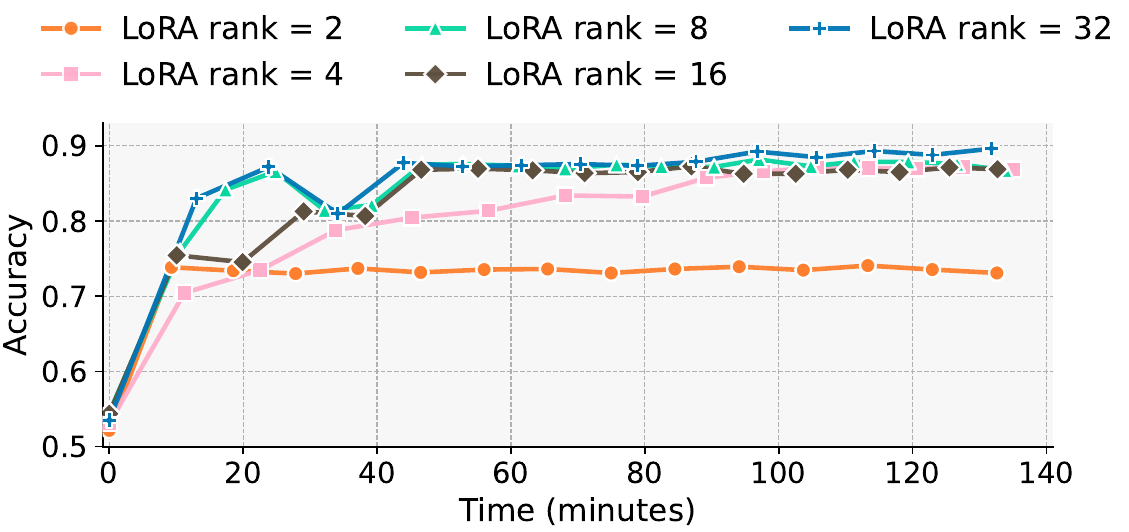}\caption{Population size 96}\end{subfigure}\hfill
  \begin{subfigure}{0.32\textwidth}\includegraphics[width=\linewidth]{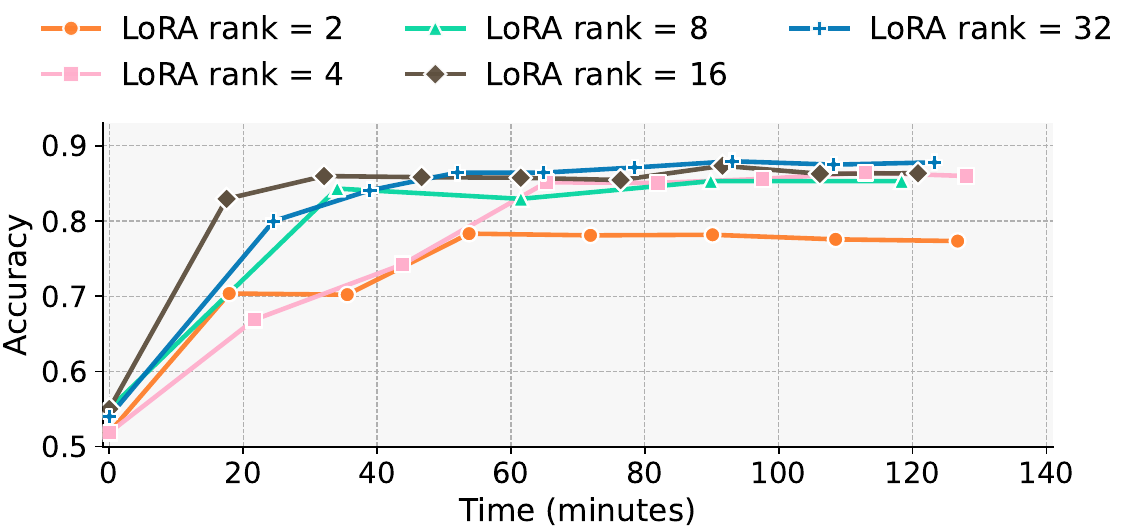}\caption{Population size 192}\end{subfigure}\hfill
  \begin{subfigure}{0.32\textwidth}\includegraphics[width=\linewidth]{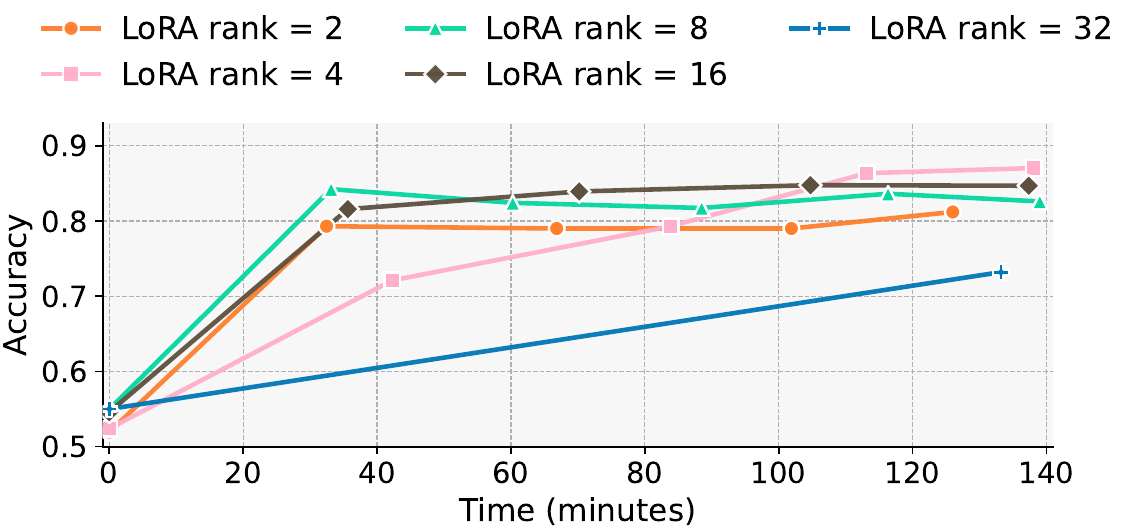}\caption{Population size 400}\end{subfigure}
  \caption{Validation accuracy over time on \textbf{GSM8K} with \textbf{Qwen2.5-Math-7B}
    when varying the LoRA rank (2, 4, 8, 16, 32)
    at fixed population sizes
    (8, 24, 48, 96, 192, 400). Batch size 100.}
  \label{fig:train_qwen_math_gsm_pops}
\end{figure*}

\begin{figure*}[p]
  \centering
  \begin{subfigure}{0.32\textwidth}\includegraphics[width=\linewidth]{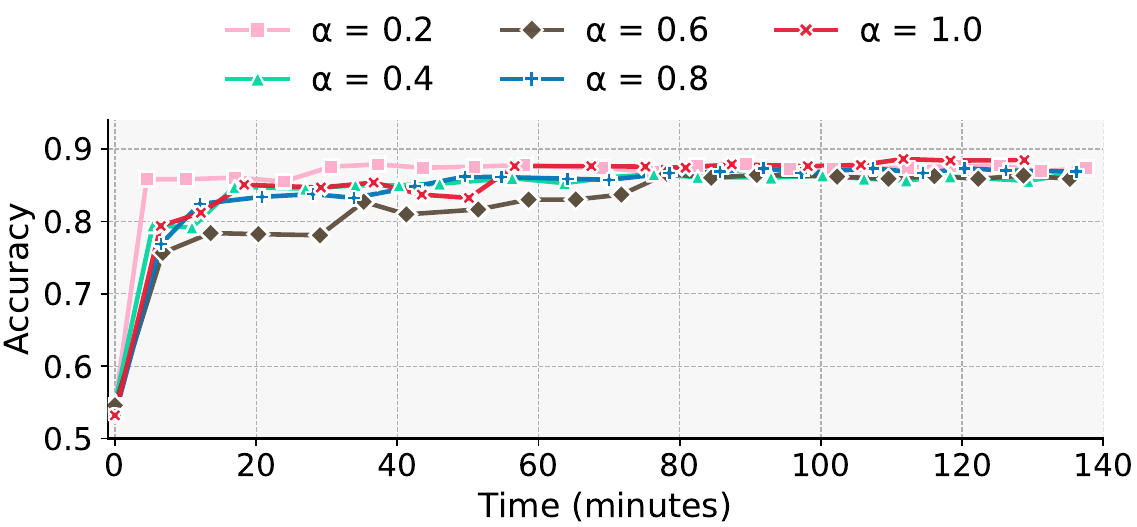}\caption{LoRA rank 8, Population size 48}\end{subfigure}\hfill
  \begin{subfigure}{0.32\textwidth}\includegraphics[width=\linewidth]{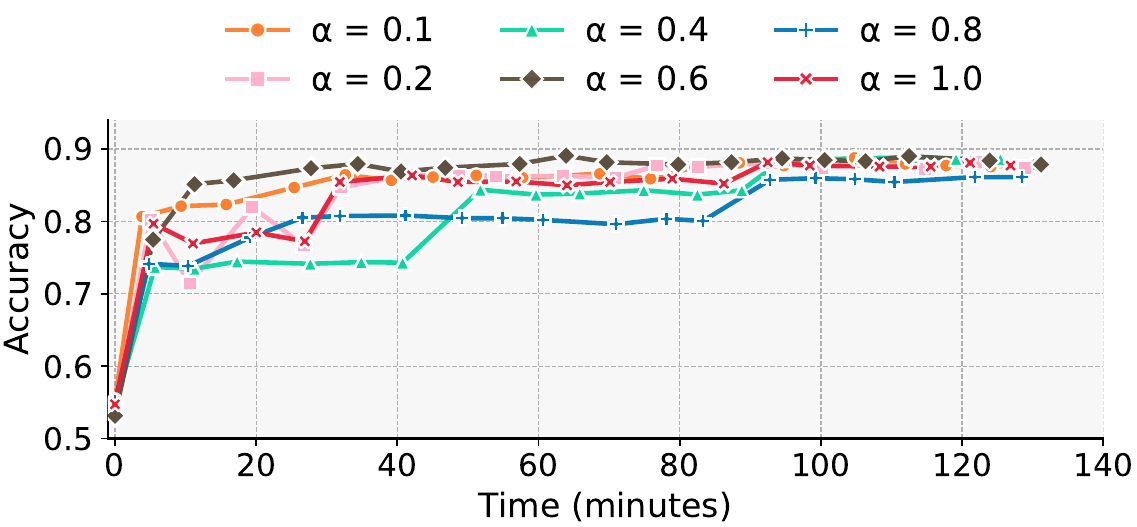}\caption{LoRA rank 16, Population size 48}\end{subfigure}\hfill
  \begin{subfigure}{0.32\textwidth}\includegraphics[width=\linewidth]{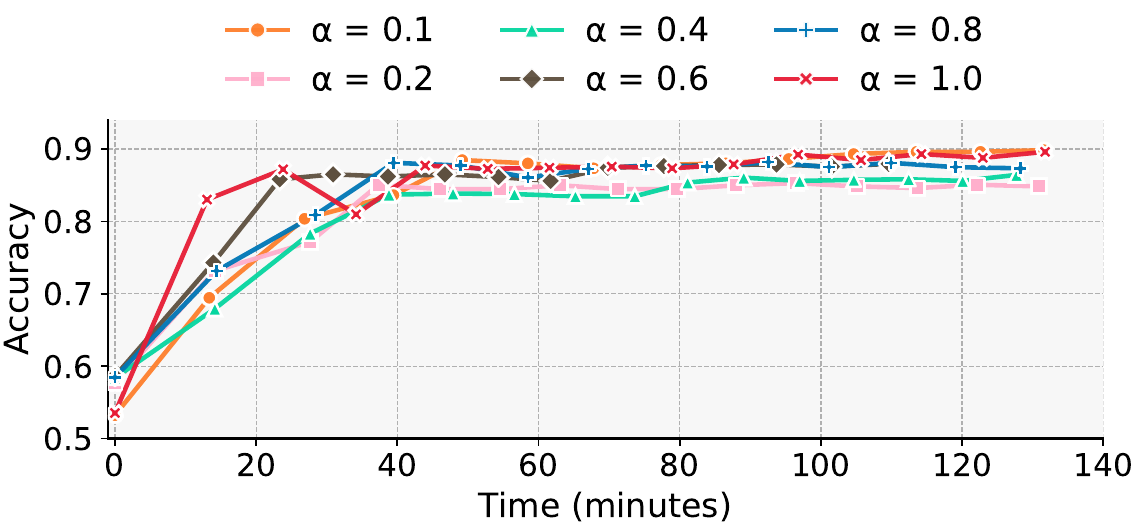}\caption{LoRA rank 32, Population size 96}\end{subfigure}
  \caption{Validation accuracy on \textbf{GSM8K} with
    \textbf{Qwen2.5-Math-7B} while varying $\alpha$.
    LoRA rank and population size are fixed to the optimal choices
    from Figure~\ref{fig:essa_hypers_qwen_math_gsm}. Batch size 100.}
  \label{fig:train_qwen_math_gsm_alphas}
\end{figure*}

\begin{figure*}[p]
  \centering
  \begin{subfigure}{0.32\textwidth}\includegraphics[width=\linewidth]{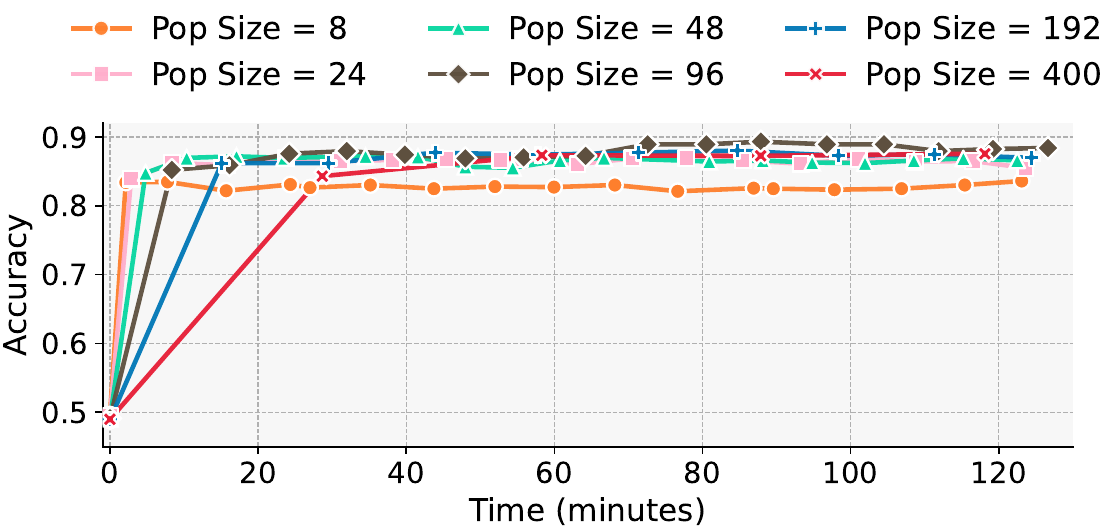}\caption{LoRA rank 2}\end{subfigure}\hfill
  \begin{subfigure}{0.32\textwidth}\includegraphics[width=\linewidth]{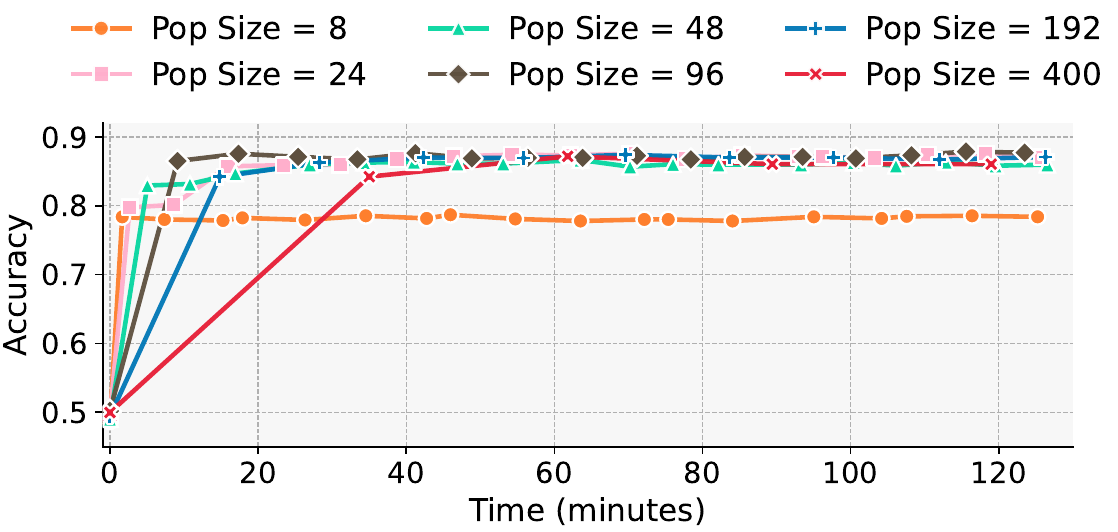}\caption{LoRA rank 4}\end{subfigure}\hfill
  \begin{subfigure}{0.32\textwidth}\includegraphics[width=\linewidth]{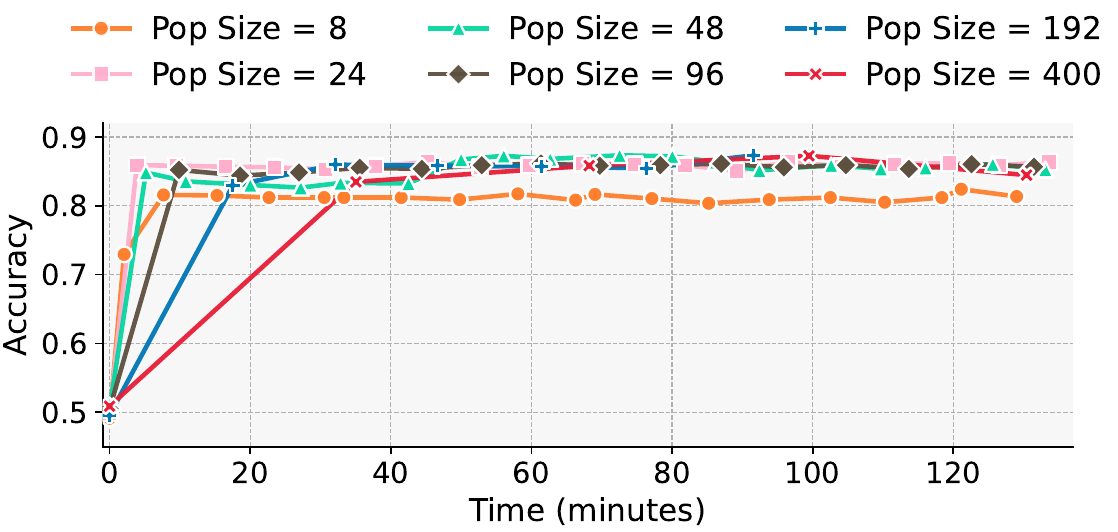}\caption{LoRA rank 8}\end{subfigure}\\[6pt]
  \hspace*{\fill}%
  \begin{subfigure}{0.32\textwidth}\includegraphics[width=\linewidth]{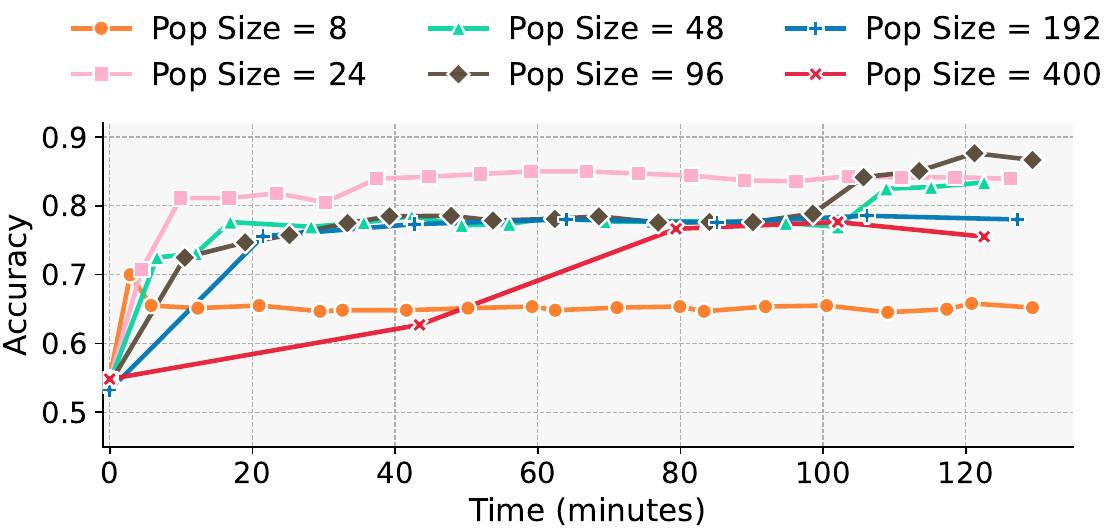}\caption{LoRA rank 16}\end{subfigure}\hfill
  \begin{subfigure}{0.32\textwidth}\includegraphics[width=\linewidth]{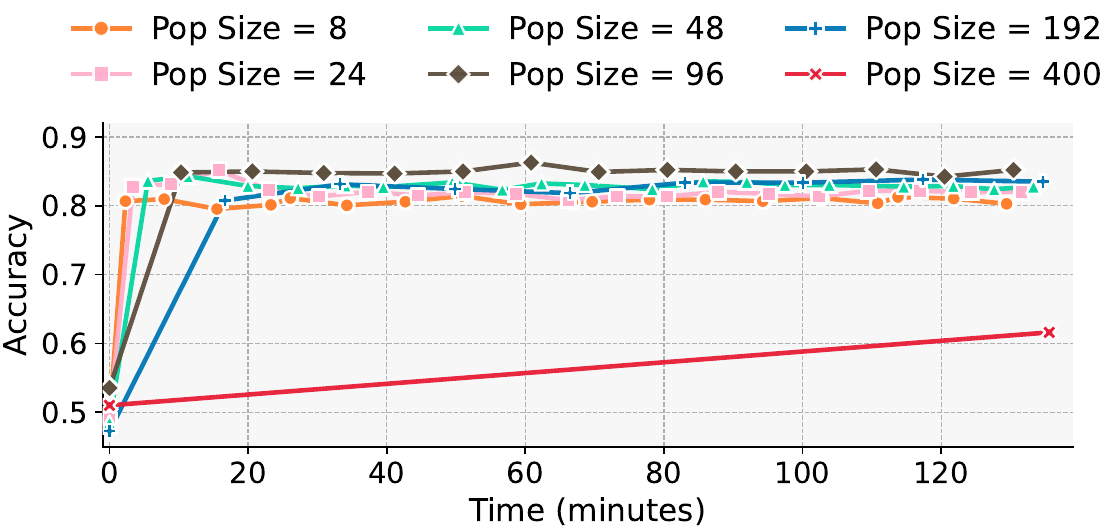}\caption{LoRA rank 32}\end{subfigure}%
  \hspace*{\fill}
  \caption{Validation accuracy over time on \textbf{GSM8K} with \textbf{Qwen2.5-7B}
    when varying the \textsc{ESSA} population size
    (8, 24, 48, 96, 192, 400) at fixed LoRA ranks
    (2, 4, 8, 16, 32). Batch size 100.}
  \label{fig:train_qwen_gsm_loras}
\end{figure*}

\begin{figure*}[p]
  \centering
  \begin{subfigure}{0.32\textwidth}\includegraphics[width=\linewidth]{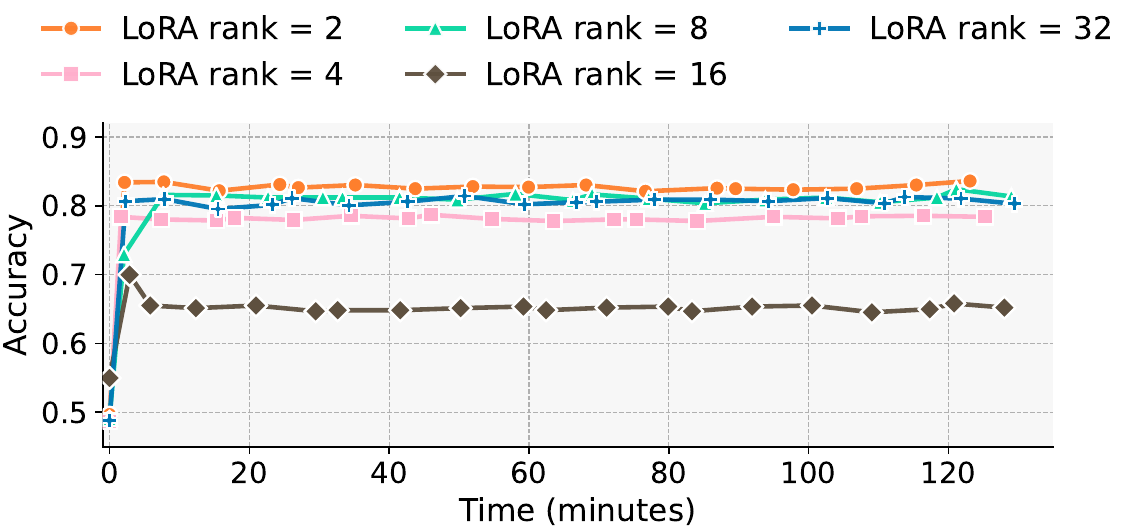}\caption{Population size 8}\end{subfigure}\hfill
  \begin{subfigure}{0.32\textwidth}\includegraphics[width=\linewidth]{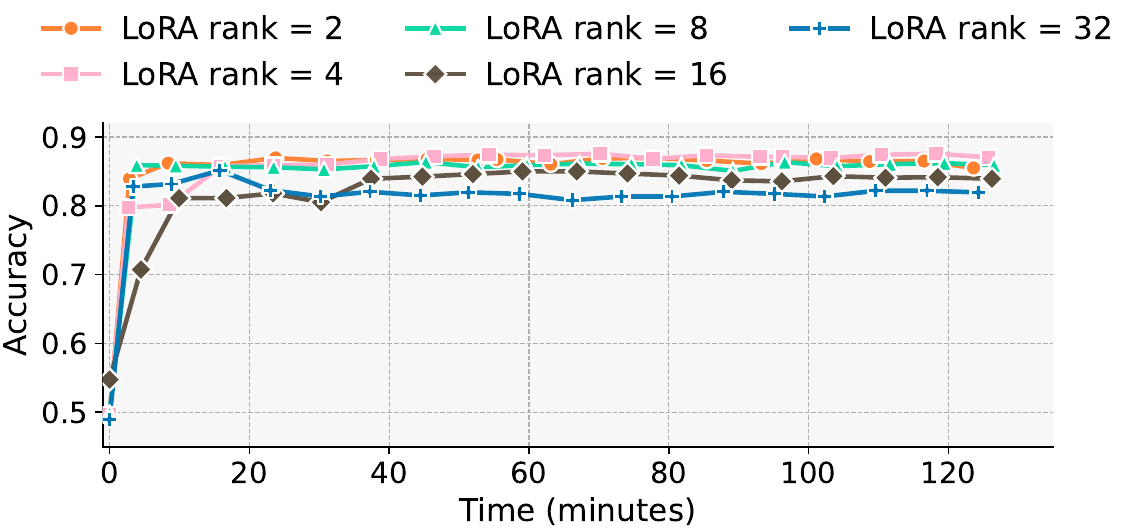}\caption{Population size 24}\end{subfigure}\hfill
  \begin{subfigure}{0.32\textwidth}\includegraphics[width=\linewidth]{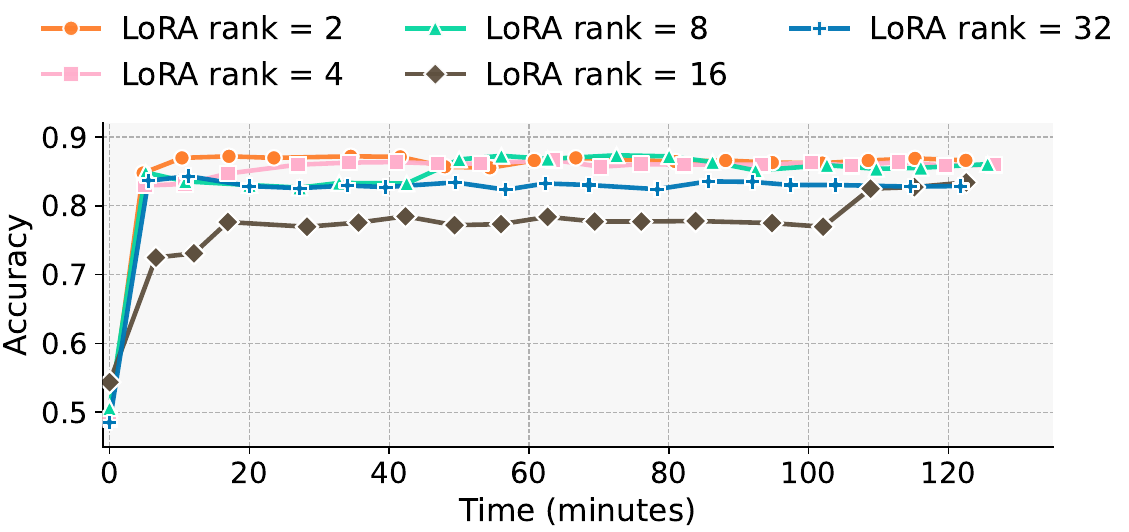}\caption{Population size 48}\end{subfigure}\\[6pt]
  \begin{subfigure}{0.32\textwidth}\includegraphics[width=\linewidth]{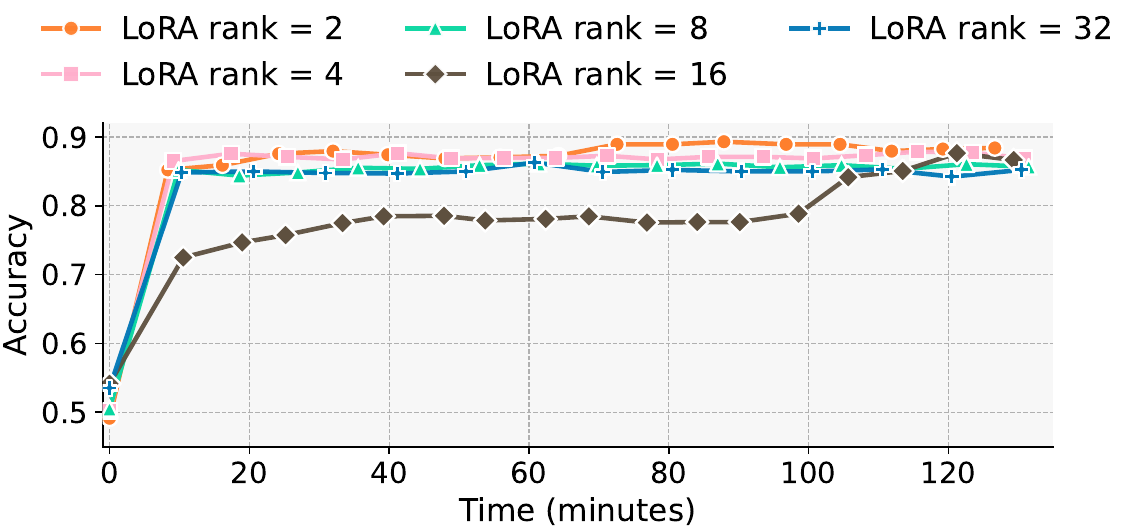}\caption{Population size 96}\end{subfigure}\hfill
  \begin{subfigure}{0.32\textwidth}\includegraphics[width=\linewidth]{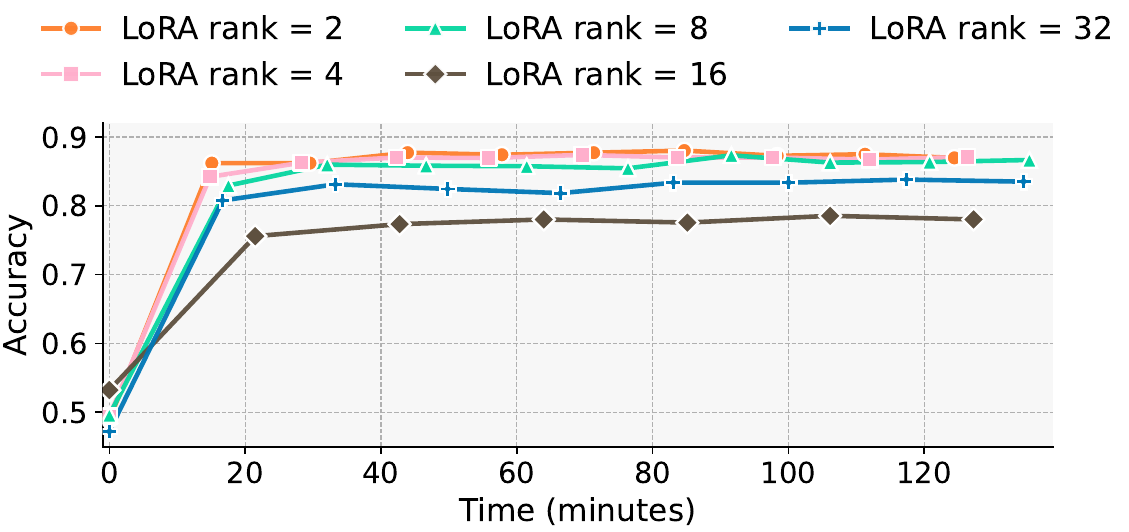}\caption{Population size 192}\end{subfigure}\hfill
  \begin{subfigure}{0.32\textwidth}\includegraphics[width=\linewidth]{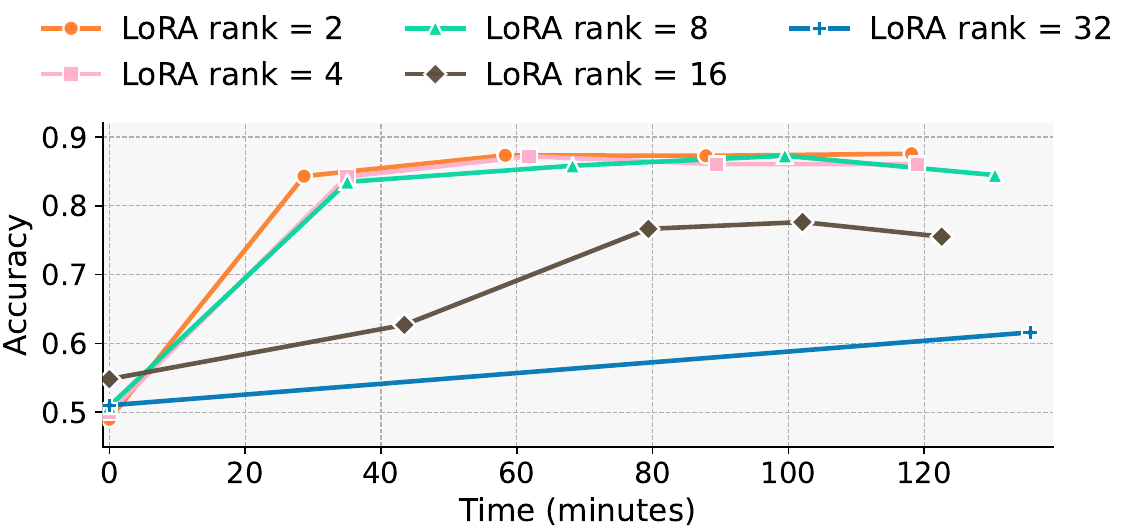}\caption{Population size 400}\end{subfigure}
  \caption{Validation accuracy over time on \textbf{GSM8K} with \textbf{Qwen2.5-7B}
    when varying the LoRA rank (2, 4, 8, 16, 32)
    at fixed population sizes
    (8, 24, 48, 96, 192, 400). Batch size 100.}
  \label{fig:train_qwen_gsm_pops}
\end{figure*}

\begin{figure*}[p]
  \centering
  \begin{subfigure}{0.32\textwidth}\includegraphics[width=\linewidth]{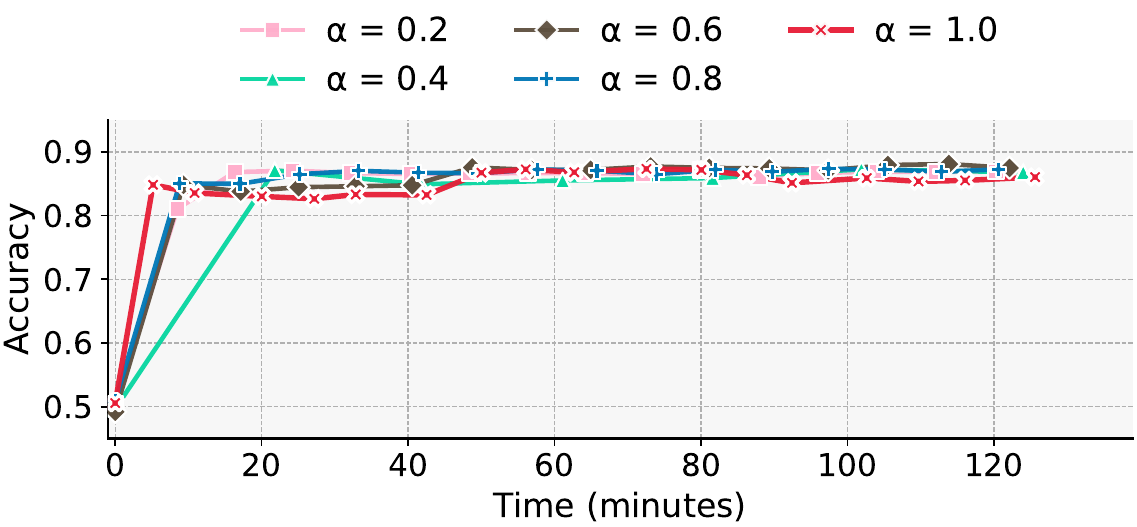}\caption{LoRA rank 8, Population size 48}\end{subfigure}\hfill
  \begin{subfigure}{0.32\textwidth}\includegraphics[width=\linewidth]{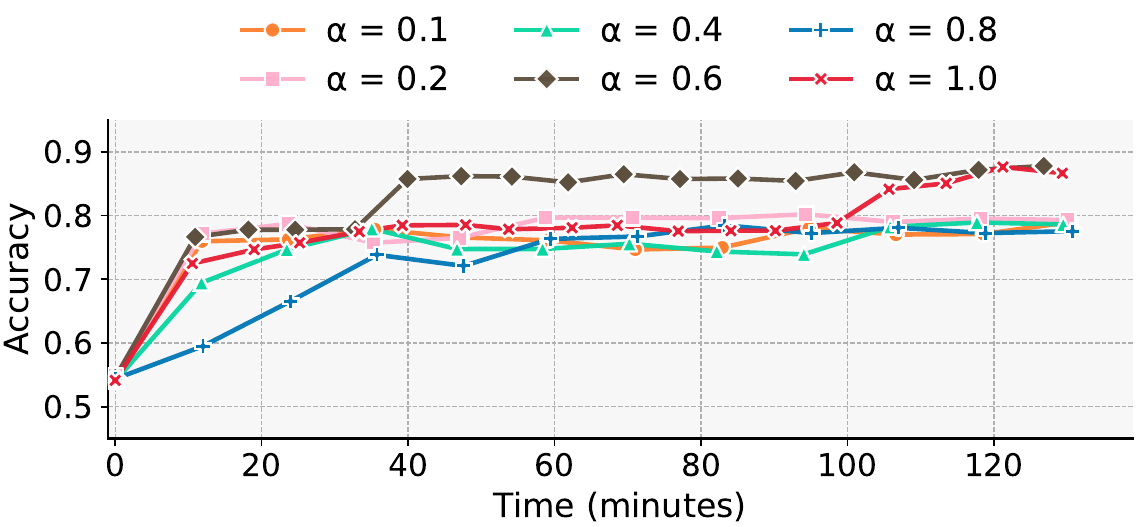}\caption{LoRA rank 16, Population size 96}\end{subfigure}\hfill
  \begin{subfigure}{0.32\textwidth}\includegraphics[width=\linewidth]{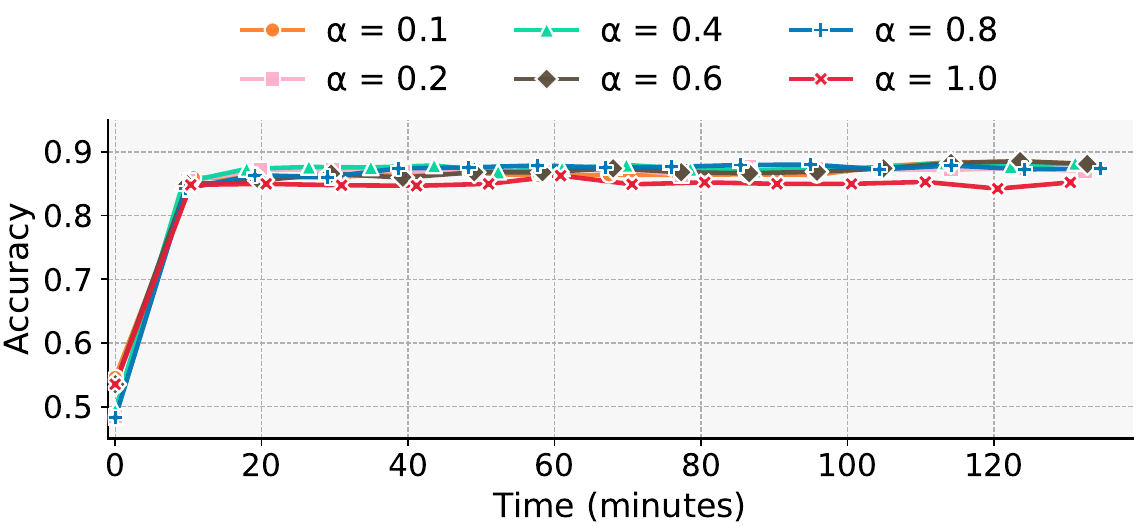}\caption{LoRA rank 32, Population size 96}\end{subfigure}
  \caption{Validation accuracy on \textbf{GSM8K} with \textbf{Qwen2.5-7B}
    while varying $\alpha$.
    LoRA rank and population size are fixed to the optimal choices from
    Figure~\ref{fig:essa_hypers_qwen_gsm}. Batch size 100.}
  \label{fig:train_qwen_gsm_alphas}
\end{figure*}

\begin{figure*}[p]
  \centering
  \begin{subfigure}{0.32\textwidth}\includegraphics[width=\linewidth]{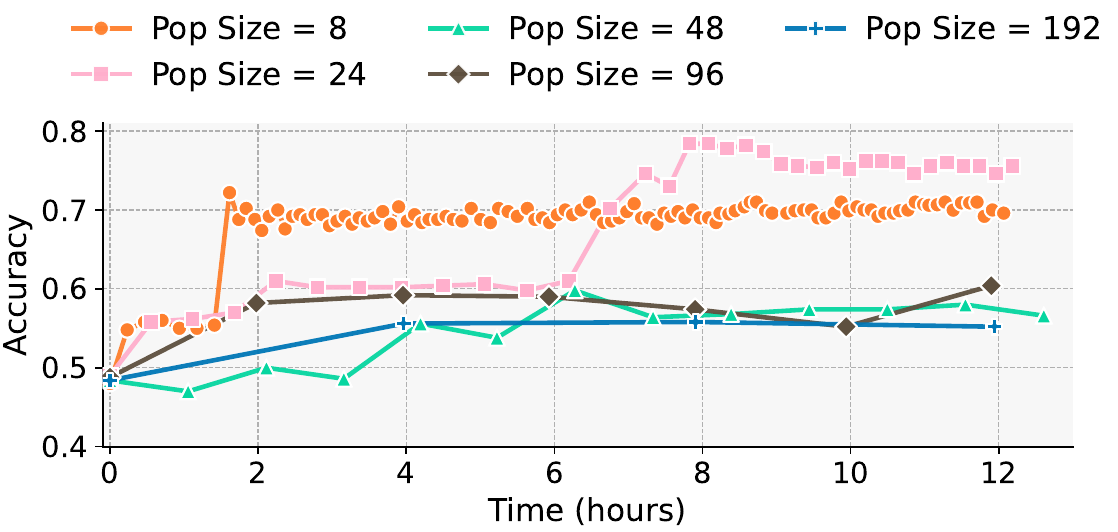}\caption{LoRA rank 2}\end{subfigure}\hfill
  \begin{subfigure}{0.32\textwidth}\includegraphics[width=\linewidth]{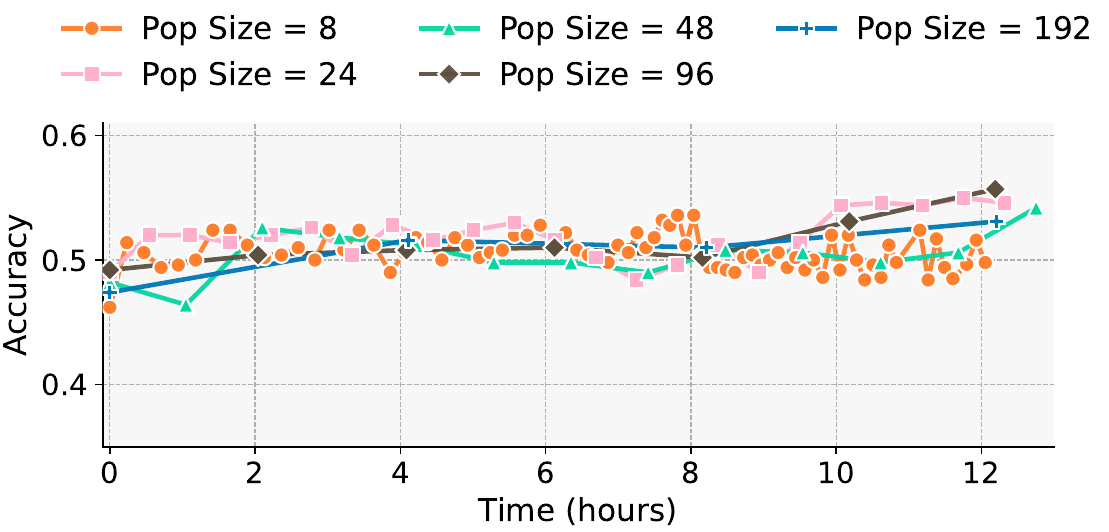}\caption{LoRA rank 4}\end{subfigure}\hfill
  \begin{subfigure}{0.32\textwidth}\includegraphics[width=\linewidth]{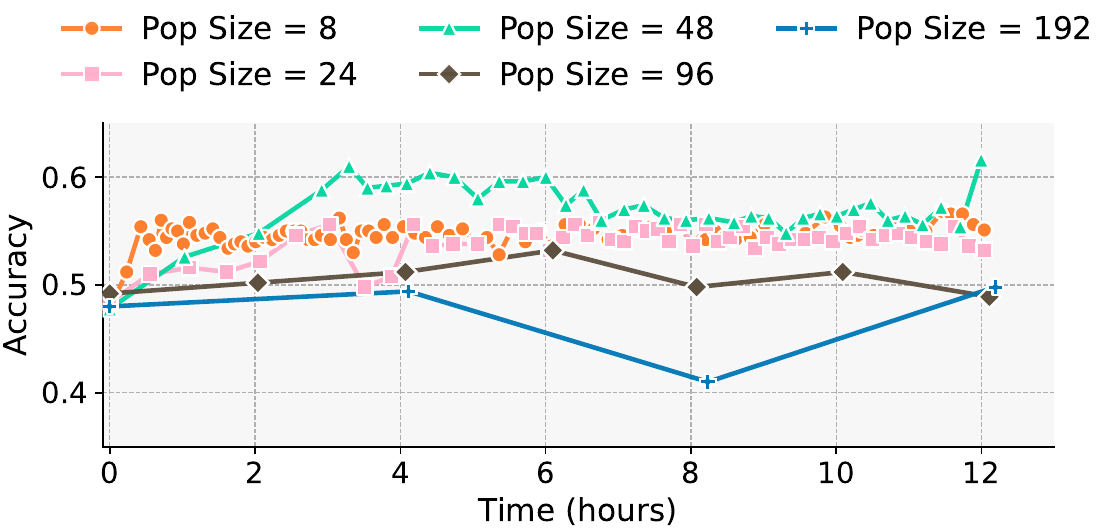}\caption{LoRA rank 8}\end{subfigure}\\[6pt]
  \hspace*{\fill}%
  \begin{subfigure}{0.32\textwidth}\includegraphics[width=\linewidth]{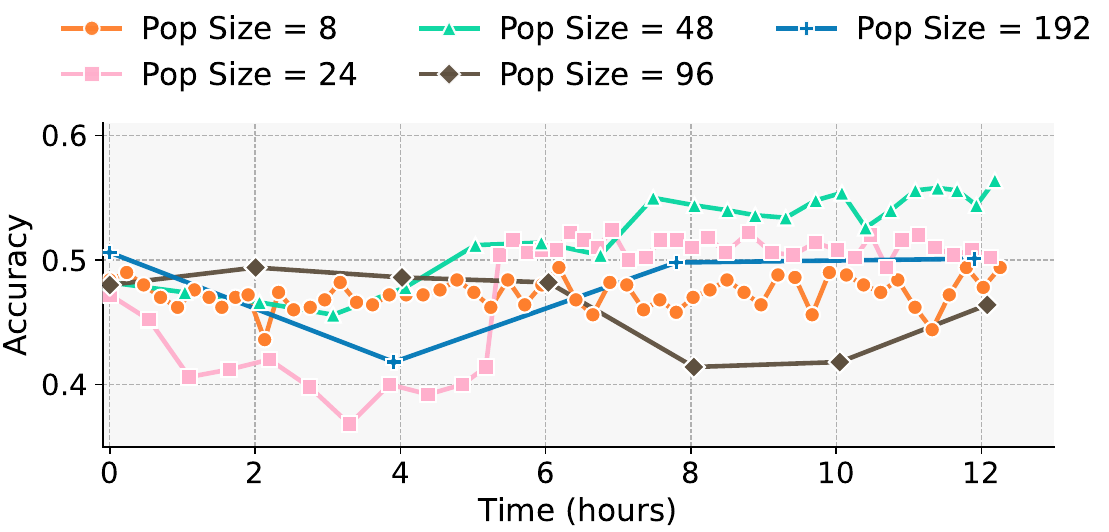}\caption{LoRA rank 16}\end{subfigure}\hfill
  \begin{subfigure}{0.32\textwidth}\includegraphics[width=\linewidth]{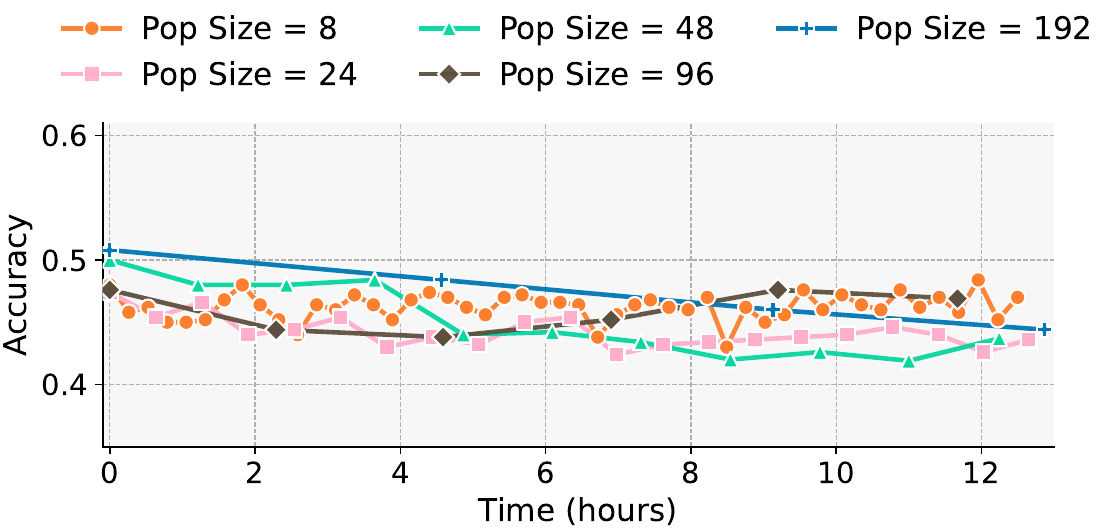}\caption{LoRA rank 32}\end{subfigure}%
  \hspace*{\fill}
  \caption{Validation accuracy over time on \textbf{PRM800K} with \textbf{Qwen2.5-Math-7B}
    when varying the \textsc{ESSA} population size
    (8, 24, 48, 96, 192) at fixed LoRA ranks
    (2, 4, 8, 16, 32). Batch size 300.}
  \label{fig:train_qwen_math_prm_loras}
\end{figure*}

\begin{figure*}[p]
  \centering
  \begin{subfigure}{0.32\textwidth}\includegraphics[width=\linewidth]{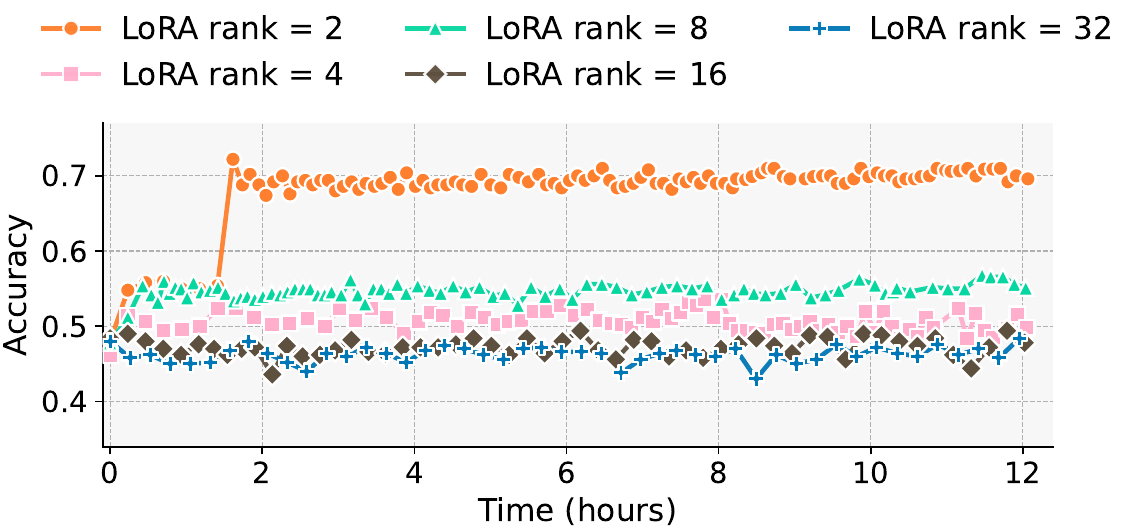}\caption{Population size 8}\end{subfigure}\hfill
  \begin{subfigure}{0.32\textwidth}\includegraphics[width=\linewidth]{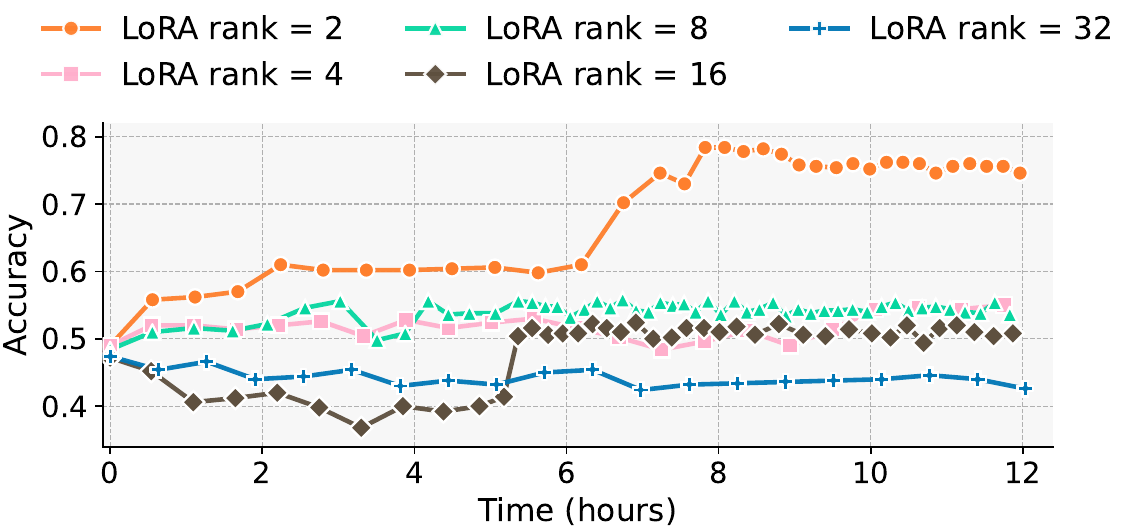}\caption{Population size 24}\end{subfigure}\hfill
  \begin{subfigure}{0.32\textwidth}\includegraphics[width=\linewidth]{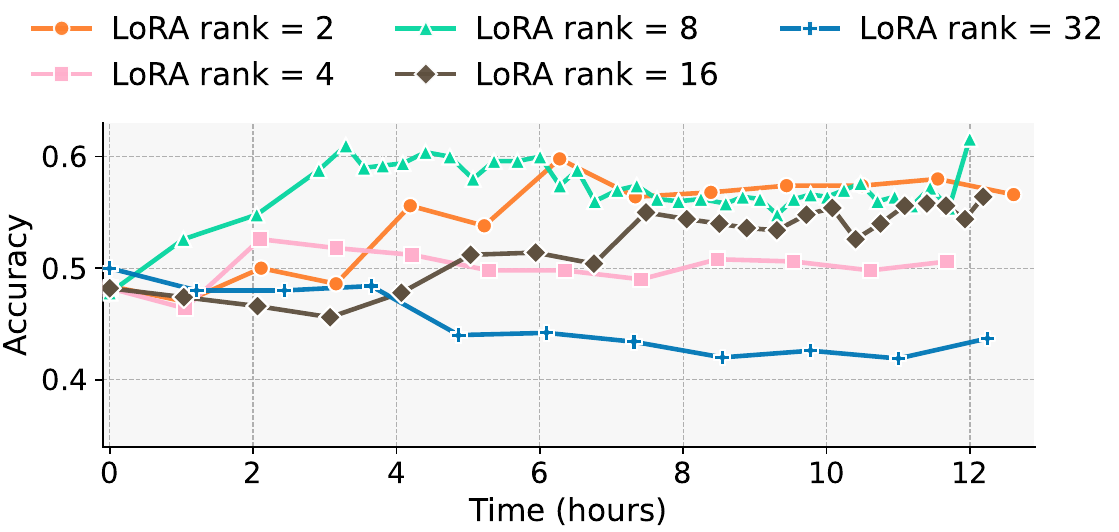}\caption{Population size 48}\end{subfigure}\\[6pt]
  \hspace*{\fill}%
  \begin{subfigure}{0.32\textwidth}\includegraphics[width=\linewidth]{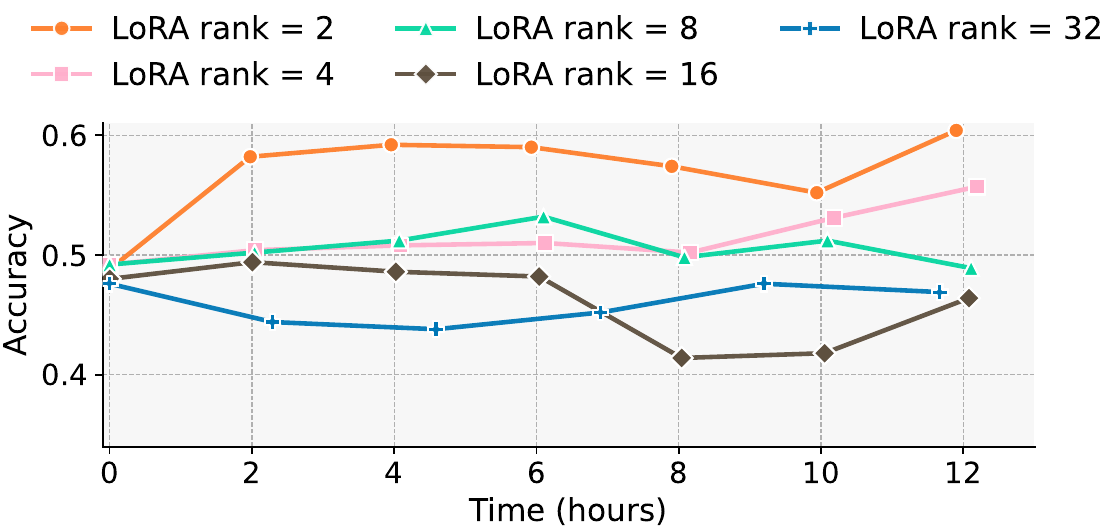}\caption{Population size 96}\end{subfigure}\hfill
  \begin{subfigure}{0.32\textwidth}\includegraphics[width=\linewidth]{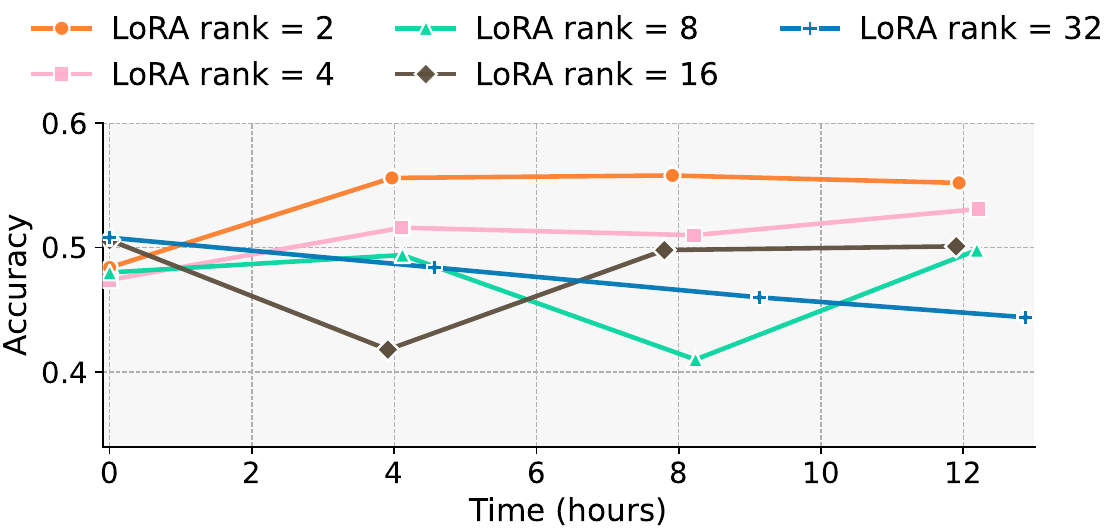}\caption{Population size 192}\end{subfigure}%
  \hspace*{\fill}
  \caption{Validation accuracy over time on \textbf{PRM800K} with \textbf{Qwen2.5-Math-7B}
    when varying the LoRA rank (2, 4, 8, 16, 32)
    at fixed population sizes
    (8, 24, 48, 96, 192). Batch size 300.}
  \label{fig:train_qwen_math_prm_pops}
\end{figure*}

\begin{figure*}[p]
  \centering
  \begin{subfigure}{0.32\textwidth}\includegraphics[width=\linewidth]{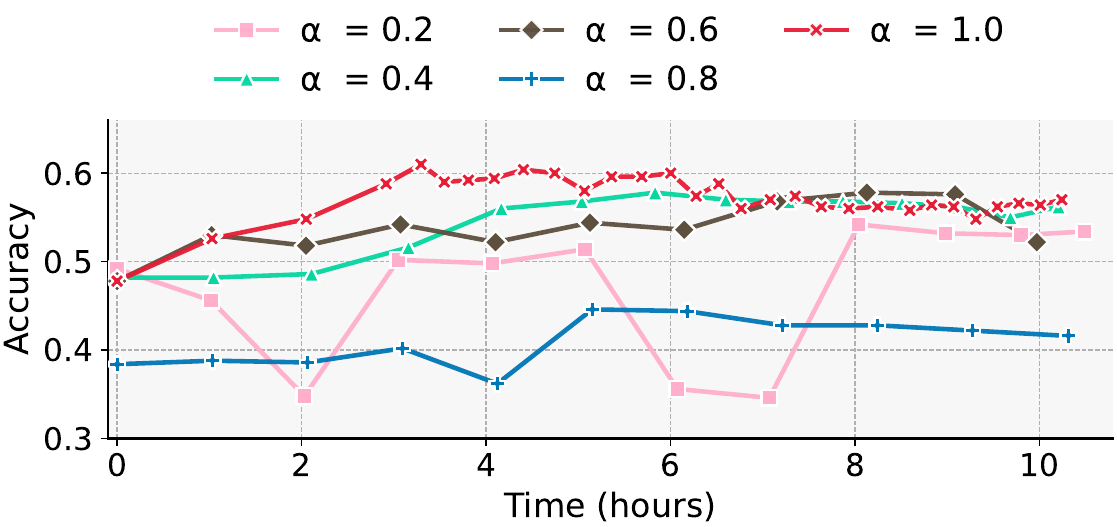}\caption{LoRA rank 8, Population size 48}\end{subfigure}\hfill
  \begin{subfigure}{0.32\textwidth}\includegraphics[width=\linewidth]{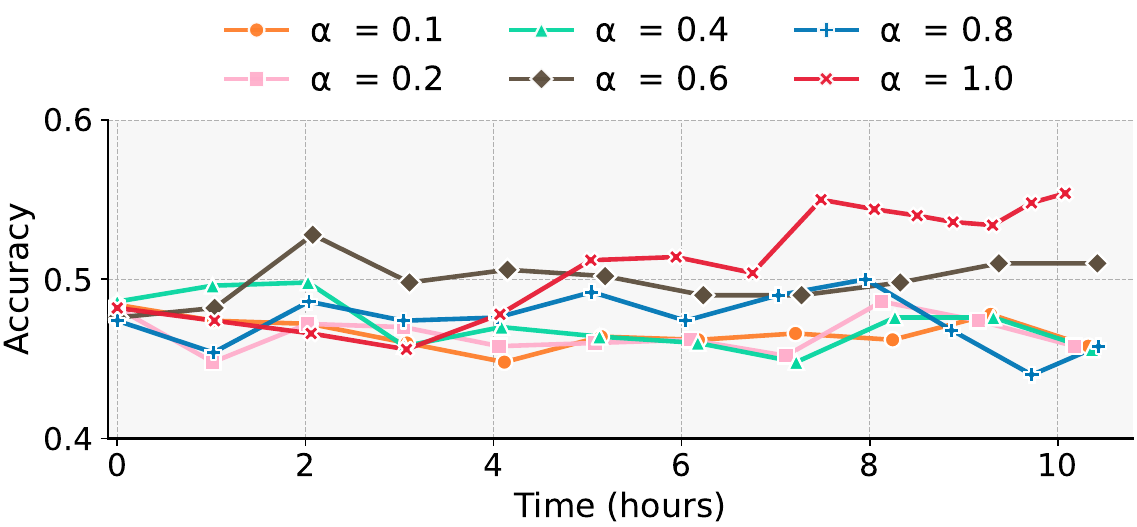}\caption{LoRA rank 16, Population size 48}\end{subfigure}\hfill
  \begin{subfigure}{0.32\textwidth}\includegraphics[width=\linewidth]{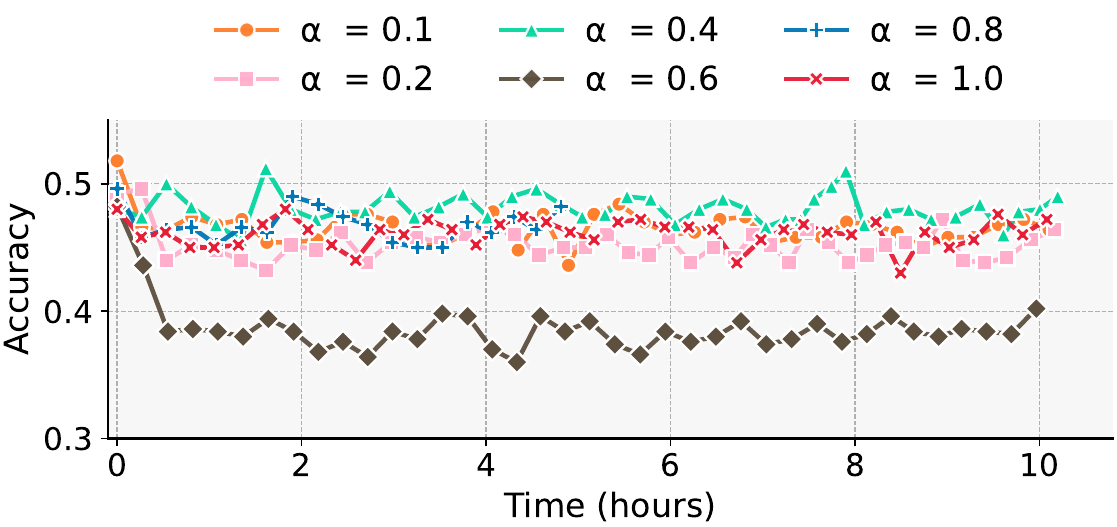}\caption{LoRA rank 32, Population size 8}\end{subfigure}
  \caption{Validation accuracy over time on \textbf{PRM800K} with \textbf{Qwen2.5-Math-7B}
    when varying $\alpha$. LoRA rank and population size are fixed to the optimal choices
    from Figure~\ref{fig:essa_hypers_qwen_math_prm}. Batch size 300.}
  \label{fig:train_qwen_math_prm_alphas}
\end{figure*}

\begin{figure*}[p]
  \centering
  \begin{subfigure}{0.32\textwidth}\includegraphics[width=\linewidth]{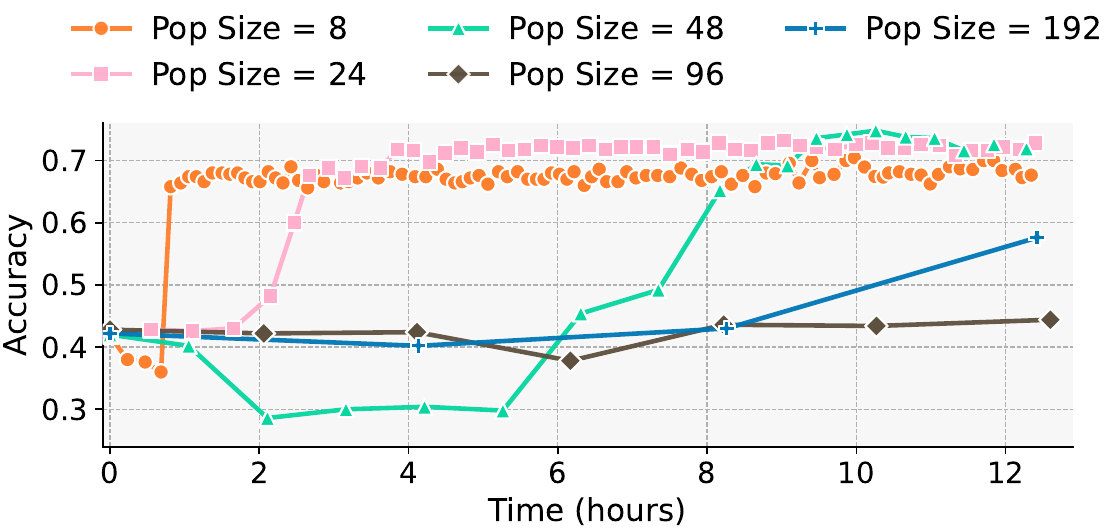}\caption{LoRA rank 2}\end{subfigure}\hfill
  \begin{subfigure}{0.32\textwidth}\includegraphics[width=\linewidth]{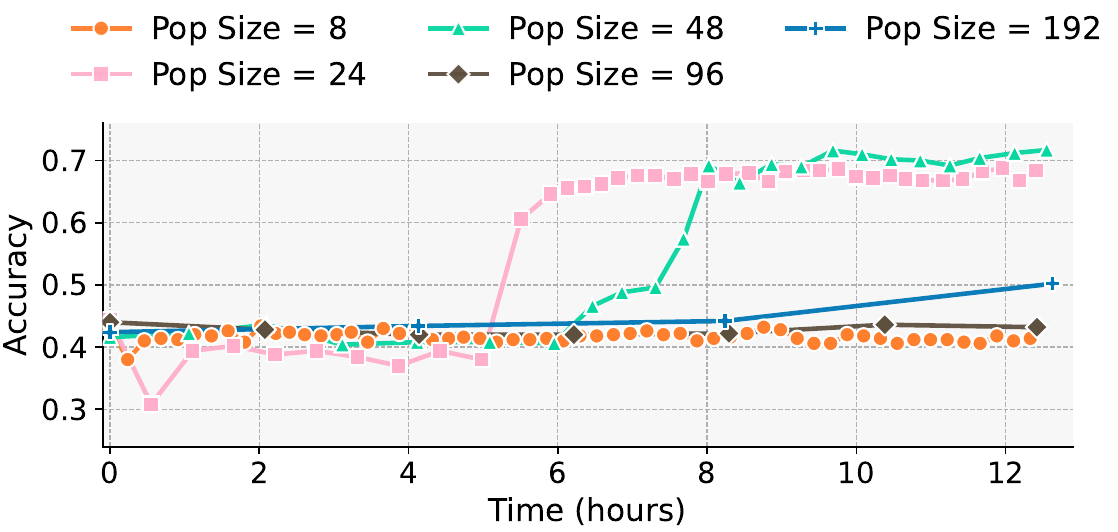}\caption{LoRA rank 4}\end{subfigure}\hfill
  \begin{subfigure}{0.32\textwidth}\includegraphics[width=\linewidth]{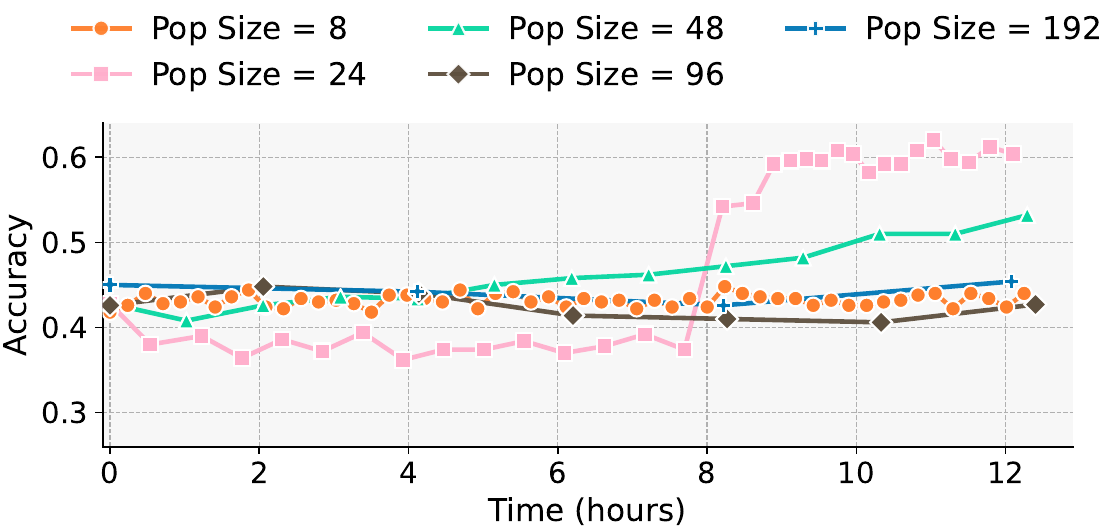}\caption{LoRA rank 8}\end{subfigure}\\[6pt]
  \hspace*{\fill}%
  \begin{subfigure}{0.32\textwidth}\includegraphics[width=\linewidth]{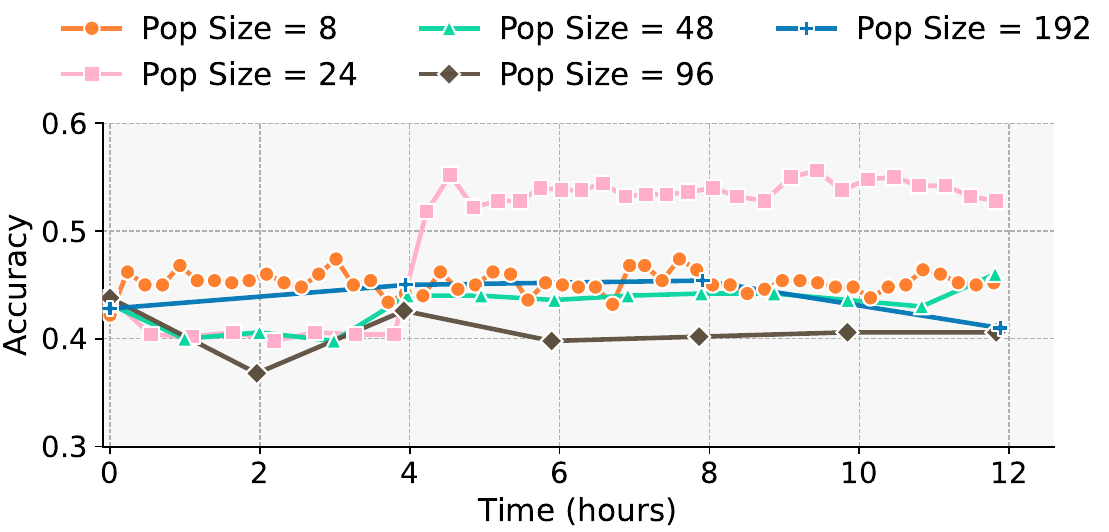}\caption{LoRA rank 16}\end{subfigure}\hfill
  \begin{subfigure}{0.32\textwidth}\includegraphics[width=\linewidth]{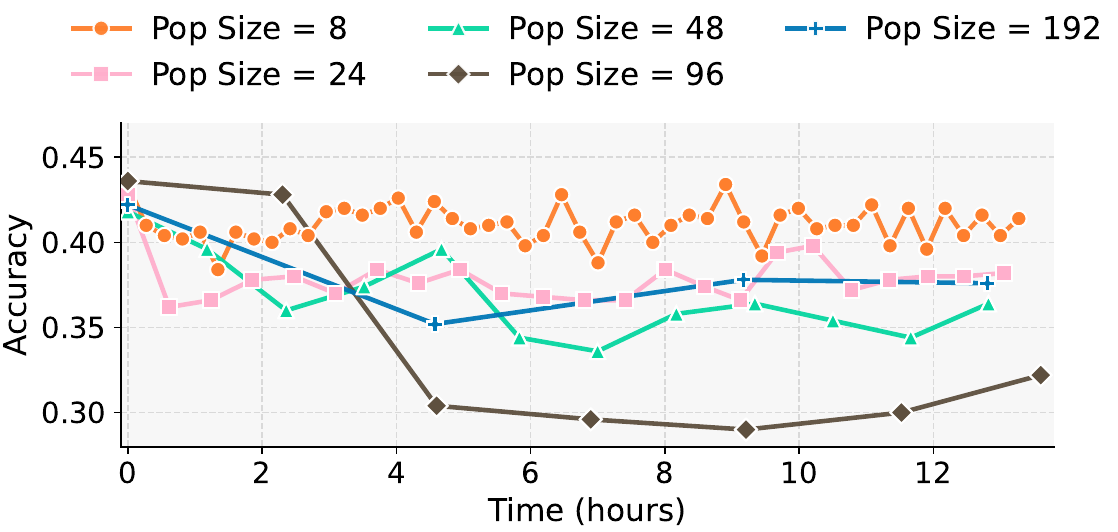}\caption{LoRA rank 32}\end{subfigure}%
  \hspace*{\fill}
  \caption{Validation accuracy over time on \textbf{PRM800K} with \textbf{Qwen2.5-7B}
    when varying the \textsc{ESSA} population size
    (8, 24, 48, 96, 192) at fixed LoRA ranks
    (2, 4, 8, 16, 32). Batch size 300.}
  \label{fig:train_qwen_prm_loras}
\end{figure*}

\begin{figure*}[p]
  \centering
  \begin{subfigure}{0.32\textwidth}\includegraphics[width=\linewidth]{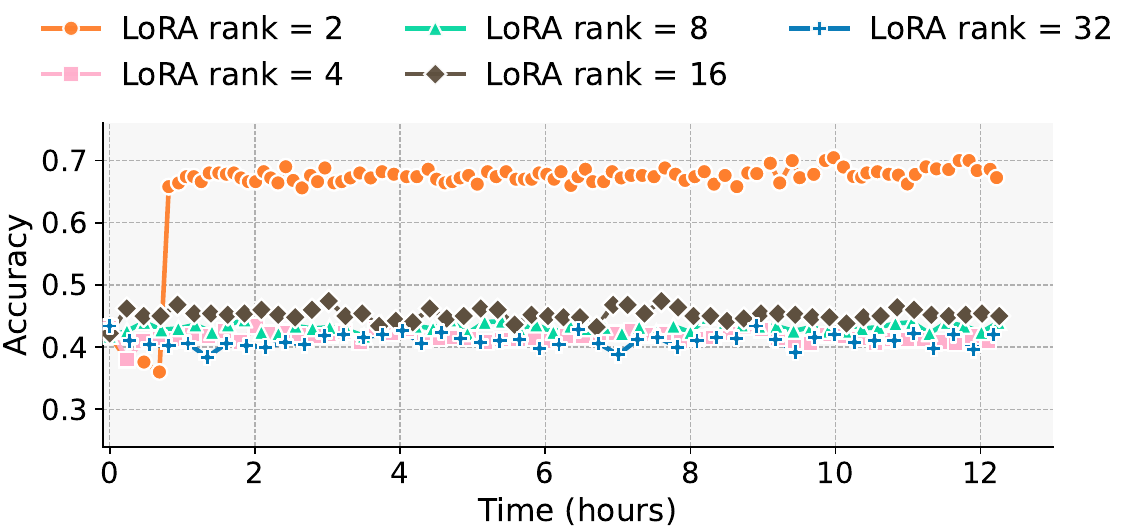}\caption{Population size 8}\end{subfigure}\hfill
  \begin{subfigure}{0.32\textwidth}\includegraphics[width=\linewidth]{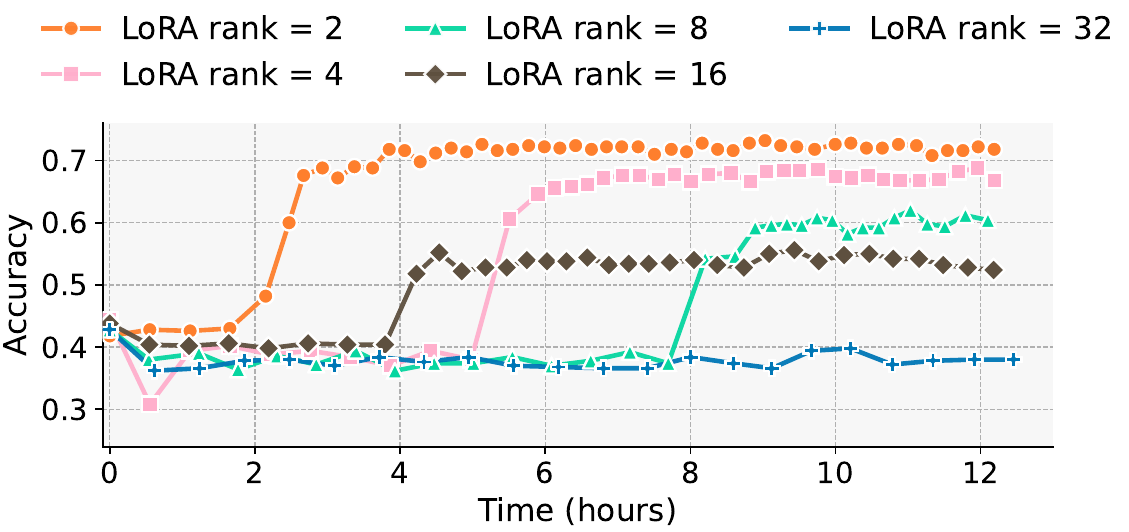}\caption{Population size 24}\end{subfigure}\hfill
  \begin{subfigure}{0.32\textwidth}\includegraphics[width=\linewidth]{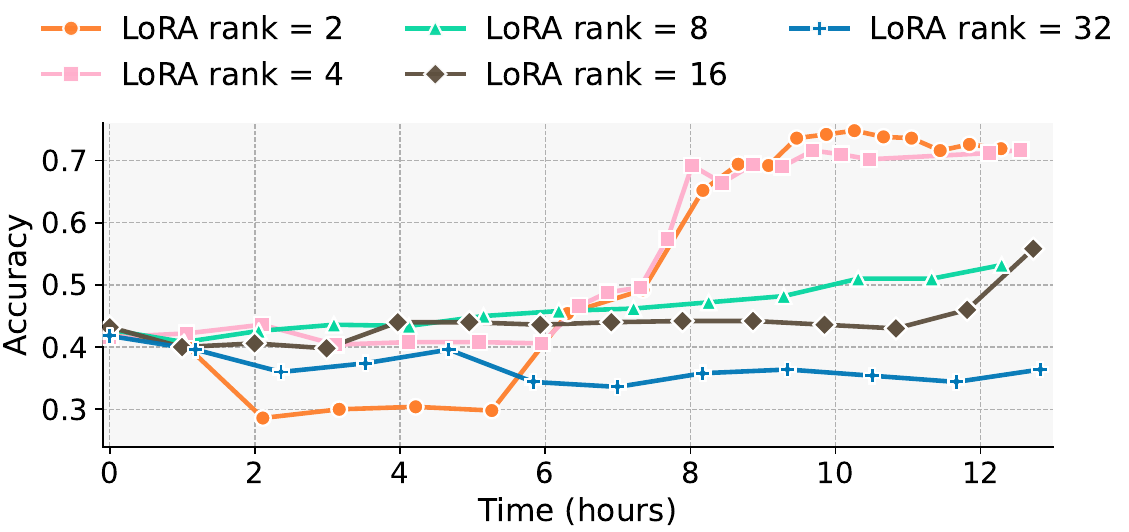}\caption{Population size 48}\end{subfigure}\\[6pt]
  \hspace*{\fill}%
  \begin{subfigure}{0.32\textwidth}\includegraphics[width=\linewidth]{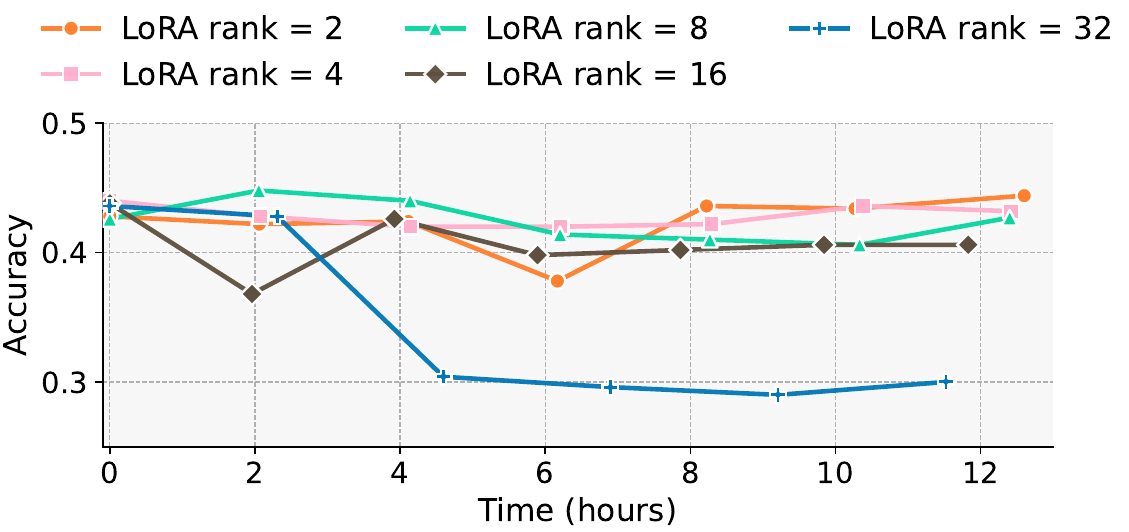}\caption{Population size 96}\end{subfigure}\hfill
  \begin{subfigure}{0.32\textwidth}\includegraphics[width=\linewidth]{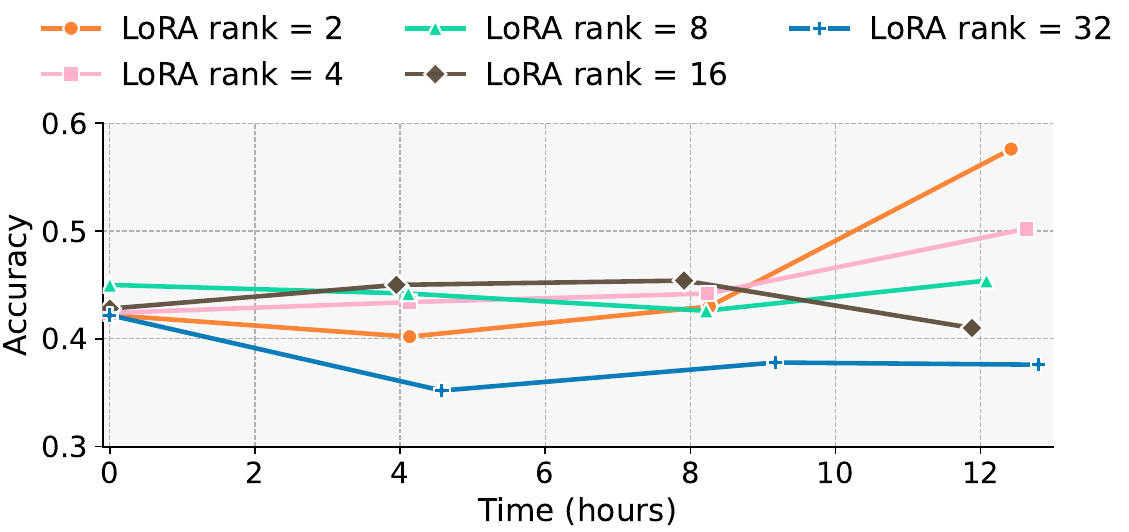}\caption{Population size 192}\end{subfigure}%
  \hspace*{\fill}
  \caption{Validation accuracy over time on \textbf{PRM800K} with \textbf{Qwen2.5-7B}
    when varying the LoRA rank (2, 4, 8, 16, 32)
    at fixed population sizes
    (8, 24, 48, 96, 192). Batch size 300.}
  \label{fig:train_qwen_prm_pops}
\end{figure*}

\begin{figure*}[p]
  \centering
  \begin{subfigure}{0.32\textwidth}\includegraphics[width=\linewidth]{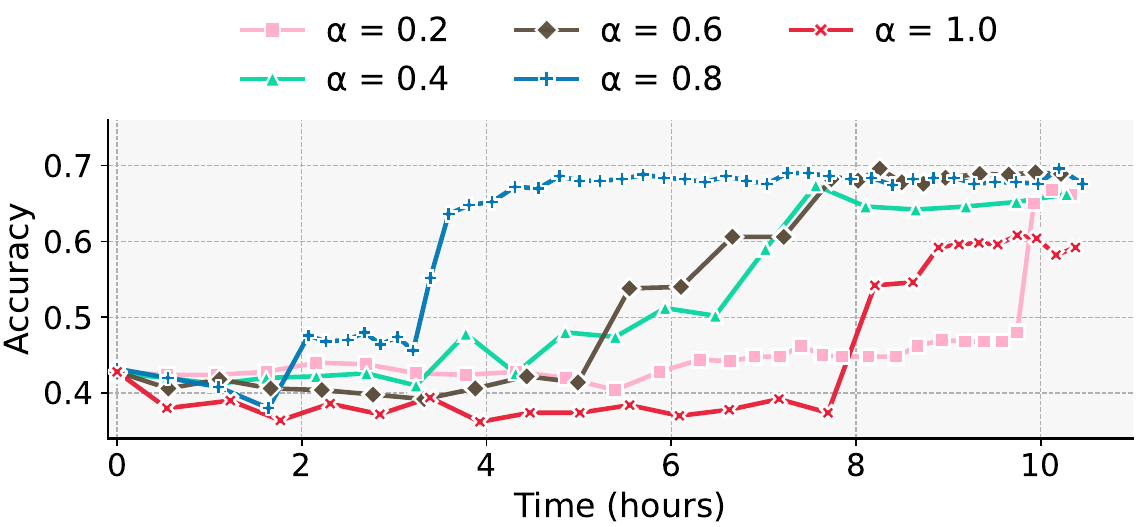}\caption{LoRA rank 8, Population size 24}\end{subfigure}\hfill
  \begin{subfigure}{0.32\textwidth}\includegraphics[width=\linewidth]{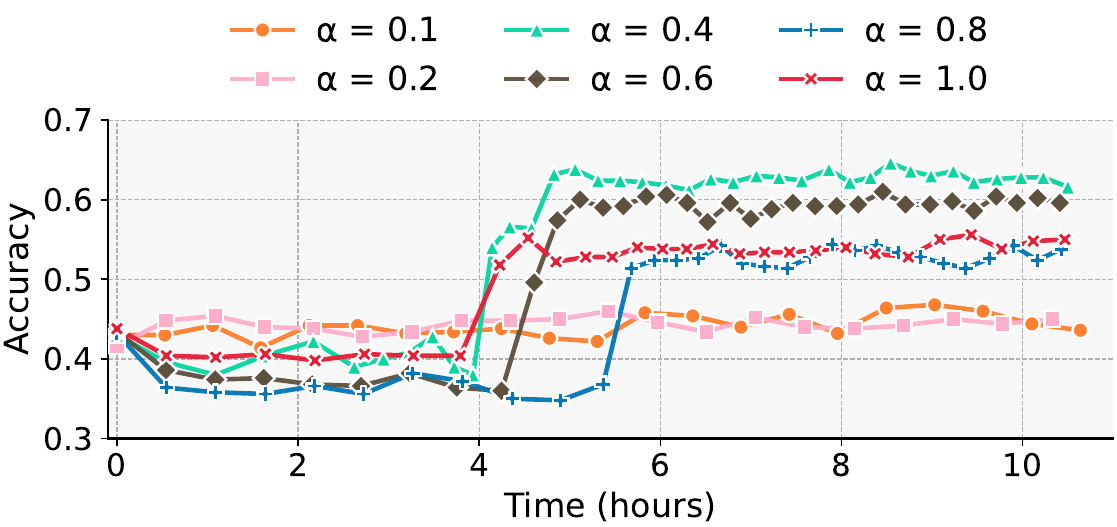}\caption{LoRA rank 16, Population size 24}\end{subfigure}\hfill
  \begin{subfigure}{0.32\textwidth}\includegraphics[width=\linewidth]{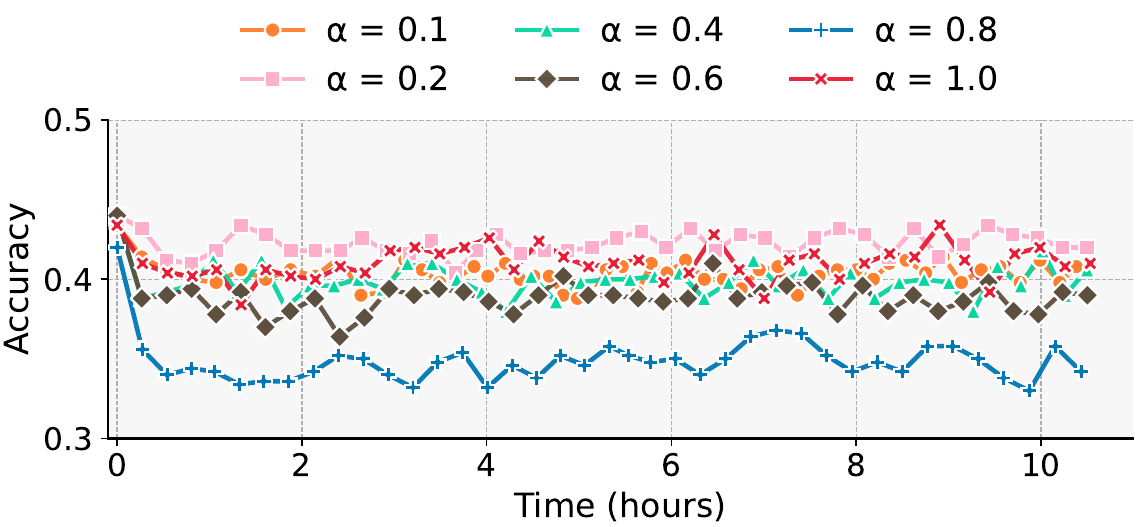}\caption{LoRA rank 32, Population size 8}\end{subfigure}
  \caption{Validation accuracy over time on \textbf{PRM800K} with
    \textbf{Qwen2.5-7B} while varying $\alpha$.
    LoRA rank and population size are fixed to the optimal choices
    from Figure~\ref{fig:essa_hypers_qwen_prm}. Batch size 300.}
  \label{fig:train_qwen_prm_alphas}
\end{figure*}

\clearpage

\begingroup
\makeatletter
\setlength{\@dblfptop}{0pt}
\setlength{\@dblfpsep}{56pt}
\setlength{\@dblfpbot}{0pt plus 1fil}
\makeatother

\begin{figure*}[p]
  \centering
  \begin{subfigure}{0.32\textwidth}\includegraphics[width=\linewidth]{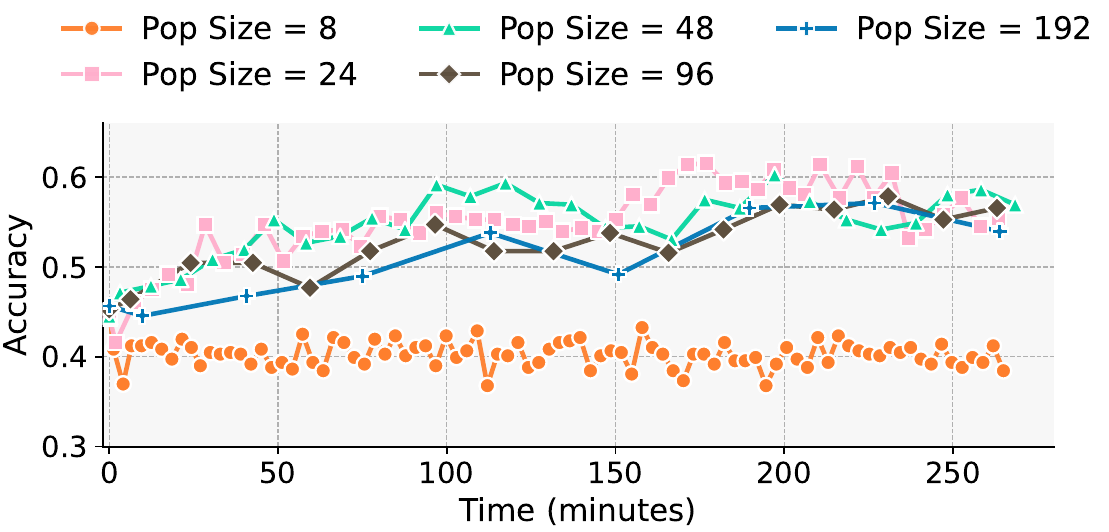}\caption{LoRA rank 2}\end{subfigure}\hfill
  \begin{subfigure}{0.32\textwidth}\includegraphics[width=\linewidth]{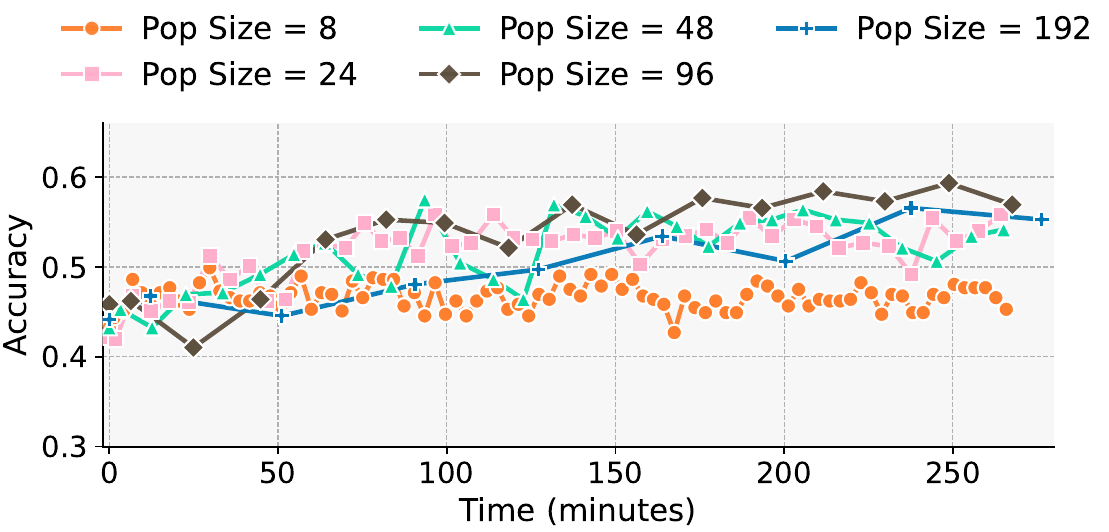}\caption{LoRA rank 4}\end{subfigure}\hfill
  \begin{subfigure}{0.32\textwidth}\includegraphics[width=\linewidth]{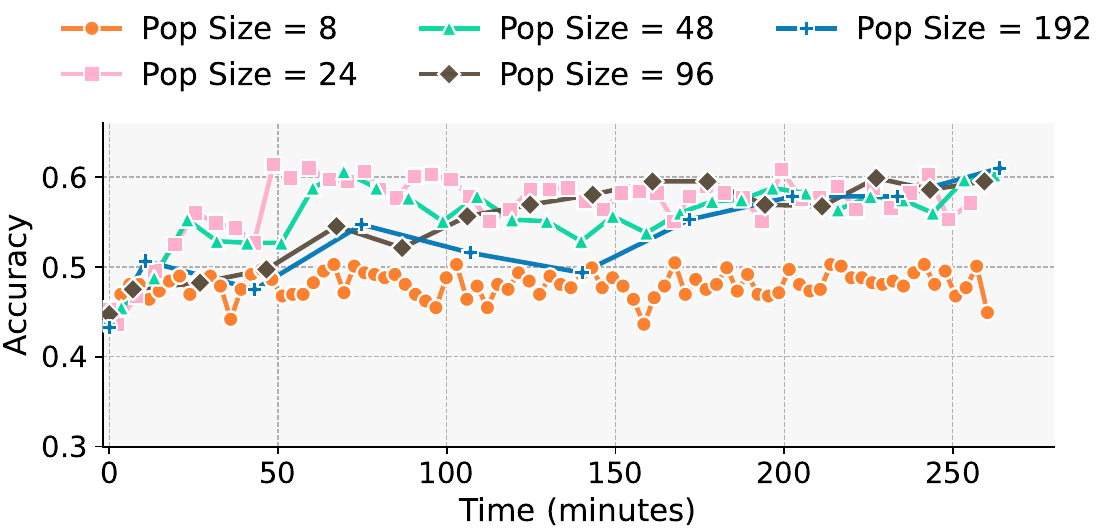}\caption{LoRA rank 8}\end{subfigure}\\[6pt]
  \hspace*{\fill}%
  \begin{subfigure}{0.32\textwidth}\includegraphics[width=\linewidth]{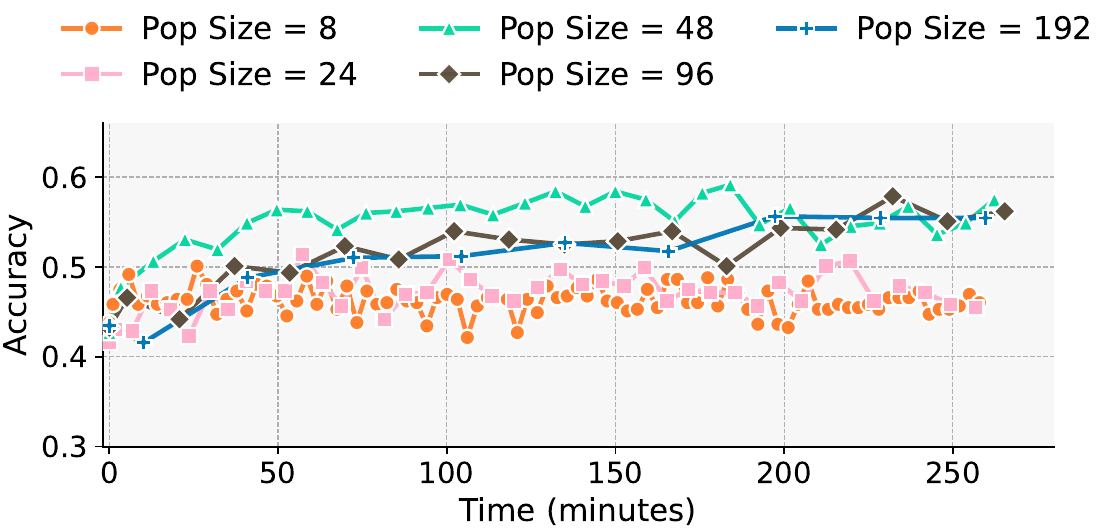}\caption{LoRA rank 16}\end{subfigure}\hfill
  \begin{subfigure}{0.32\textwidth}\includegraphics[width=\linewidth]{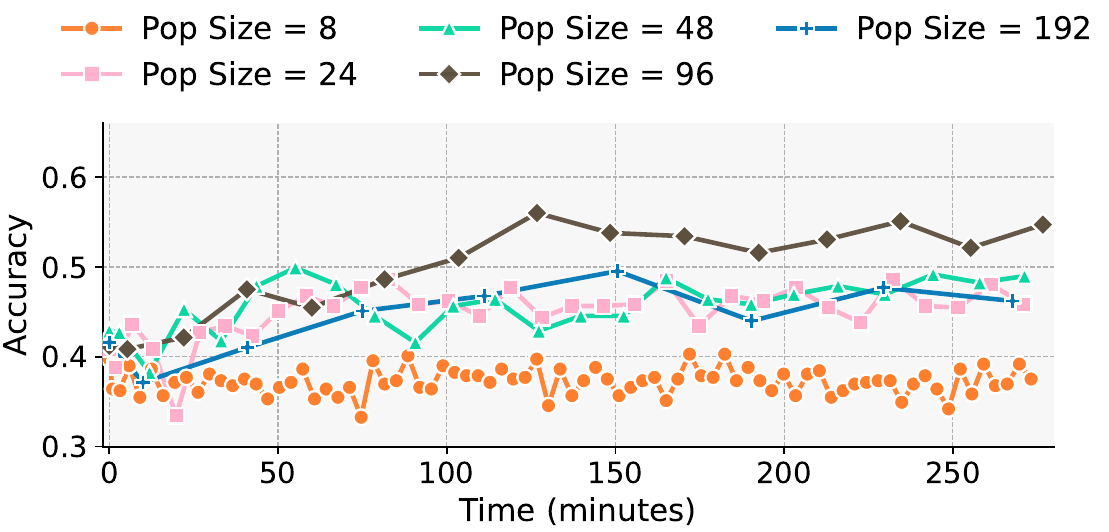}\caption{LoRA rank 32}\end{subfigure}%
  \hspace*{\fill}
  \caption{Validation accuracy over time on \textbf{IFEval} with \textbf{Llama-3.1-8B}
    when varying the \textsc{ESSA} population size
    (8, 24, 48, 96, 192) at fixed LoRA ranks
    (2, 4, 8, 16, 32). Batch size 500, $\alpha\!=\!1.0$.}
  \label{fig:train_llama_ifeval_loras}
\end{figure*}

\begin{figure*}[p]
  \centering
  \begin{subfigure}{0.32\textwidth}\includegraphics[width=\linewidth]{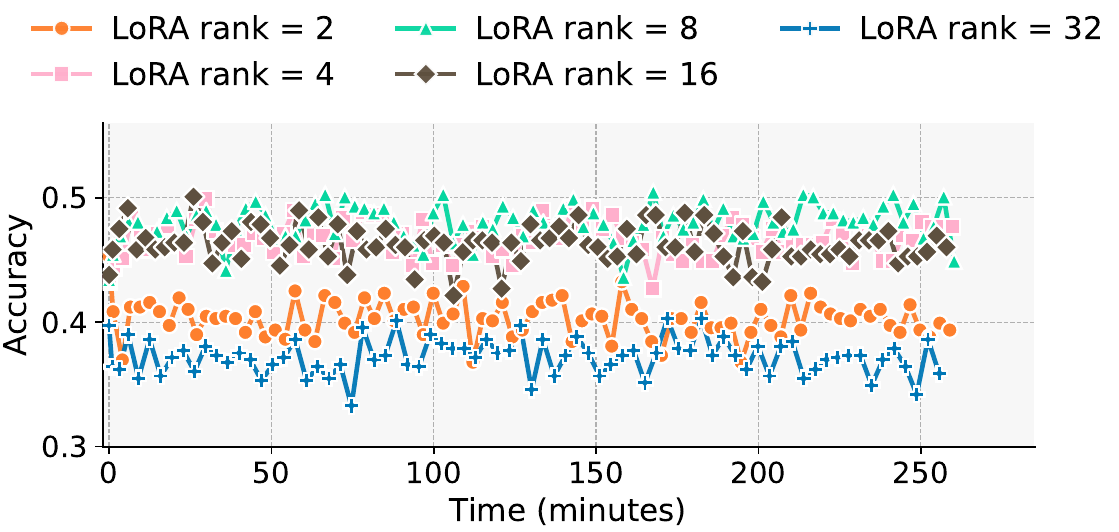}\caption{Population size 8}\end{subfigure}\hfill
  \begin{subfigure}{0.32\textwidth}\includegraphics[width=\linewidth]{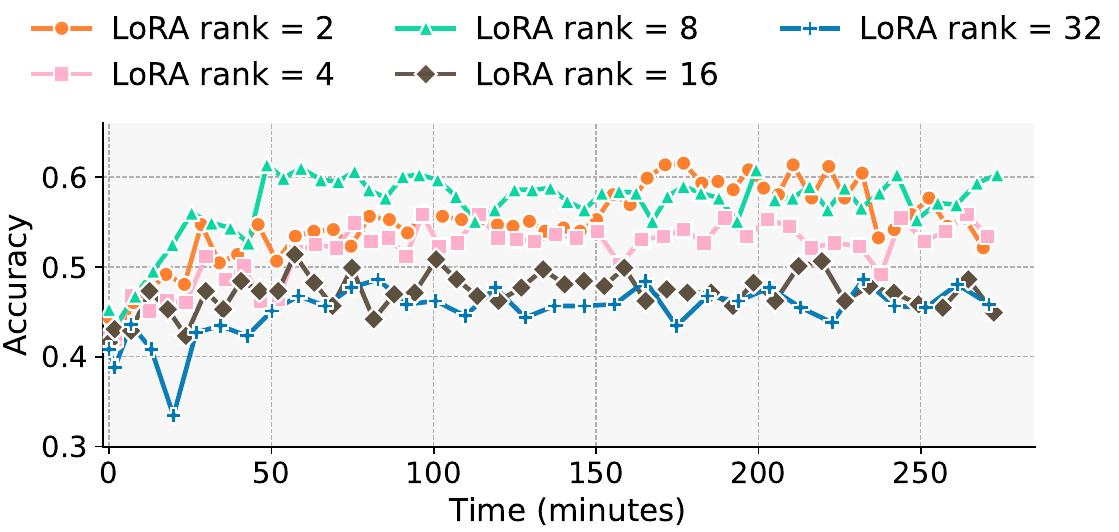}\caption{Population size 24}\end{subfigure}\hfill
  \begin{subfigure}{0.32\textwidth}\includegraphics[width=\linewidth]{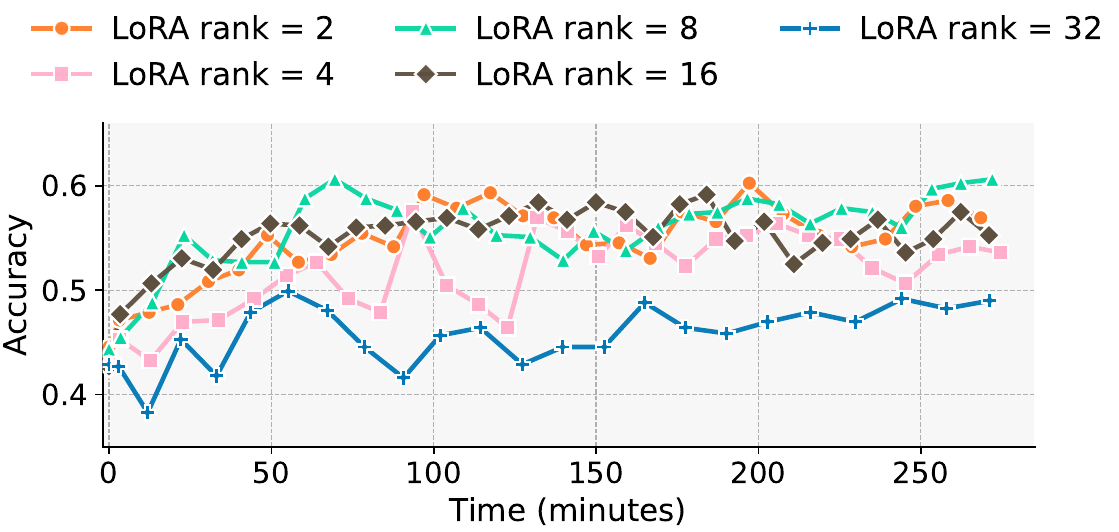}\caption{Population size 48}\end{subfigure}\\[6pt]
  \hspace*{\fill}%
  \begin{subfigure}{0.32\textwidth}\includegraphics[width=\linewidth]{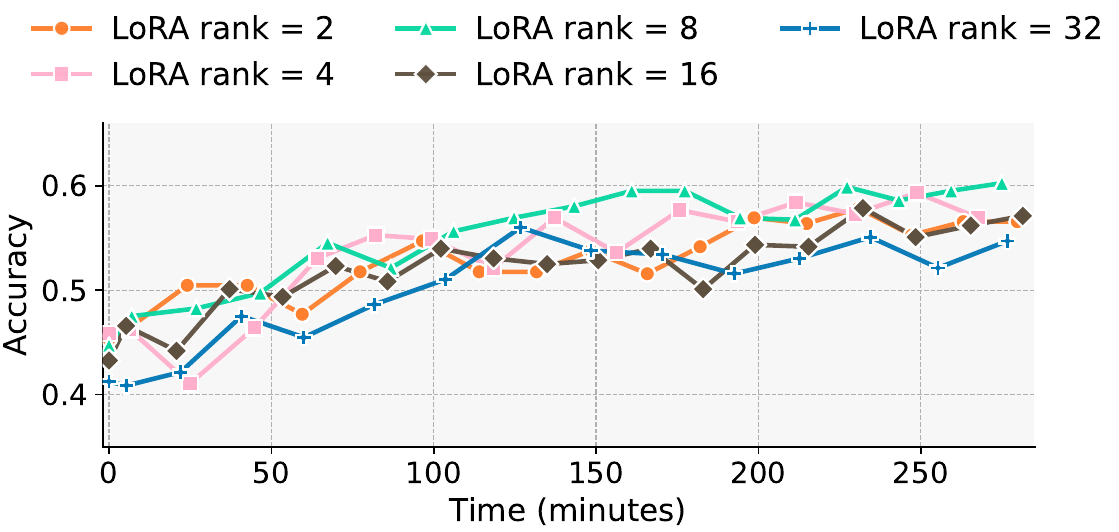}\caption{Population size 96}\end{subfigure}\hfill
  \begin{subfigure}{0.32\textwidth}\includegraphics[width=\linewidth]{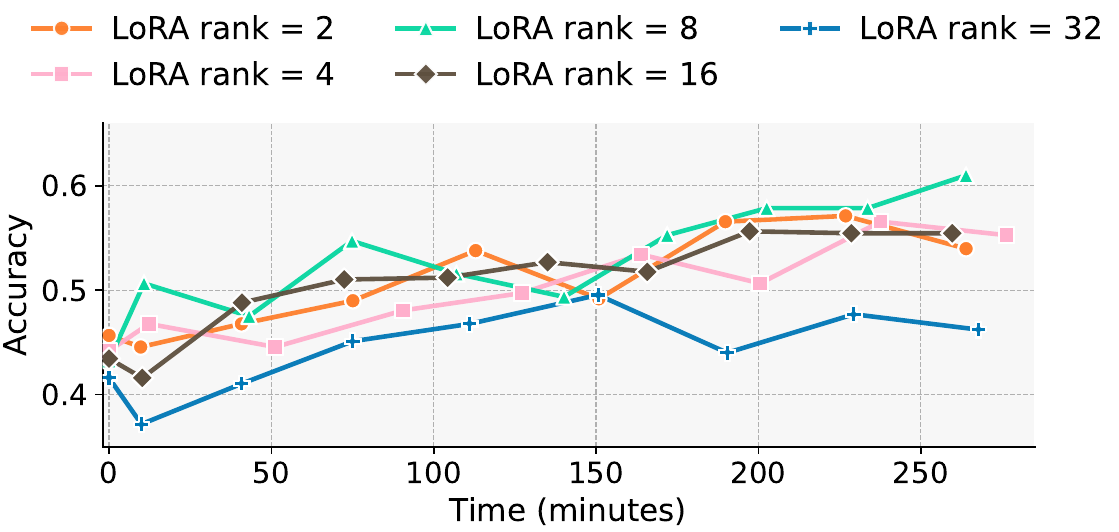}\caption{Population size 192}\end{subfigure}%
  \hspace*{\fill}
  \caption{Validation accuracy over time on \textbf{IFEval} with \textbf{Llama-3.1-8B}
    when varying the LoRA rank (2, 4, 8, 16, 32)
    at fixed population sizes
    (8, 24, 48, 96, 192). Batch size 500, $\alpha\!=\!1.0$.}
  \label{fig:train_llama_ifeval_pops}
\end{figure*}

\clearpage
\endgroup

\section{Comparison to LoRA-GRPO}
\label{app:full_grpo}

This section provides additional ESSA--LoRA-GRPO comparisons
supporting the main-text results in
Sections~\ref{sec:main_comparison}, \ref{sec:optimizer_controls},
\ref{sec:nlp}, and~\ref{sec:scale_math}.

\subsection{School Math}
\label{app:school}

This subsection extends the GSM8K comparison in
Section~\ref{sec:main_comparison}. Figures
\ref{fig:train_qwen_math_gsm_grpo} and
\ref{fig:train_qwen_gsm_grpo} present the results for
Qwen2.5-Math-7B and Qwen2.5-7B, respectively.
Figure~\ref{fig:train_qwen_gsm_3b} presents the results for
Qwen2.5-3B on one GPU. This experiment shows that ESSA's benefits
are not restricted to multi-GPU setups but persist when only one
device is available. Figure~\ref{fig:train_qwen_gsm_base} presents
the results for Qwen2.5-7B and Qwen2.5-Math-7B on GSM8K without an
SFT warm-start. LoRA matrix $A$ is initialized using Kaiming
initialization, while matrix $B$ is initialized to zero. In these
experiments, runs starting from randomly initialized LoRA adapters
achieve lower absolute accuracy.

\subsection{Advanced Math}
\label{app:advanced_math}

This subsection complements the PRM800K$\rightarrow$MATH500
analysis in Section~\ref{sec:optimizer_controls} and the
large-model advanced-math results in Section~\ref{sec:scale_math}.
Figures~\ref{fig:train_qwen_math_prm_grpo} and
\ref{fig:train_qwen_prm_grpo} present the full results for
Qwen2.5-Math-7B and Qwen2.5-7B, respectively. ESSA climbs to high
accuracy more rapidly than LoRA-GRPO and often reaches a higher
plateau, particularly when the LoRA rank is small or moderate.

\subsection{HelpSteer2 dynamics}
\label{app:helpsteer}

This subsection provides the full training dynamics and endpoint
diagnostics for the HelpSteer2 experiment summarized in
Section~\ref{sec:nlp}. Figure~\ref{fig:helpsteer} reports the
HelpSteer2 validation reward over time at three LoRA ranks for
Llama-3.1-8B. Both ESSA and LoRA-GRPO improve over the SFT
initialization. We additionally examine response length, Distinct-2,
generation entropy, and relative parameter drift at the endpoints of
the training trajectories.

\begin{table}[H]
\centering
\footnotesize
\setlength{\tabcolsep}{4pt}
\begin{tabular}{lccc}
\toprule
\textbf{Diagnostic} & \textbf{SFT} & \textbf{ESSA} &
\textbf{LoRA-GRPO} \\
\midrule
Mean response length & 626   & 745   & 722   \\
Distinct-2           & 0.201 & 0.613 & 0.733 \\
Generation entropy   & 0.642 & 0.735 & 1.973 \\
Relative parameter drift & 0.000 & 2.403 & 4.233 \\
\bottomrule
\end{tabular}
\caption{Endpoint diagnostics on \textbf{HelpSteer2} with
\textbf{Llama-3.1-8B} at LoRA rank 16. ESSA uses population size
42, $\alpha\!=\!1.0$, and batch size 100. LoRA-GRPO uses learning
rate $1\!\times\!10^{-5}$, global batch size 512, and minibatch size
64. Mean response length increases moderately for both methods
($+19\%$ for ESSA and $+15\%$ for LoRA-GRPO), providing no
indication of disproportionate length growth. Relative parameter
drift is $\lVert\Delta W_{\mathrm{final}}-\Delta W_{\mathrm{SFT}}\rVert_F/
\lVert\Delta W_{\mathrm{SFT}}\rVert_F$ (zero at the SFT checkpoint).
ESSA remains closer to the SFT policy in both generation entropy
and relative parameter drift, while both methods increase
Distinct-2.}
\label{tab:helpsteer_diagnostics}
\end{table}

\subsection{HH-RLHF dynamics}
\label{app:hh_rlhf}

This subsection provides the full training dynamics for the
safety/helpfulness experiment summarized in
Section~\ref{sec:nlp}. Figure~\ref{fig:hh_rlhf} shows the in-domain
reward over time for Mistral-7B trained on Anthropic HH-RLHF with
the OpenAssistant DeBERTa-v3 reward model. ESSA crosses LoRA-GRPO
within the first hour and reaches a higher plateau by the end of
training.

\newpage

\section{Precision Analysis}
\label{app:precision}

This section provides the full training dynamics for the precision
study summarized in Section~\ref{sec:sensitivity}.

Figure~\ref{fig:precision_time} shows the full training curves of
ESSA on PRM800K with Qwen2.5-32B, LoRA rank $8$, population size
$64$, $\alpha\!=\!1.0$, and per-candidate batch size $256$ under
three numerical precisions: BF16, INT8, and INT4. Shaded areas
indicate one standard deviation over five independent seeds.

Across all three precisions, accuracy rises quickly within the first
hour and then remains stable. Reducing precision from BF16 to INT8
or INT4 produces only a small reduction in final accuracy, consistent
with the main-text results in Table~\ref{tab:precision}. In particular,
the difference between BF16 and INT4 is 1.1 test-accuracy percentage
points. These results show that ESSA can use aggressive weight
quantization while retaining similar final performance.

\begin{figure}[!t]
\centering
\includegraphics[width=0.46\textwidth]{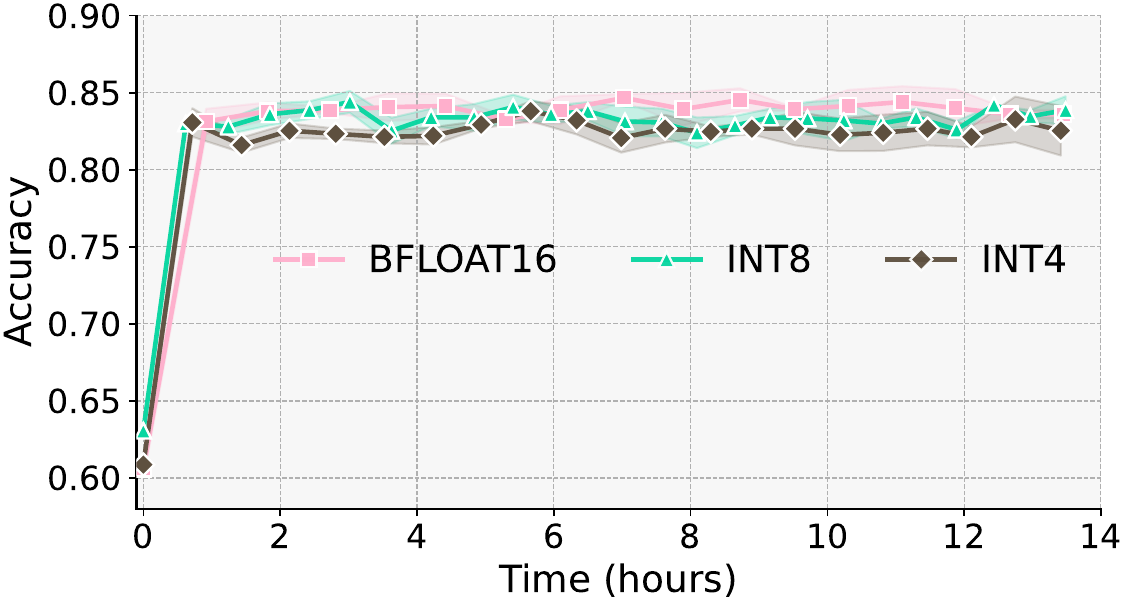}
\caption{Validation accuracy over training time for \textbf{Qwen2.5-32B} on \textbf{PRM800K}
under different weight precisions (BFLOAT16, INT8, INT4).
Settings: LoRA rank $8$, pop. $64$, batch size $256$, $\alpha\!=\!1.0$.}
\label{fig:precision_time}
\end{figure}

\clearpage

\begin{figure*}[p]
  \centering
  \begin{subfigure}{0.32\textwidth}\includegraphics[width=\linewidth]{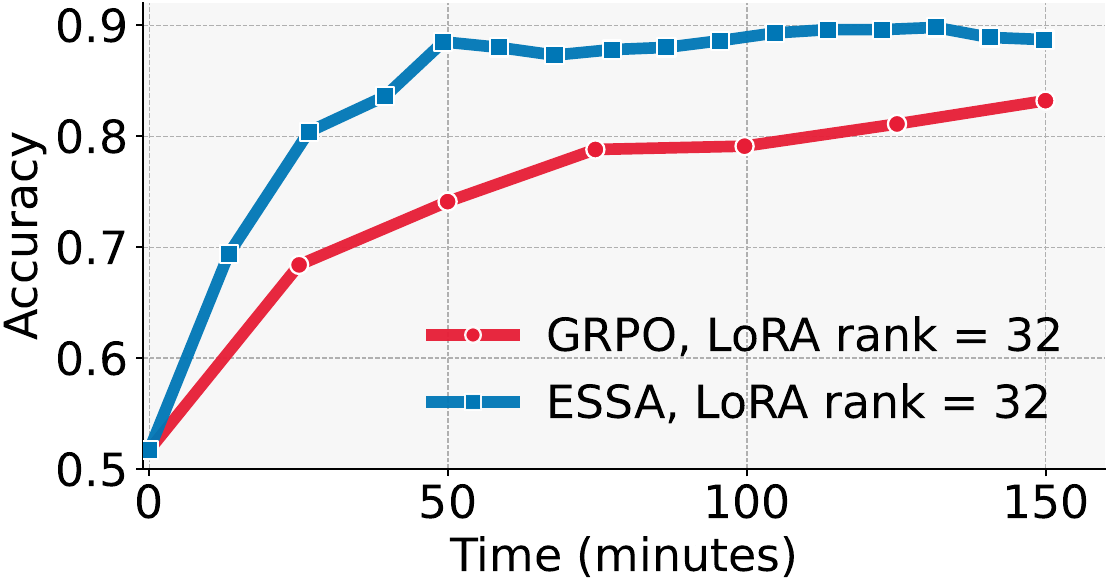}\caption{Population size 96, $\alpha\!=\!1.0$}\end{subfigure}\hfill
  \begin{subfigure}{0.32\textwidth}\includegraphics[width=\linewidth]{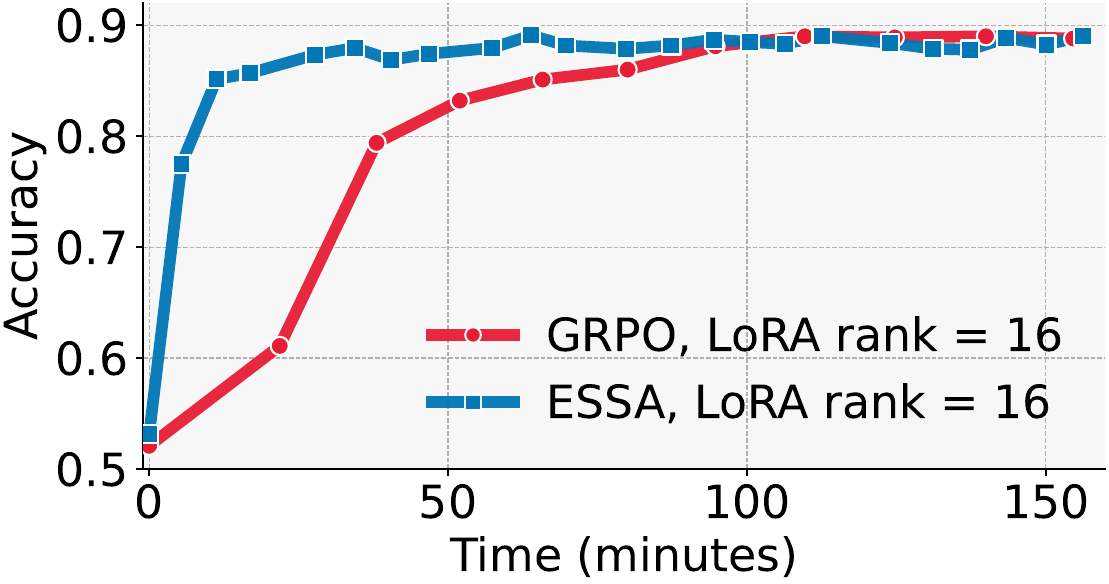}\caption{Population size 48, $\alpha\!=\!0.6$}\end{subfigure}\hfill
  \begin{subfigure}{0.32\textwidth}\includegraphics[width=\linewidth]{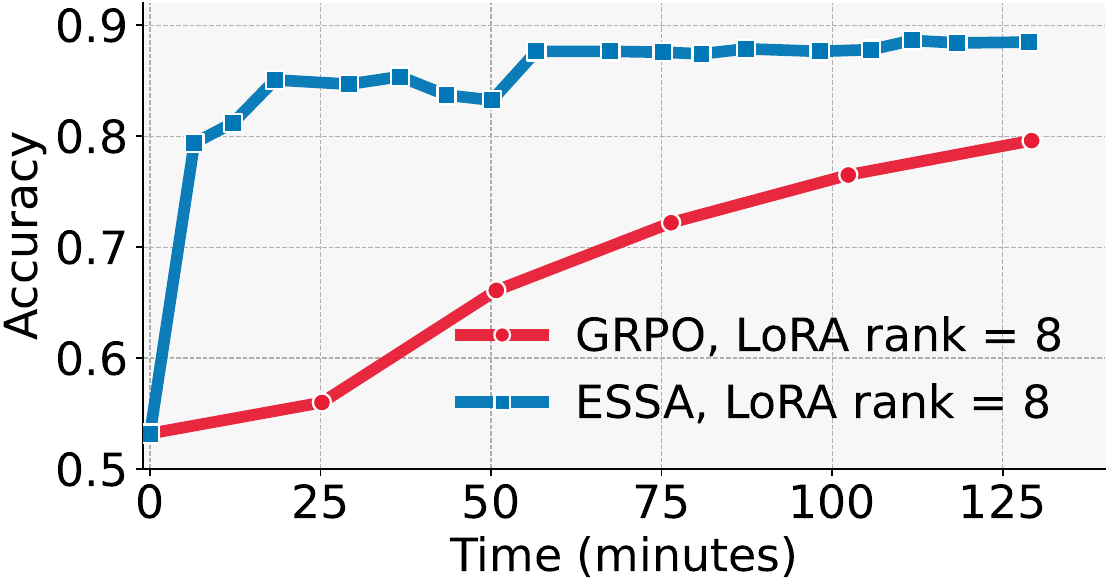}\caption{Population size 48, $\alpha\!=\!1.0$}\end{subfigure}\\[6pt]
  \hspace*{\fill}%
  \begin{subfigure}{0.32\textwidth}\includegraphics[width=\linewidth]{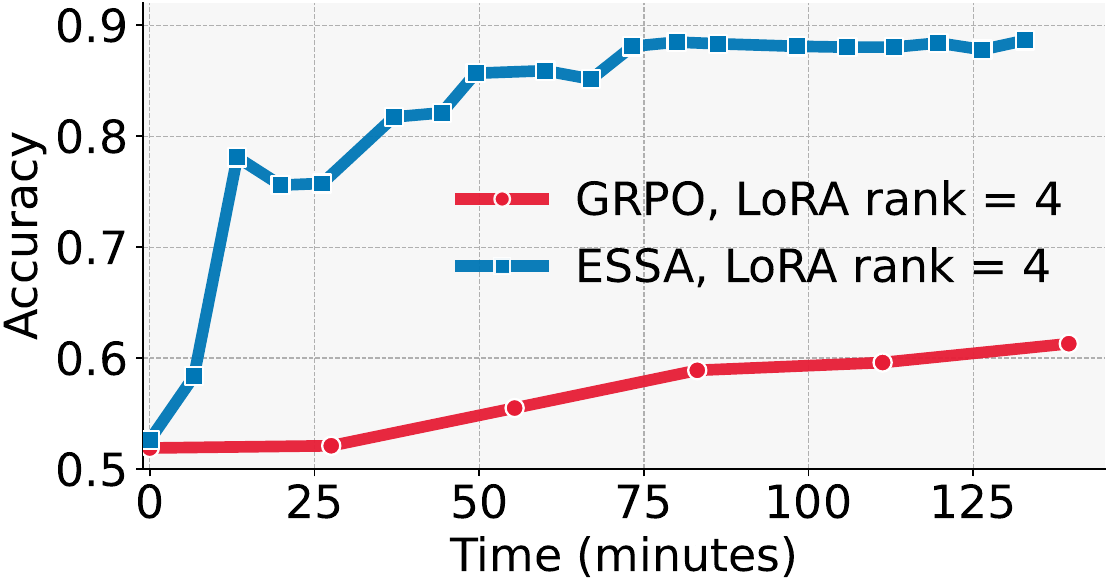}\caption{Population size 48, $\alpha\!=\!1.0$}\end{subfigure}\hfill
  \begin{subfigure}{0.32\textwidth}\includegraphics[width=\linewidth]{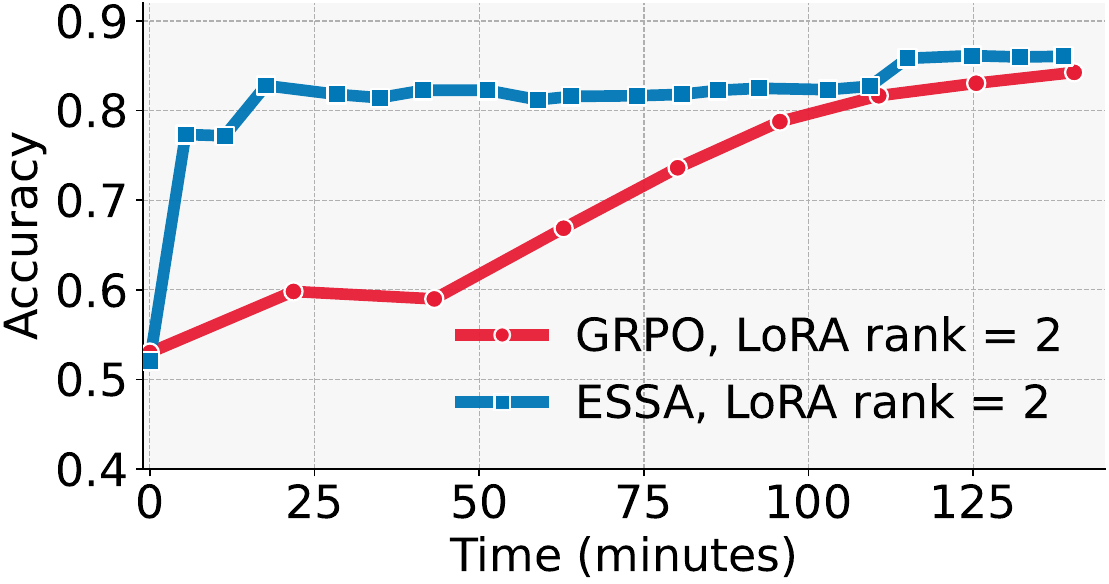}\caption{Population size 48, $\alpha\!=\!1.0$}\end{subfigure}%
  \hspace*{\fill}
  \caption{Validation accuracy over time on \textbf{GSM8K} with \textbf{Qwen2.5-Math-7B}.
  Panels (a)-(e) correspond to LoRA ranks 32, 16, 8, 4, and 2, respectively. ESSA (blue): batch size 100. LoRA-GRPO (red): lr $1\!\times\!10^{-5}$, global batch 512, mini batch 64.}
  \label{fig:train_qwen_math_gsm_grpo}
\end{figure*}

\begin{figure*}[p]
  \centering
  \begin{subfigure}{0.32\textwidth}\includegraphics[width=\linewidth]{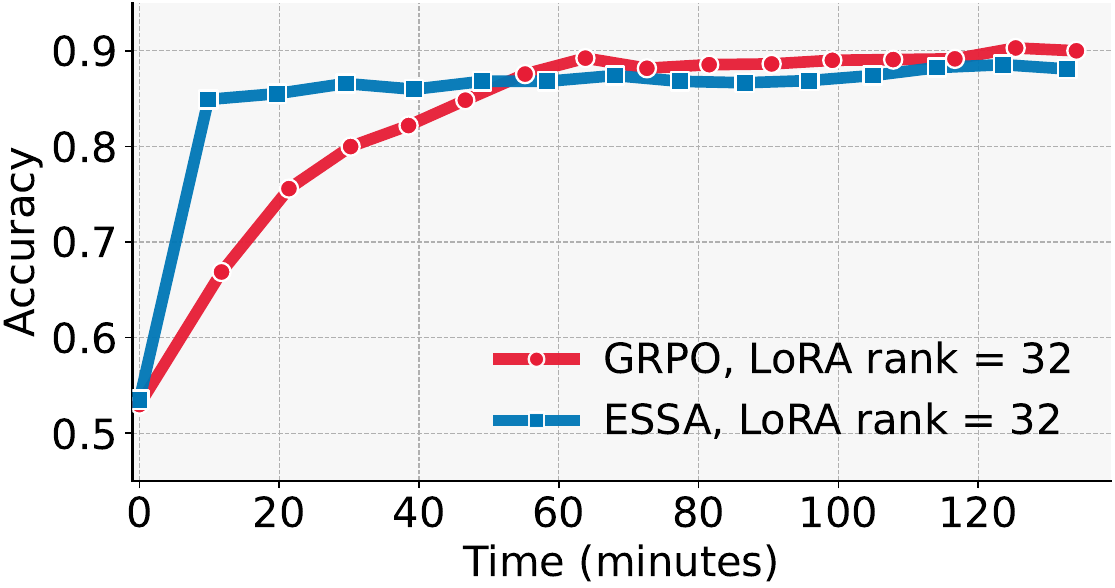}\caption{Population size 96, $\alpha\!=\!0.6$}\end{subfigure}\hfill
  \begin{subfigure}{0.32\textwidth}\includegraphics[width=\linewidth]{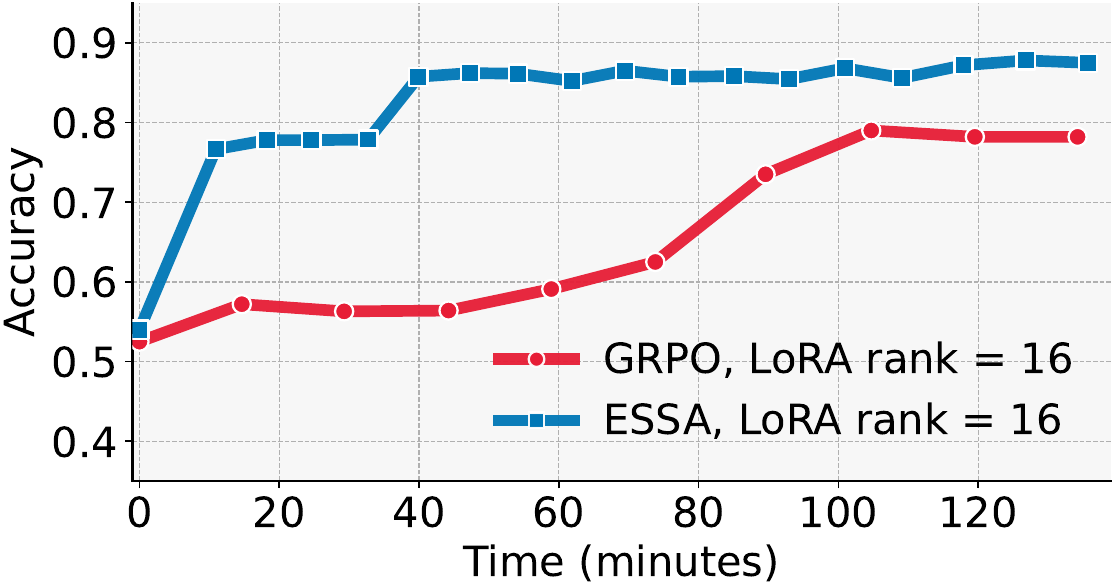}\caption{Population size 96, $\alpha\!=\!0.6$}\end{subfigure}\hfill
  \begin{subfigure}{0.32\textwidth}\includegraphics[width=\linewidth]{images/QWEN7B-GSM8K/8_48_6_grpo_essa.pdf}\caption{Population size 48, $\alpha\!=\!0.6$}\end{subfigure}\\[6pt]
  \hspace*{\fill}%
  \begin{subfigure}{0.32\textwidth}\includegraphics[width=\linewidth]{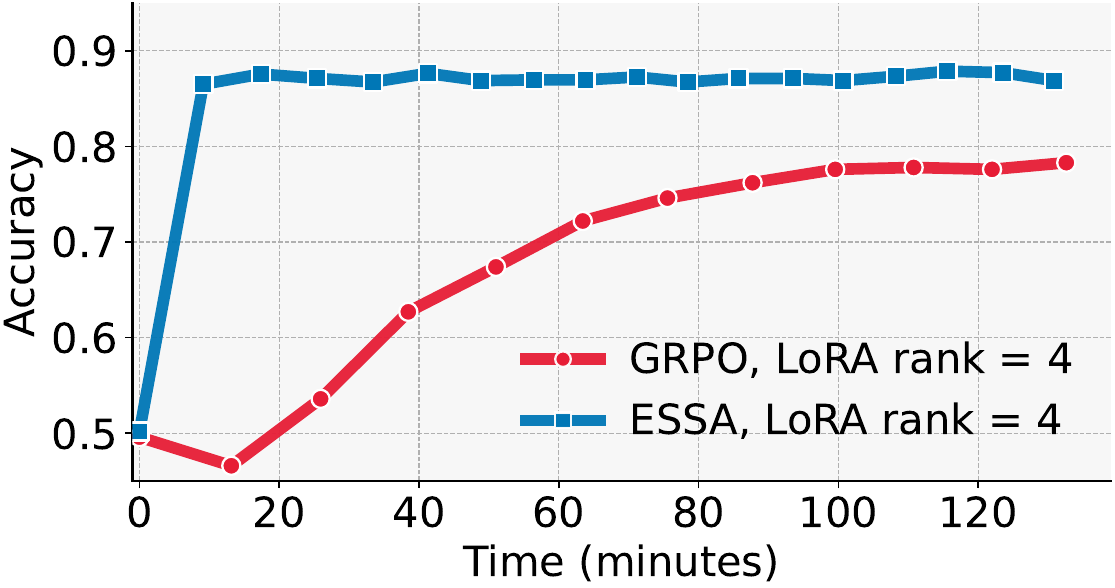}\caption{Population size 96, $\alpha\!=\!1.0$}\end{subfigure}\hfill
  \begin{subfigure}{0.32\textwidth}\includegraphics[width=\linewidth]{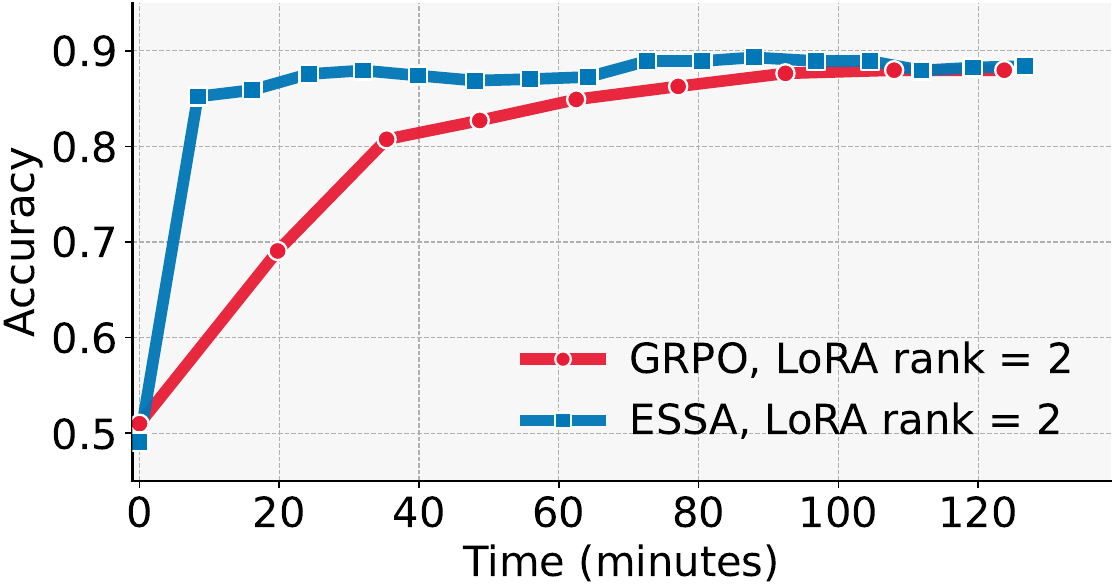}\caption{Population size 96, $\alpha\!=\!1.0$}\end{subfigure}%
  \hspace*{\fill}
  \caption{Validation accuracy over time on \textbf{GSM8K} with \textbf{Qwen2.5-7B}.
  Panels (a)-(e) correspond to LoRA ranks 32, 16, 8, 4, and 2, respectively. ESSA (blue): batch size 100. LoRA-GRPO (red): lr $1\!\times\!10^{-5}$, global batch 512, mini batch 64.}
  \label{fig:train_qwen_gsm_grpo}
\end{figure*}

\begin{figure*}[p]
  \centering
  \begin{subfigure}{0.45\textwidth}
    \includegraphics[width=\linewidth]{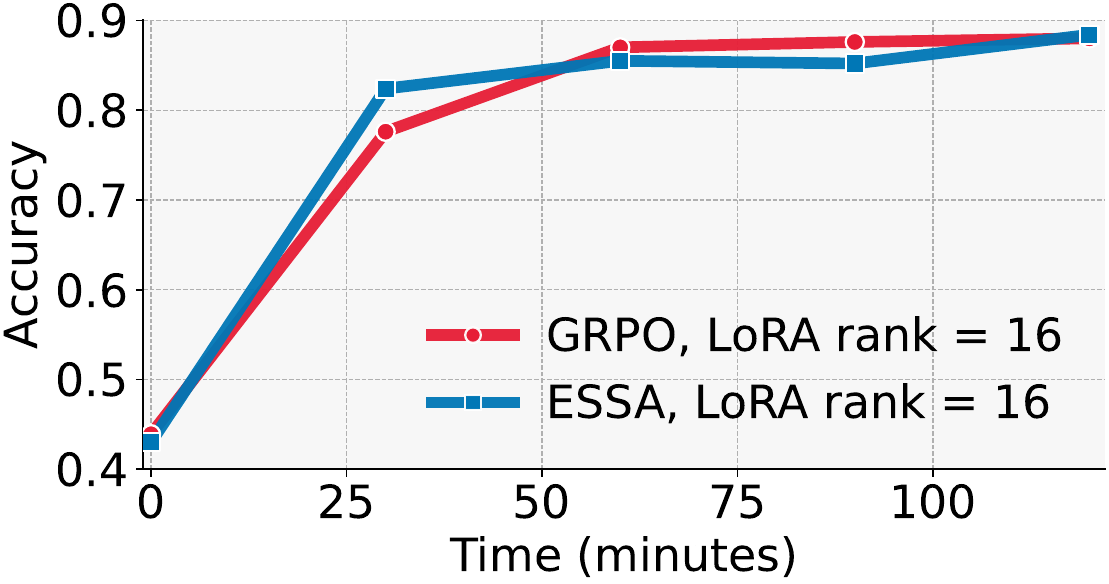}
    \caption{Population size 96, $\alpha\!=\!1.0$}
  \end{subfigure}\hfill
  \begin{subfigure}{0.45\textwidth}
    \includegraphics[width=\linewidth]{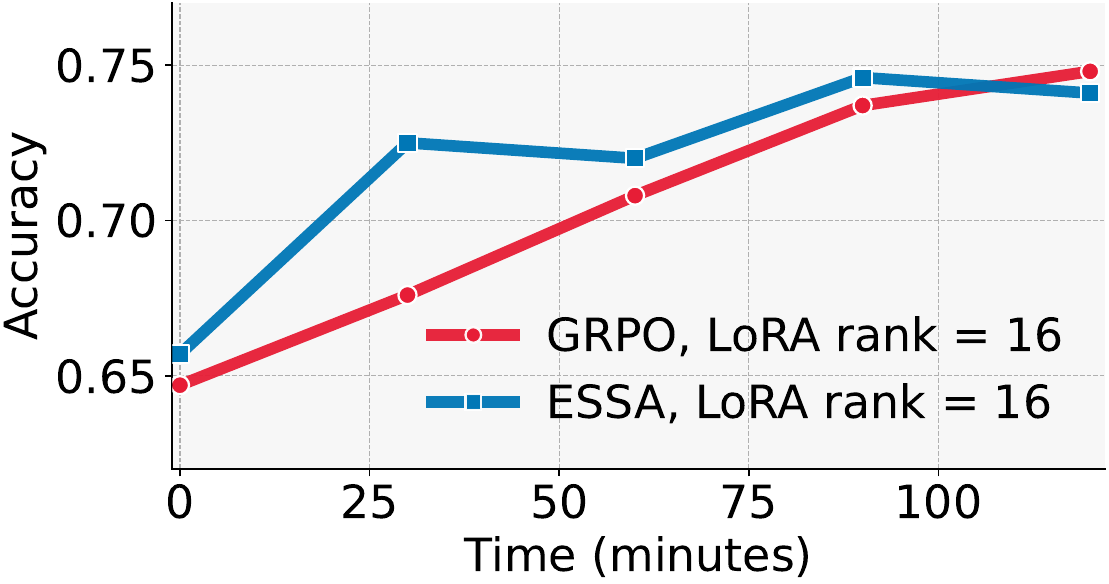}
    \caption{Population size 96, $\alpha\!=\!1.0$}
  \end{subfigure}
  \caption{Validation accuracy over time on \textbf{GSM8K} with \textbf{Qwen2.5-7B} (a) and \textbf{Qwen2.5-Math-7B} (b) without SFT warm-start. ESSA (blue): batch size 100. LoRA-GRPO (red): lr $1\!\times\!10^{-5}$, global batch 512, mini batch 64.}
  \label{fig:train_qwen_gsm_base}
\end{figure*}

\begin{figure*}[p]
  \centering
  \includegraphics[width=0.45\textwidth]{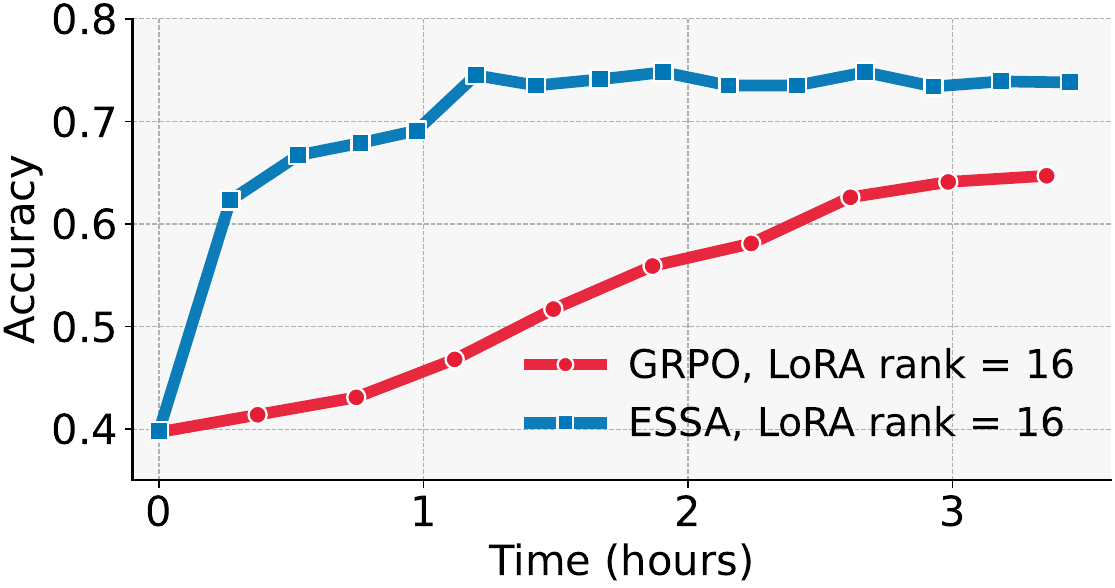}
  \caption{Validation accuracy over time on \textbf{GSM8K} with \textbf{Qwen2.5-3B} on one GPU. ESSA (blue): population size 48, $\alpha\!=\!1.0$, batch size 100. LoRA-GRPO (red): lr $1\!\times\!10^{-5}$, global batch 512, mini batch 64.}
  \label{fig:train_qwen_gsm_3b}
\end{figure*}

\begin{figure*}[p]
  \centering
  \begin{subfigure}{0.32\textwidth}\includegraphics[width=\linewidth]{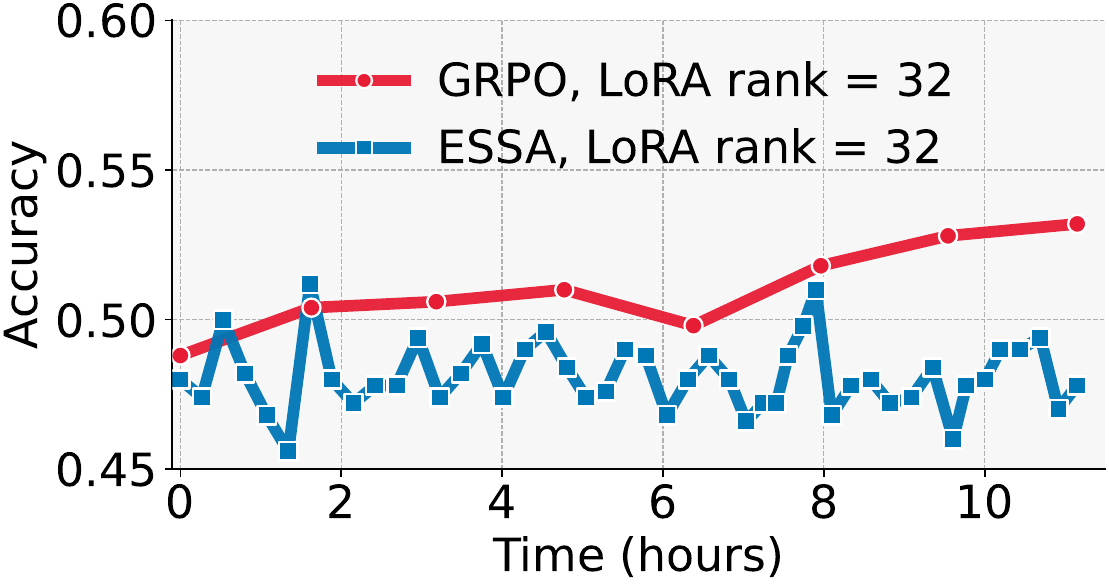}\caption{Population size 8, $\alpha\!=\!0.4$}\end{subfigure}\hfill
  \begin{subfigure}{0.32\textwidth}\includegraphics[width=\linewidth]{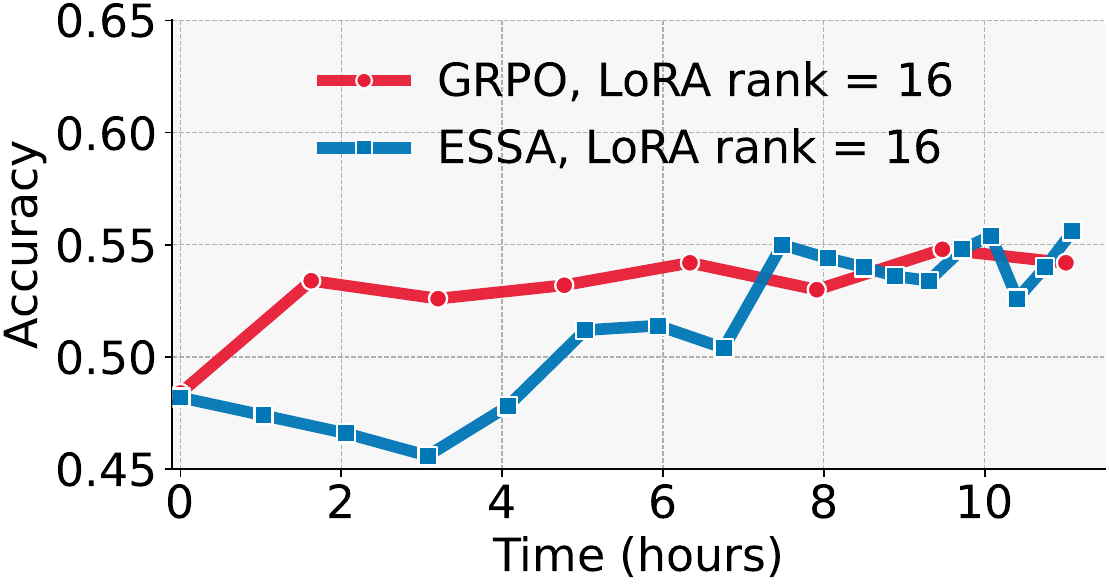}\caption{Population size 48, $\alpha\!=\!1.0$}\end{subfigure}\hfill
  \begin{subfigure}{0.32\textwidth}\includegraphics[width=\linewidth]{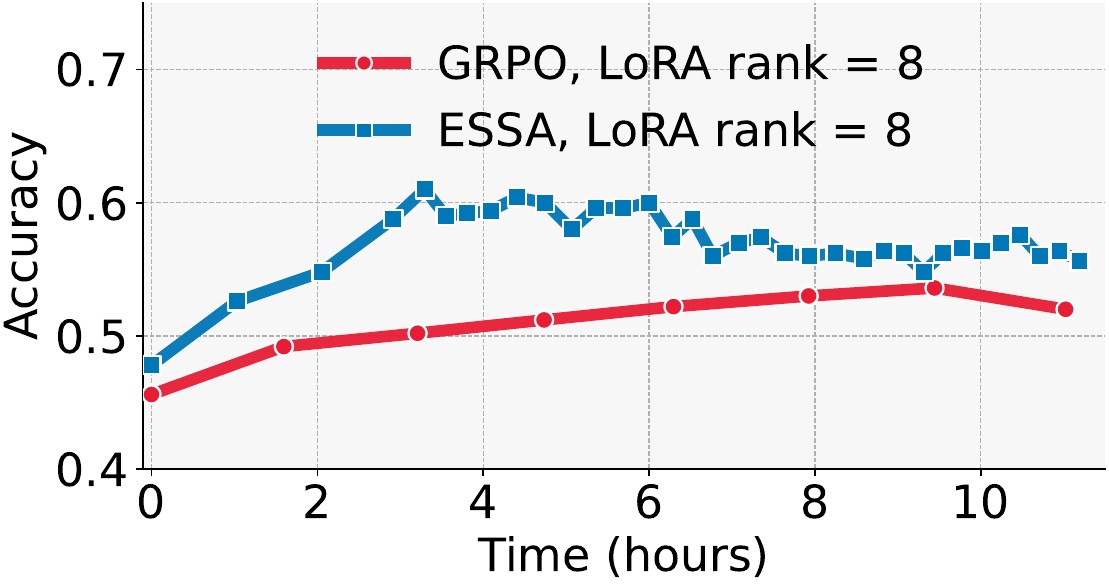}\caption{Population size 48, $\alpha\!=\!1.0$}\end{subfigure}\\[6pt]
  \hspace*{\fill}%
  \begin{subfigure}{0.32\textwidth}\includegraphics[width=\linewidth]{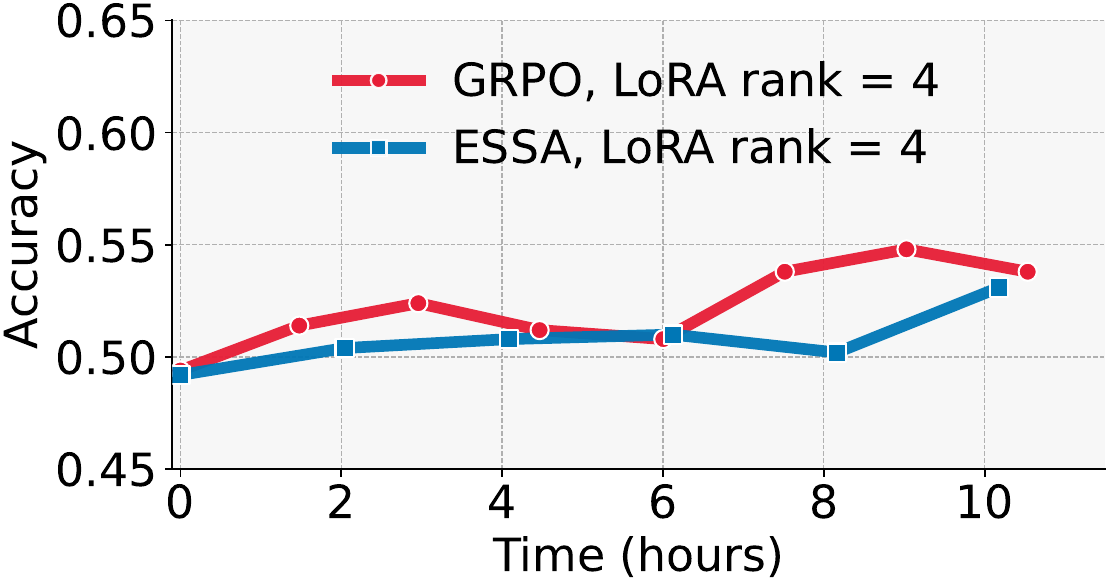}\caption{Population size 96, $\alpha\!=\!1.0$}\end{subfigure}\hfill
  \begin{subfigure}{0.32\textwidth}\includegraphics[width=\linewidth]{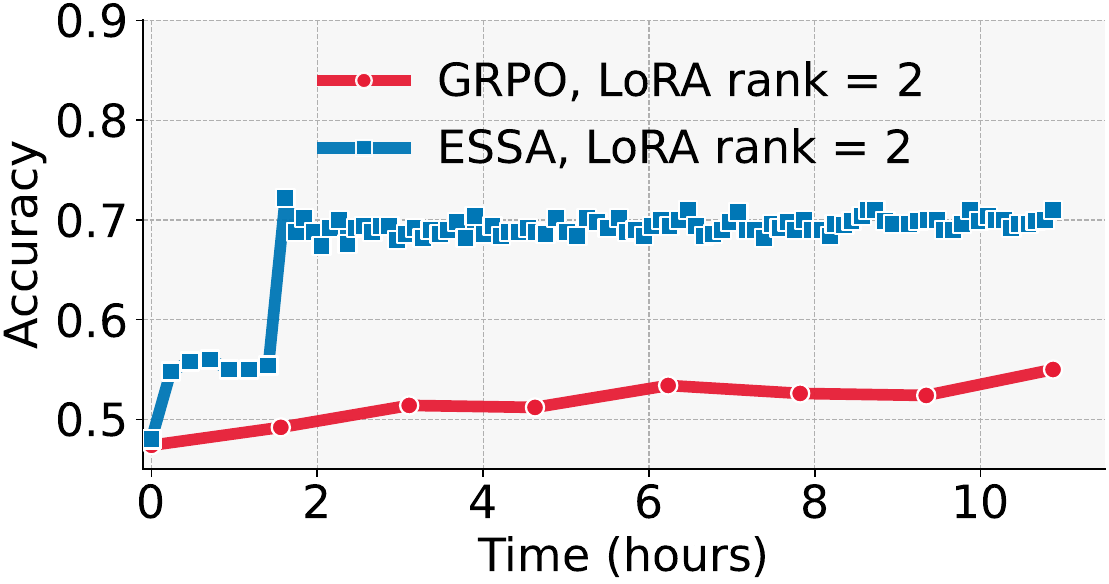}\caption{Population size 8, $\alpha\!=\!1.0$}\end{subfigure}%
  \hspace*{\fill}
  \caption{Validation accuracy over time on \textbf{PRM800K} with \textbf{Qwen2.5-Math-7B}.
  Panels (a)-(e) correspond to LoRA ranks 32, 16, 8, 4, and 2, respectively. ESSA (blue): batch size 300. LoRA-GRPO (red): lr $1\!\times\!10^{-5}$, global batch 512, mini batch 64.}
  \label{fig:train_qwen_math_prm_grpo}
\end{figure*}

\begin{figure*}[p]
  \centering
  \begin{subfigure}{0.32\textwidth}\includegraphics[width=\linewidth]{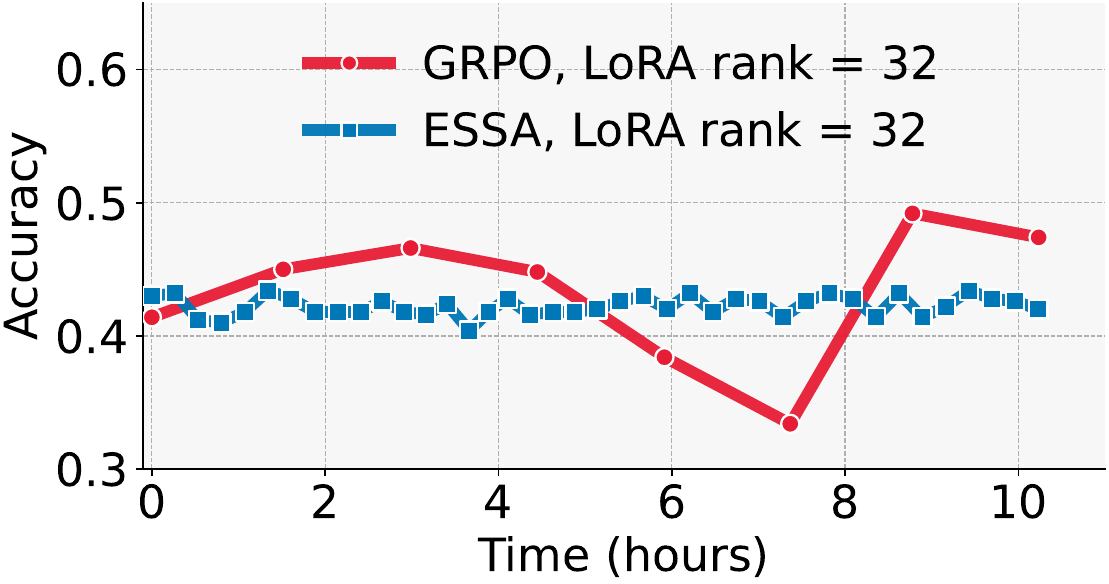}\caption{Population size 8, $\alpha\!=\!0.2$}\end{subfigure}\hfill
  \begin{subfigure}{0.32\textwidth}\includegraphics[width=\linewidth]{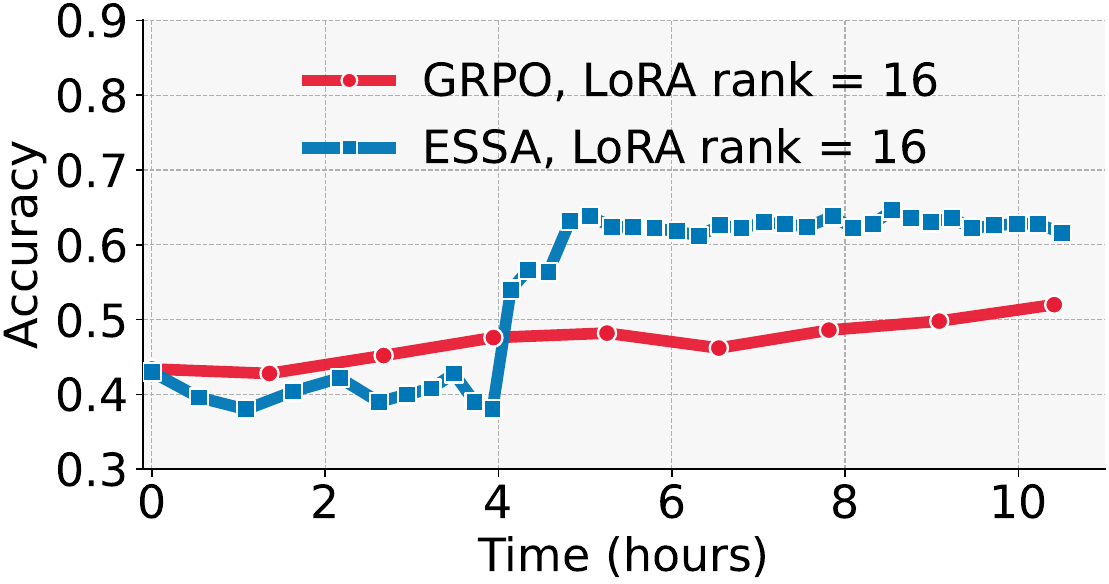}\caption{Population size 24, $\alpha\!=\!0.4$}\end{subfigure}\hfill
  \begin{subfigure}{0.32\textwidth}\includegraphics[width=\linewidth]{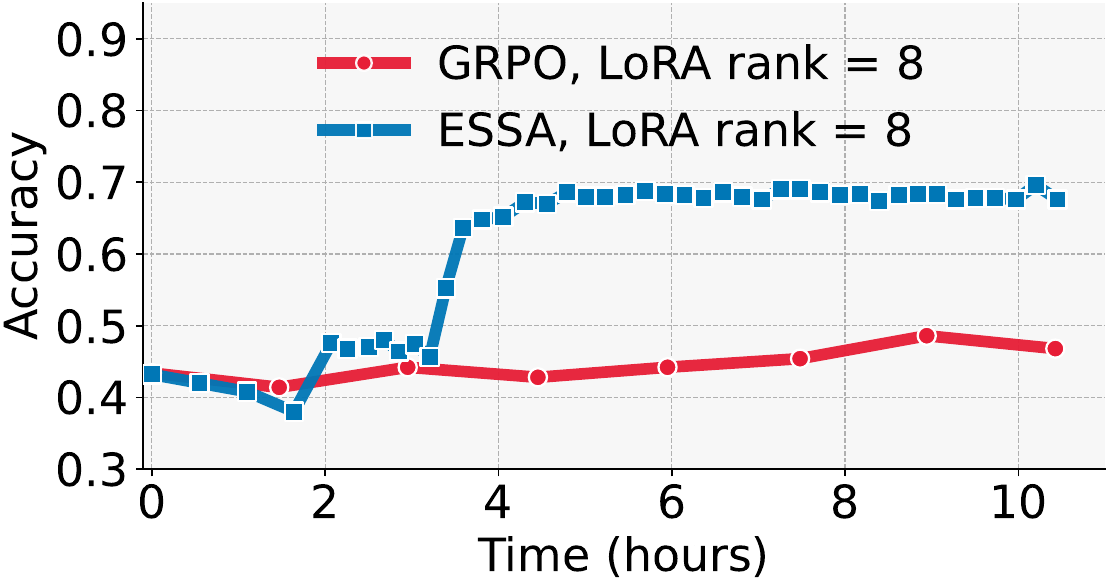}\caption{Population size 24, $\alpha\!=\!0.8$}\end{subfigure}\\[6pt]
  \hspace*{\fill}%
  \begin{subfigure}{0.32\textwidth}\includegraphics[width=\linewidth]{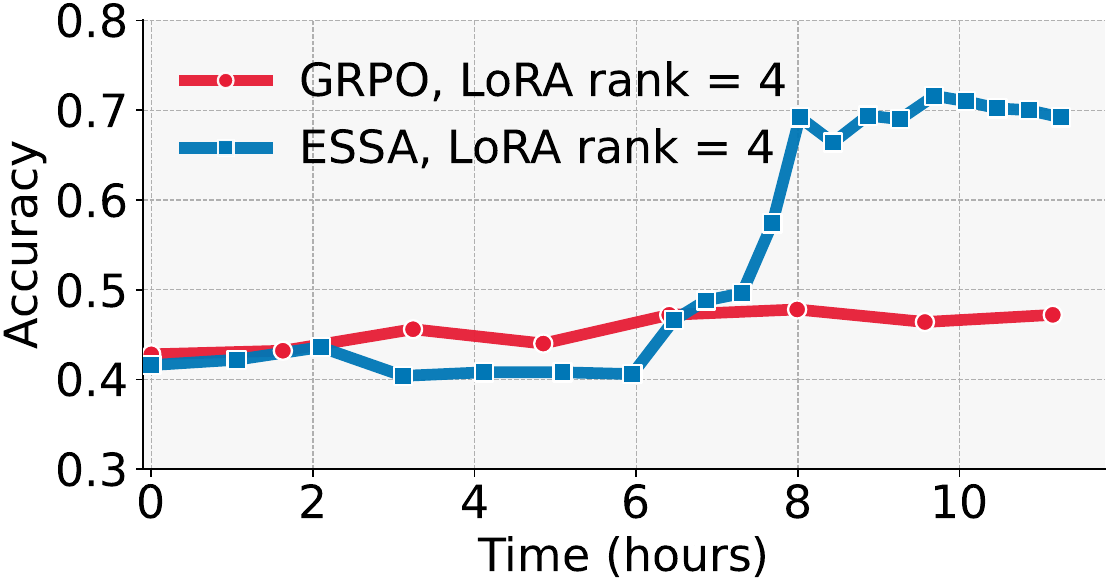}\caption{Population size 48, $\alpha\!=\!1.0$}\end{subfigure}\hfill
  \begin{subfigure}{0.32\textwidth}\includegraphics[width=\linewidth]{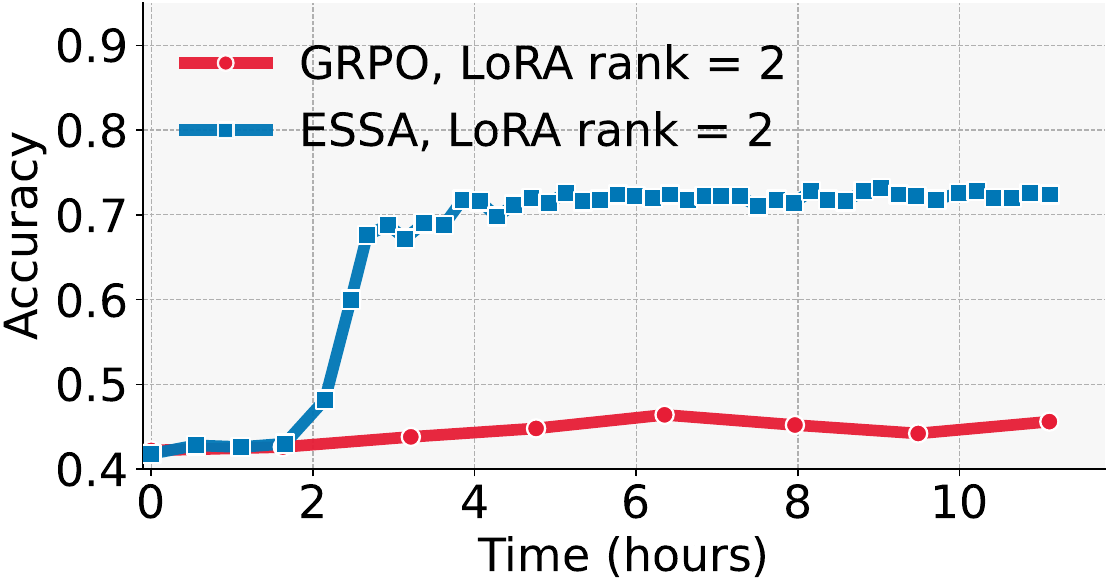}\caption{Population size 24, $\alpha\!=\!1.0$}\end{subfigure}%
  \hspace*{\fill}
  \caption{Validation accuracy over time on \textbf{PRM800K} with \textbf{Qwen2.5-7B}.
  Panels (a)-(e) correspond to LoRA ranks 32, 16, 8, 4, and 2, respectively. ESSA (blue): batch size 300. LoRA-GRPO (red): lr $1\!\times\!10^{-5}$, global batch 512, mini batch 64.}
  \label{fig:train_qwen_prm_grpo}
\end{figure*}

\begin{figure*}[p]
  \centering
  \begin{subfigure}{0.32\textwidth}\includegraphics[width=\linewidth]{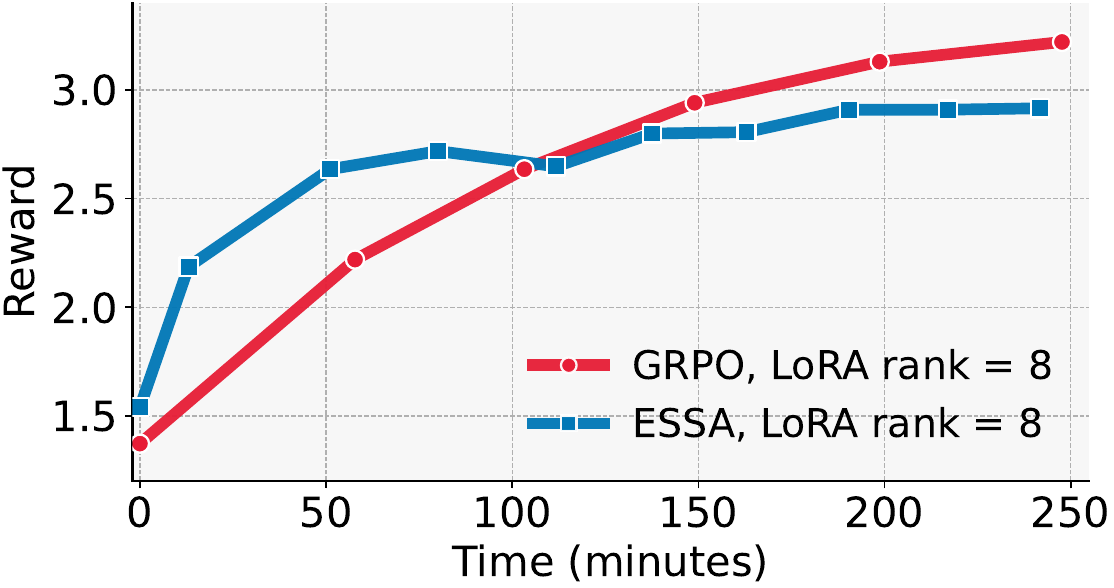}\caption{Population size 36, $\alpha\!=\!1.0$}\end{subfigure}\hfill
  \begin{subfigure}{0.32\textwidth}\includegraphics[width=\linewidth]{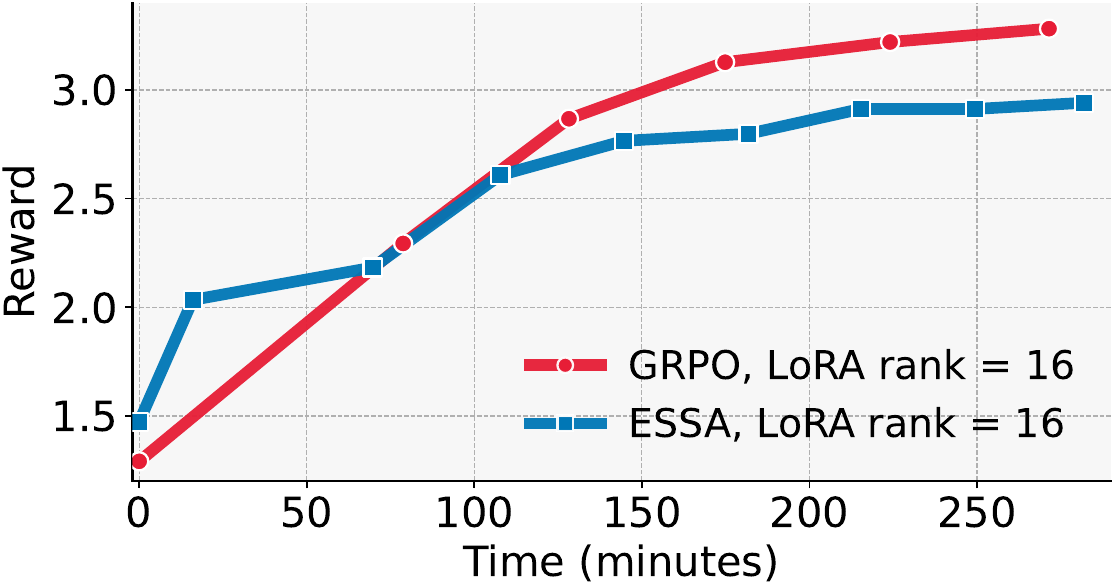}\caption{Population size 42, $\alpha\!=\!1.0$}\end{subfigure}\hfill
  \begin{subfigure}{0.32\textwidth}\includegraphics[width=\linewidth]{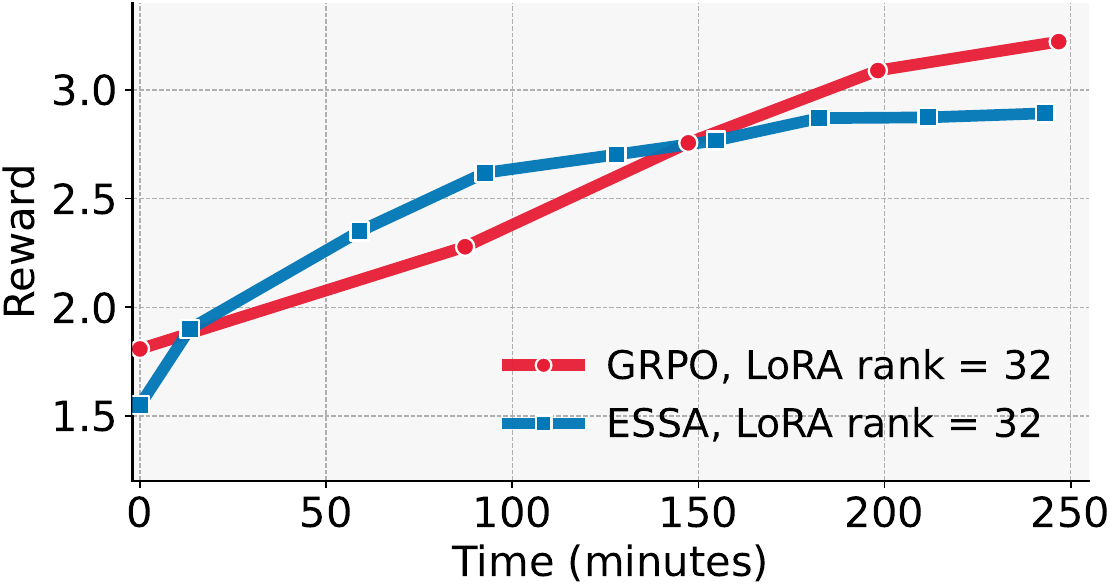}\caption{Population size 36, $\alpha\!=\!1.0$}\end{subfigure}
  \caption{Validation reward over time on \textbf{HelpSteer2} with \textbf{Llama-3.1-8B.} Panels (a)-(c) correspond to LoRA ranks 8, 16, 32 respectively. ESSA (blue): batch size 100. LoRA-GRPO (red): lr $1\!\times\!10^{-5}$, global batch 512, mini batch 64.}
  \label{fig:helpsteer}
\end{figure*}

\begin{figure*}[p]
  \centering
  \includegraphics[width=0.45\textwidth]{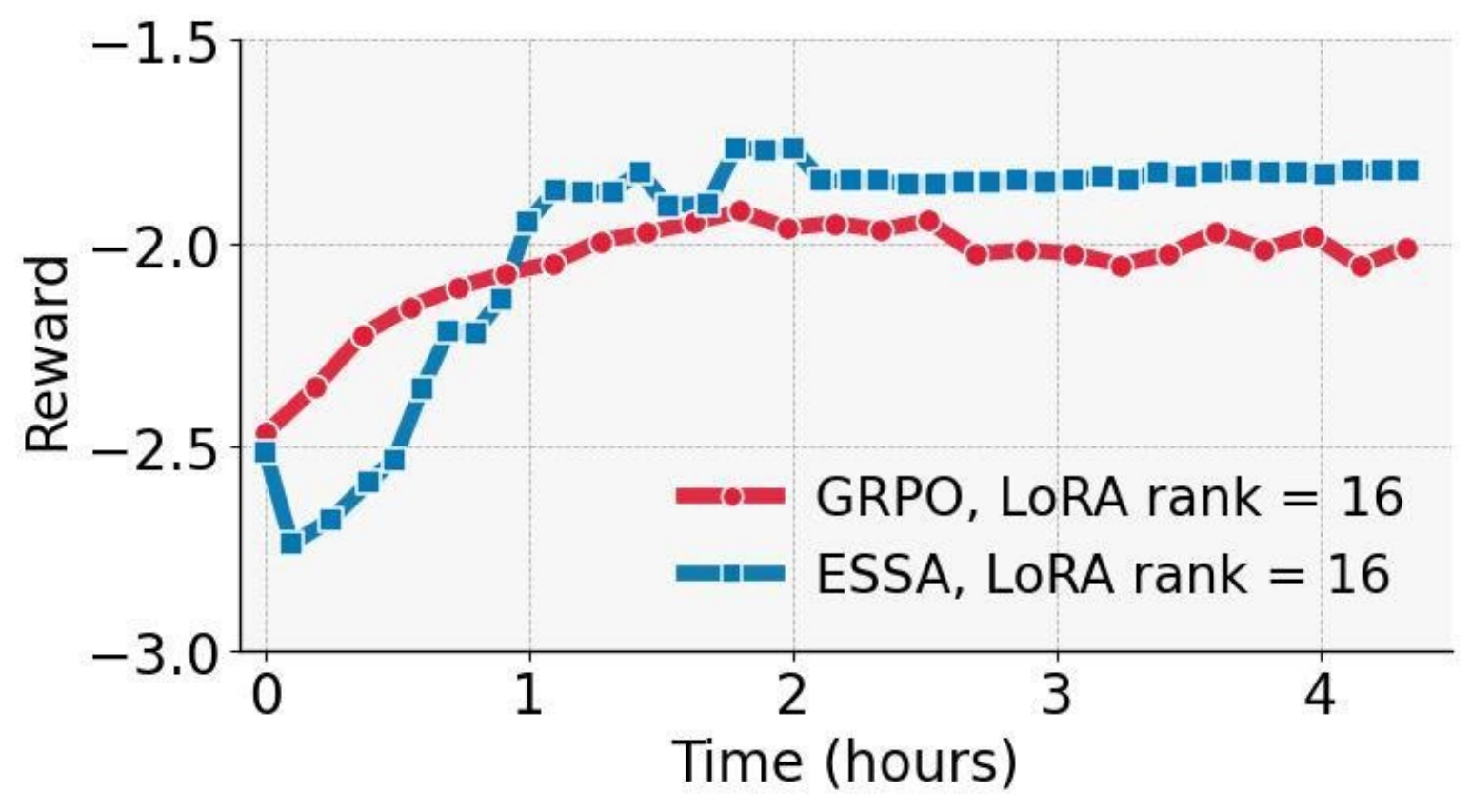}
  \caption{Validation reward over time on \textbf{HH-RLHF} with \textbf{Mistral-7B}, LoRA rank 16. ESSA (blue): population size 32, $\alpha\!=\!1.0$, batch size 100. LoRA-GRPO (red): lr $1\!\times\!10^{-5}$, global batch 512, mini batch 64.}
  \label{fig:hh_rlhf}
\end{figure*}

\clearpage

\twocolumn[
\begin{minipage}{\textwidth}
\centering

\scriptsize
\setlength{\tabcolsep}{4pt}
\resizebox{\linewidth}{!}{%
\begin{tabular}{lllllll}
\toprule
Task & Optimizer & SFT init. & Per-step candidates
& Wall-clock & $G=Q$ & Final test \\
\midrule
GSM8K
& CMA-ES (ESSA)
& 0.780 & 96 & 2 h & 441{,}600
& $0.867\pm0.010$ \\
& Plain $(\mu,\lambda)$-ES
& 0.780 & 96 & 2 h & 441{,}600
& $0.893\pm0.006$ \\
& ARS
& 0.780 & 96 & 2 h & 441{,}600
& $0.884\pm0.004$ \\
& Random search
& 0.780 & 96 & 2 h & 441{,}600
& $0.842\pm0.005$ \\
& MeZO (SPSA)
& 0.780 & 96 & 2 h & 441{,}600
& $0.826\pm0.008$ \\
\addlinespace[2pt]
PRM800K$\rightarrow$MATH500
& CMA-ES (ESSA)
& 0.420 & 96 & 8 h & 4{,}608{,}000
& $0.706\pm0.009$ \\
& Plain $(\mu,\lambda)$-ES
& 0.420 & 96 & 8 h & 4{,}608{,}000
& $0.736\pm0.008$ \\
& ARS
& 0.420 & 96 & 8 h & 4{,}608{,}000
& $0.754\pm0.006$ \\
& Random search
& 0.420 & 96 & 8 h & 4{,}608{,}000
& $0.636\pm0.004$ \\
& MeZO (SPSA)
& 0.420 & 96 & 8 h & 4{,}608{,}000
& $0.472\pm0.010$ \\
\bottomrule
\end{tabular}%
}
\captionof{table}{Alternative optimizers on Qwen2.5-7B with LoRA rank 16.
GSM8K uses population size 96 and batch size 100;
PRM800K$\rightarrow$MATH500 uses population size 96 and batch size 300.}
\label{tab:other-es-configs}
\vspace{1em}
\end{minipage}
]

\section{Optimizer and Subspace Controls}
\label{app:optimizer-controls}

This section provides additional results for the optimizer and
subspace controls presented in
Section~\ref{sec:optimizer_controls}.

\subsection{Optimizer convergence}

Table~\ref{tab:optimizer-convergence} compares validation accuracy at
selected optimization steps. CMA-ES improves fastest initially, while
ARS achieves the highest late-stage accuracy.

\subsection{Random-basis control}

We isolate the contribution of the SFT-derived singular vectors
by replacing them with random orthogonal bases. The resulting
search space is
$W_{\mathrm{SFT}} +
U_{\mathrm{rand}}\operatorname{diag}(\delta)V_{\mathrm{rand}}^\top$,
so $\delta=0$ exactly recovers the same SFT model. The search
dimension, CMA-ES configuration, and 37-step budget are held
fixed.

The random basis still improves test accuracy from 0.777 to 0.823 (Table~\ref{tab:random_basis}), showing that low-dimensional search around a strong SFT
policy already provides useful directions. The SFT-SVD basis
nevertheless converges faster and reaches higher validation and
test accuracy. Thus, the learned basis is beneficial but is not
the sole explanation for ESSA's performance.

\subsection{Sensitivity to SFT strength}
ESSA reaches similar final accuracy across all five SFT fractions (Table~\ref{tab:sft_strength}), with no monotonic relationship between SFT fraction
and final performance. The method therefore does not require a
precisely optimized SFT checkpoint, although it still assumes a
warm start containing learned structure.

\begin{table}[t]
\centering
\small
\setlength{\tabcolsep}{2.5pt}
\begin{tabular}{lccccc}
\toprule
\textbf{Optimizer}
& \textbf{21}
& \textbf{54}
& \textbf{107}
& \textbf{160}
& \textbf{214} \\
\midrule
CMA-ES
& 0.732 & 0.724 & 0.734 & 0.706 & 0.708 \\
Plain $(\mu,\lambda)$-ES
& 0.702 & 0.732 & 0.714 & 0.694 & 0.696 \\
ARS
& 0.516 & 0.682 & 0.726 & 0.728 & 0.744 \\
Random search
& 0.556 & 0.572 & 0.602 & 0.652 & 0.646 \\
MeZO (SPSA)
& 0.490 & 0.484 & 0.514 & 0.508 & 0.502 \\
\bottomrule
\end{tabular}
\caption{PRM800K validation accuracy for Qwen2.5-7B with LoRA rank 16,
population size 96, and batch size 300.}
\label{tab:optimizer-convergence}
\end{table}

\begingroup
\makeatletter
\setlength{\@fptop}{0pt}
\setlength{\@fpsep}{6pt}
\setlength{\@fpbot}{0pt plus 1fil}
\makeatother
\setlength{\floatsep}{6pt}
\setlength{\textfloatsep}{8pt}

\begin{table}[t]
\centering
\small
\setlength{\tabcolsep}{3pt}
\resizebox{\columnwidth}{!}{%
\begin{tabular}{lccc}
\toprule
\textbf{Basis}
& \textbf{SFT test}
& \textbf{Best val.}
& \textbf{Test @ best val.} \\
\midrule
SFT-SVD
& 0.777 & \textbf{0.916} & \textbf{0.838} \\
Random orthogonal
& 0.777 & 0.884 & 0.823 \\
\bottomrule
\end{tabular}%
}
\caption{Random-basis control on Qwen2.5-7B/GSM8K with
LoRA rank 8, population size 96, and a matched 37-step budget.}
\label{tab:random_basis}
\end{table}

\begin{table}[t]
\centering
\small
\setlength{\tabcolsep}{2.5pt}
\begin{tabular}{ccc}
\toprule
\textbf{SFT fraction}
& \textbf{SFT only}
& \textbf{SFT$\rightarrow$ESSA} \\
\midrule
5\%   & 0.501 & $0.710 \pm 0.016$ \\
25\%  & 0.447 & $0.699 \pm 0.023$ \\
50\%  & 0.395 & $\mathbf{0.728 \pm 0.024}$ \\
75\%  & 0.475 & $0.710 \pm 0.014$ \\
100\% & 0.458 & $0.721 \pm 0.023$ \\
\bottomrule
\end{tabular}
\caption{MATH500 accuracy as a function of the fraction of
PRM800K SFT data used to construct the warm start.}
\label{tab:sft_strength}
\end{table}

\clearpage
\endgroup

\clearpage

\clearpage

\begingroup
\raggedbottom

\setlength{\abovedisplayskip}{5pt}
\setlength{\belowdisplayskip}{5pt}
\setlength{\abovedisplayshortskip}{3pt}
\setlength{\belowdisplayshortskip}{3pt}

\section{Experimental Configuration and Implementation Details}
\label{app:implementation}

\paragraph{Data splits.}
Table~\ref{tab:data_splits} summarizes the training,
selection-validation, and final-test splits.

\paragraph{Main experimental configurations.}
Table~\ref{tab:main-reported-configs} reports the configurations and
compute accounting for the reported experiments. The run-level
learning-rate sweeps for the main-text LoRA-GRPO comparisons on GSM8K
and IFEval are reported in
Tables~\ref{tab:grpo-run-level-gsm-ranks} and
\ref{tab:grpo-run-level-ifeval}, respectively. The SVD-GRPO sweeps for
the optimizer ablation on PRM800K$\rightarrow$MATH500 are reported in
Table~\ref{tab:grpo-run-level-prm}. These experiments use the
nine-value learning-rate set
\[
\begin{aligned}
\mathcal{L}=\{&10^{-6},5\!\times\!10^{-6},10^{-5},\\
&5\!\times\!10^{-5},10^{-4},5\!\times\!10^{-4},\\
&10^{-3},5\!\times\!10^{-3},10^{-2}\}.
\end{aligned}
\]
Each generated response is scored once, so the policy-generation count
$G$ equals the reward-query count $Q$ and is reported jointly as
$G=Q$. For the SFT-seeded methods, wall-clock budgets exclude the
shared SFT stage, whose cost is reported separately in
Table~\ref{tab:sft_hrs}.

\paragraph{SFT cost.}
SFT is shared by ESSA and the post-SFT LoRA baselines. Full-parameter GRPO in Table~\ref{tab:main_compare} is initialized without SFT and is therefore excluded
from this shared-cost statement. Table~\ref{tab:sft_hrs} reports the
SFT GPU-hours and the initial held-out metric observed before online
optimization.

\paragraph{System prompts.}
\begin{itemize}
  \raggedright
  \setlength{\itemsep}{1pt}
  \setlength{\parskip}{0pt}
  \setlength{\parsep}{0pt}

  \item \textbf{Math reasoning:}
  \texttt{Please reason step by step and place your final answer in
  \textbackslash boxed\{\}.}

  \item \textbf{Countdown:}
  \texttt{You are a helpful assistant.}

  \item \textbf{Instruction following:}
  \texttt{You are a helpful, honest, and concise assistant.}
\end{itemize}

\paragraph{ESSA hyperparameter grid.}
We evaluate ESSA using task-specific subsets of the hyperparameter
values reported in Table~\ref{tab:essa_grid}; not every Cartesian
combination is evaluated. The CMA-ES initial step size is fixed at
$\sigma_0=0.32$ in all experiments.

\paragraph{Software.}
We train the gradient baselines--LoRA-GRPO, Online DPO, PPO,
full-parameter GRPO, and SVD-GRPO--using the
\texttt{verl} library~\citep{verl}. ESSA uses
vLLM~\citep{vllm} for policy inference and EvoTorch for CMA-ES.
Unless otherwise noted, the experiments use NVIDIA H100-equivalent
GPUs.
\clearpage
\endgroup

\providecommand{\sweeptbd}{\textit{TBD}}
\providecommand{\notselected}{\textit{TBD}}

\begin{table*}[t]
\centering
\small
\setlength{\tabcolsep}{5pt}
\resizebox{\textwidth}{!}{%
\begin{tabular}{lrrrr}
\toprule
\textbf{Dataset}
& \textbf{SFT train}
& \textbf{Online train}
& \textbf{Selection val}
& \textbf{Final test} \\
\midrule

GSM8K
& 1.5K
& 5.5K
& 500 from GSM8K train
& GSM8K test (1,319) \\

PRM800K $\rightarrow$ MATH500
& 3.6K
& 7.4K
& 500 from PRM800K
& MATH500 (500) \\

IFEval-like $\rightarrow$ IFEval
& 1.7K
& 2.9K
& 500 from source train
& IFEval (541) \\

\midrule

HelpSteer2
& 19K
& 1.3K
& 500
& 500 \\

HH-RLHF
& 5K
& 10K
& 1.0K
& held-out judge set \\

Countdown
& ---
& 1.0K
& 1.0K
& 1.0K \\

\bottomrule
\end{tabular}%
}
\caption{Training, selection, and final-evaluation
splits. Online-train counts exclude selection-validation
examples. For GSM8K and IFEval, the official benchmark
sets are never used for model or hyperparameter selection.
For PRM800K$\rightarrow$MATH500, selection is performed
on 500 held-out PRM800K problems and MATH500 is used
only for final reporting.}
\label{tab:data_splits}
\end{table*}

\begin{table*}[p]
\centering
\scriptsize
\setlength{\tabcolsep}{4pt}
\renewcommand{\arraystretch}{1.02}

\begin{tabularx}{\textwidth}{
  @{}
  p{0.27\textwidth}
  c
  X
  c
  c
  @{}
}
\toprule
\textbf{Backbone (task)}
& \textbf{LoRA rank}
& \textbf{Reported in}
& \textbf{SFT epochs}
& \textbf{SFT GPU-hrs} \\
\midrule

Qwen2.5-7B (GSM8K)
& 2
& Sec.~\ref{sec:main_comparison};
  Apps.~\ref{app:hyperparams} and~\ref{app:school}
& 1 & 0.10 \\
& 4  & & 1 & 0.15 \\
& 8  & & 1 & 0.16 \\
& 16 & & 1 & 0.18 \\
& 32 & & 1 & 0.21 \\

\addlinespace[2pt]

Qwen2.5-Math-7B (GSM8K)
& 2
& Apps.~\ref{app:hyperparams} and~\ref{app:school}
& 1 & 0.10 \\
& 4  & & 1 & 0.12 \\
& 8  & & 1 & 0.15 \\
& 16 & & 1 & 0.16 \\
& 32 & & 1 & 0.19 \\

\addlinespace[2pt]

Qwen2.5-7B (PRM800K)
& 2
& Sec.~\ref{sec:optimizer_controls};
  Apps.~\ref{app:hyperparams} and~\ref{app:advanced_math}
& 1 & 0.31 \\
& 4  & & 1 & 0.35 \\
& 8  & & 1 & 0.38 \\
& 16 & & 1 & 0.43 \\
& 32 & & 1 & 0.48 \\

\addlinespace[2pt]

Qwen2.5-Math-7B (PRM800K)
& 2
& Apps.~\ref{app:hyperparams} and~\ref{app:advanced_math}
& 1 & 0.28 \\
& 4  & & 1 & 0.32 \\
& 8  & & 1 & 0.35 \\
& 16 & & 1 & 0.40 \\
& 32 & & 1 & 0.44 \\

\addlinespace[2pt]

Qwen2.5-32B (PRM800K)
& 8
& Secs.~\ref{sec:scale_math} and~\ref{sec:sensitivity}
& 3 & 9.80 \\
& 16
& Sec.~\ref{sec:scaling}
& 3 & 11.24 \\

\addlinespace[2pt]

Qwen2.5-72B (PRM800K)
& 4
& Sec.~\ref{sec:scale_math}
& 3 & 51.97 \\

\addlinespace[2pt]

Llama-3.1-8B (IFEval-like)
& 2
& Secs.~\ref{sec:main_comparison} and~\ref{sec:nlp};
  App.~\ref{app:hyperparams}
& 1 & 0.35 \\
& 4  & & 1 & 0.41 \\
& 8  & & 1 & 0.45 \\
& 16 & & 1 & 0.50 \\
& 32 & & 1 & 0.56 \\

\addlinespace[2pt]

Llama-3.1-8B (HelpSteer2)
& 8
& Sec.~\ref{sec:nlp}; App.~\ref{app:helpsteer}
& 3 & 3.38 \\
& 16 & & 3 & 3.74 \\
& 32 & & 3 & 4.01 \\

\addlinespace[2pt]

Mistral-7B (HH-RLHF)
& 16
& Sec.~\ref{sec:nlp}; App.~\ref{app:hh_rlhf}
& 3 & 3.24 \\

\bottomrule
\end{tabularx}

\caption{SFT cost for the checkpoints used in the reported experiments.
GPU-hours are measured on NVIDIA H100 80GB GPUs and reported separately
for each LoRA rank. For every referenced SFT-seeded training curve, the
first plotted point is the validation performance of the corresponding
SFT checkpoint.}
\label{tab:sft_hrs}
\end{table*}

\begin{table*}[t]
\centering
\footnotesize
\setlength{\tabcolsep}{5pt}
\renewcommand{\arraystretch}{1.08}

\begin{tabular*}{\textwidth}{
  @{\extracolsep{\fill}}
  lll
  @{}
}
\toprule
\textbf{Parameter}
& \textbf{Values evaluated}
& \textbf{Default} \\
\midrule

LoRA rank ($r$)
& 2, 4, 8, 16, 32
& 8 \\

Population size ($P$)
& 8, 24, 32, 36, 42, 48, 64, 96, 128, 192, 400
& 96 \\

Retained fraction ($\alpha$)
& 0.1, 0.2, 0.4, 0.6, 0.8, 1.0
& 1.0 \\

Per-candidate batch size ($B_{\mathrm{E}}$)
& 100, 200, 256, 300, 500
& 100 \\

Weight precision
& BF16, INT8, INT4
& BF16 \\

CMA-ES initial step size ($\sigma_0$)
& 0.32
& 0.32 \\

\bottomrule
\end{tabular*}

\caption{ESSA hyperparameter values used or evaluated across experiments.
The listed values form the union of task-specific configurations; not
every Cartesian combination was evaluated. Experiments use up to
128 NVIDIA H100 80GB GPUs.}
\label{tab:essa_grid}
\end{table*}

\clearpage

\begin{table*}[t]
\centering

\scriptsize
\setlength{\tabcolsep}{2.4pt}
\renewcommand{\arraystretch}{1.06}

\resizebox{\textwidth}{!}{%
\begin{tabular}{lllrlrlrll}
\toprule
Source & Model / task & Method & GPUs & Configuration
& SFT init. & Wall-clock & $G=Q$ & Reported result & Prec. \\
\midrule

Tab.~1 & Qwen2.5-7B / GSM8K
& ESSA & 8 & $16/96/100/1.0$ & 0.723 & 2 h & 489{,}600
& $0.867\pm0.010$ & BF16 \\
& & LoRA-GRPO & 8 & $16/1\!\times\!10^{-4}/512/64/8$
& 0.723 & 2 h & 184{,}320
& $0.823\pm0.012$ & BF16 \\
& & Online DPO & 8 & $16/5\!\times\!10^{-5}/512/64/8$
& 0.723 & 2 h & 401{,}408
& $0.818\pm0.001$ & BF16 \\
& & PPO & 8 & $16/1\!\times\!10^{-4}/512/64/1$
& 0.723 & 2 h & 15{,}360
& $0.791\pm0.010$ & BF16 \\
& & Full-parameter GRPO & 8 & $\text{--}/1\!\times\!10^{-5}/512/64/8$
& -- & 2 h & 94{,}208
& $0.870\pm0.009$ & BF16 \\

\addlinespace[2pt]
Tab.~1 & Llama-3.1-8B / IFEval
& ESSA & 8 & $8/24/500/1.0$ & 0.451 & 4 h & 1{,}764{,}000
& $0.555\pm0.009$ & BF16 \\
& & LoRA-GRPO & 8 & $8/5\!\times\!10^{-5}/512/64/8$
& 0.451 & 4 h & 122{,}880
& $0.483\pm0.002$ & BF16 \\
& & Online DPO & 8 & $8/1\!\times\!10^{-5}/512/64/8$
& 0.451 & 4 h & 602{,}112
& $0.559\pm0.008$ & BF16 \\
& & PPO & 8 & $8/1\!\times\!10^{-4}/512/64/1$
& 0.451 & 4 h & 51{,}200
& $0.559\pm0.007$ & BF16 \\
& & Full-parameter GRPO & 8 & $\text{--}/1\!\times\!10^{-5}/512/64/8$
& -- & 4 h & 143{,}360
& $0.692\pm0.008$ & BF16 \\

\addlinespace[2pt]
Fig.~2 & Qwen2.5-7B / GSM8K
& ESSA & 8 & $8/48/100/0.8$ & 0.724 & 2 h & 201{,}600
& $0.853\pm0.008$ & BF16 \\
& & LoRA-GRPO & 8 & $8/1\!\times\!10^{-4}/512/64/8$
& 0.724 & 2 h & 204{,}800
& $0.847\pm0.014$ & BF16 \\

\addlinespace[2pt]
Fig.~3 & Qwen2.5-7B / PRM800K $\rightarrow$ MATH500
& ESSA & 8 & $2/24/300/1.0$ & 0.420 & 8 h & 1{,}080{,}000
& $0.723\pm0.005$ & BF16 \\
& & SVD-GRPO & 8 & $2/1\!\times\!10^{-2}/512/64/8$
& 0.420 & 8 h & 204{,}800
& $0.422\pm0.008$ & BF16 \\
& & SVD-GRPO & 8 & $4/1\!\times\!10^{-2}/512/64/8$
& 0.420 & 8 h & 245{,}760
& $0.322\pm0.012$ & BF16 \\
& & SVD-GRPO & 8 & $8/1\!\times\!10^{-2}/512/64/8$
& 0.420 & 8 h & 245{,}760
& $0.488\pm0.001$ & BF16 \\
& & SVD-GRPO & 8 & $16/1\!\times\!10^{-2}/512/64/8$
& 0.420 & 8 h & 245{,}760
& $0.510\pm0.002$ & BF16 \\

\addlinespace[2pt]
Fig.~4 & Qwen2.5-32B / PRM800K $\rightarrow$ MATH500
& ESSA & 16 & $16/128/256/1.0$ & 0.631 & 206.17 min & 622{,}592
& $0.835$ target & BF16 \\
& & LoRA-GRPO & 16 & $16/1\!\times\!10^{-5}/512/64/8$
& 0.631 & 394.36 min & 204{,}800
& $0.835$ target & BF16 \\

Fig.~4 & Qwen2.5-32B / PRM800K $\rightarrow$ MATH500
& ESSA & 32 & $16/128/256/1.0$ & 0.631 & 110.71 min & 622{,}592
& $0.835$ target & BF16 \\
& & LoRA-GRPO & 32 & $16/1\!\times\!10^{-5}/512/64/8$
& 0.631 & 210.29 min & 204{,}800
& $0.835$ target & BF16 \\

Fig.~4 & Qwen2.5-32B / PRM800K $\rightarrow$ MATH500
& ESSA & 64 & $16/128/256/1.0$ & 0.631 & 30.62 min & 622{,}592
& $0.835$ target & BF16 \\
& & LoRA-GRPO & 64 & $16/1\!\times\!10^{-5}/512/64/8$
& 0.631 & 177.82 min & 204{,}800
& $0.835$ target & BF16 \\

Fig.~4 & Qwen2.5-32B / PRM800K $\rightarrow$ MATH500
& ESSA & 128 & $16/128/256/1.0$ & 0.631 & 19.72 min & 622{,}592
& $0.835$ target & BF16 \\
& & LoRA-GRPO & 128 & $16/1\!\times\!10^{-5}/512/64/8$
& 0.631 & 153.04 min & 204{,}800
& $0.835$ target & BF16 \\

\addlinespace[2pt]
Fig.~5 & Llama-3.1-8B / IFEval
& ESSA & 8 & $8/24/500/1.0$ & 0.451 & 4 h & 1{,}764{,}000
& $0.555\pm0.009$ & BF16 \\
& & LoRA-GRPO & 8 & $8/5\!\times\!10^{-5}/512/64/8$
& 0.451 & 4 h & 122{,}880
& $0.483\pm0.002$ & BF16 \\

\addlinespace[2pt]
Tab.~4 & Qwen2.5-32B / advanced math
& ESSA & 8 & $8/48/256/1.0$ & 0.631 & 8 h & 614{,}400
& $0.440$ & BF16 \\
& & LoRA-GRPO & 8 & $16/1\!\times\!10^{-5}/512/64/8$
& 0.631 & 8 h & 204{,}800
& $0.425$ & BF16 \\

\addlinespace[2pt]
Tab.~5 & Qwen2.5-72B / advanced math
& ESSA & 8 & $4/64/256/1.0$ & 0.678 & 20 h & 327{,}680
& $0.447$ & INT4 \\
& & LoRA-GRPO & 8 & $16/1\!\times\!10^{-5}/512/64/8$
& 0.678 & 20 h & 61{,}440
& $0.445$ & BF16 \\

\addlinespace[2pt]
Tab.~6 & Qwen2.5-32B / MATH500 precision
& ESSA (BF16) & 8 & $8/64/256/1.0$ & 0.631 & 8 h & 147{,}456
& $0.845\pm0.004$ & BF16 \\
& & ESSA (INT8) & 8 & $8/64/256/1.0$ & 0.631 & 8 h & 229{,}376
& $0.841\pm0.006$ & INT8 \\
& & ESSA (INT4) & 8 & $8/64/256/1.0$ & 0.631 & 8 h & 196{,}608
& $0.834\pm0.006$ & INT4 \\

\addlinespace[2pt]
Fig.~35(a) & Llama-3.1-8B / HelpSteer2
& ESSA & 8 & $8/36/100/1.0$ & 1.815 reward & 4 h & 144{,}000
& 2.909 reward & BF16 \\
& & LoRA-GRPO & 8 & $8/1\!\times\!10^{-5}/512/64/8$
& 1.815 reward & 4 h & 163{,}840
& 3.130 reward & BF16 \\

\addlinespace[2pt]
Fig.~35(b) & Llama-3.1-8B / HelpSteer2
& ESSA & 8 & $16/42/100/1.0$ & 1.815 reward & 4 h & 126{,}000
& 2.913 reward & BF16 \\
& & LoRA-GRPO & 8 & $16/1\!\times\!10^{-5}/512/64/8$
& 1.815 reward & 4 h & 163{,}840
& 3.220 reward & BF16 \\

\addlinespace[2pt]
Fig.~35(c) & Llama-3.1-8B / HelpSteer2
& ESSA & 8 & $32/36/100/1.0$ & 1.815 reward & 4 h & 126{,}000
& 2.874 reward & BF16 \\
& & LoRA-GRPO & 8 & $32/1\!\times\!10^{-5}/512/64/8$
& 1.815 reward & 4 h & 122{,}880
& 3.089 reward & BF16 \\

\addlinespace[2pt]
Fig.~36 & Mistral-7B / HH-RLHF
& ESSA & 8 & $16/32/100/1.0$ & $-2.500$ reward & 4 h
& 729{,}600 & $12/76/12$ & BF16 \\
& & LoRA-GRPO & 8 & $16/1\!\times\!10^{-5}/512/64/8$
& $-2.500$ reward & 4 h & 294{,}912
& \textit{pairwise result above} & BF16 \\

\bottomrule
\end{tabular}%
}
\caption{Configurations and accounting for the reported main
experiments. Within each experiment, ESSA is listed first, followed by
the corresponding baselines. ESSA configurations are reported as
$r/P/B_{\mathrm{E}}/\alpha$, where $r$ is the LoRA rank, $P$ is the
population size, $B_{\mathrm{E}}$ is the per-candidate batch size, and
$\alpha$ is the fraction of singular-value coordinates optimized.
Gradient-based configurations are reported as
$r/\eta/B_{\mathrm{G}}/b/n$, where $\eta$ is the learning rate,
$B_{\mathrm{G}}$ is the global batch size, $b$ is the minibatch size,
and $n$ is the number of responses sampled per prompt. The displayed
$\eta$ is the learning rate of the reported run; all nine sweep values
are listed in the run-level tables. $G$ denotes the number of training
policy generations and $Q$ the number of reward queries; because each
generated response is scored once, $G=Q$. The GPUs column
reports the number of GPUs used: all non-scaling experiments use eight
GPUs, while the Figure~\ref{fig:scaling} rows report 16, 32, 64, and
128 GPUs. SFT init.\ denotes the result
of the shared supervised-fine-tuning initialization, and Prec.\ denotes
the numerical precision. For HelpSteer2, the reported result is the
validation reward at the last checkpoint within the four-hour cutoff,
and $G=Q$ is truncated at the same checkpoint. For
Mistral-7B/HH-RLHF, the reported result is the GPT-4o pairwise outcome
in the order ESSA win / tie / LoRA-GRPO win.}
\label{tab:main-reported-configs}
\end{table*}

\clearpage

\begin{table*}[p]
\centering

\footnotesize
\setlength{\tabcolsep}{4pt}
\renewcommand{\arraystretch}{1.05}

\resizebox{\textwidth}{!}{%
\begin{tabular}{llcccc}
\toprule
\textbf{Experiment}
& \textbf{Method / rank}
& \textbf{LR}
& \textbf{SFT initial test}
& \textbf{Wall-clock}
& \textbf{Final test} \\
\midrule

Qwen2.5-7B / GSM8K
& LoRA-GRPO, $r=8$
& $10^{-6}$
& 0.724
& 2 h
& $0.746 \pm 0.004$ \\

&
& $5\!\times\!10^{-6}$
& 0.724
& 2 h
& $0.755 \pm 0.003$ \\

&
& $10^{-5}$
& 0.724
& 2 h
& $0.832 \pm 0.010$ \\

&
& $5\!\times\!10^{-5}$
& 0.724
& 2 h
& $0.839 \pm 0.008$ \\

&
& $10^{-4}$
& 0.724
& 2 h
& $0.847 \pm 0.014$ \\

&
& $5\!\times\!10^{-4}$
& 0.724
& 2 h
& $0.844 \pm 0.012$ \\

&
& $10^{-3}$
& 0.724
& 2 h
& $0.842 \pm 0.010$ \\

&
& $5\!\times\!10^{-3}$
& 0.724
& 2 h
& $0.000 \pm 0.000$ \\

&
& $10^{-2}$
& 0.724
& 2 h
& $0.000 \pm 0.000$ \\

\addlinespace[4pt]

Qwen2.5-7B / GSM8K
& LoRA-GRPO, $r=16$
& $10^{-6}$
& 0.723
& 2 h
& $0.728 \pm 0.006$ \\

&
& $5\!\times\!10^{-6}$
& 0.723
& 2 h
& $0.737 \pm 0.012$ \\

&
& $10^{-5}$
& 0.723
& 2 h
& $0.789 \pm 0.005$ \\

&
& $5\!\times\!10^{-5}$
& 0.723
& 2 h
& $0.801 \pm 0.006$ \\

&
& $10^{-4}$
& 0.723
& 2 h
& $0.823 \pm 0.012$ \\

&
& $5\!\times\!10^{-4}$
& 0.723
& 2 h
& $0.805 \pm 0.015$ \\

&
& $10^{-3}$
& 0.723
& 2 h
& $0.814 \pm 0.007$ \\

&
& $5\!\times\!10^{-3}$
& 0.723
& 2 h
& $0.000 \pm 0.000$ \\

&
& $10^{-2}$
& 0.723
& 2 h
& $0.000 \pm 0.000$ \\

\bottomrule
\end{tabular}%
}
\caption{LoRA-GRPO learning-rate sweep results for Qwen2.5-7B on GSM8K.
For each run, we report the test-set performance of the checkpoint from
the final validation evaluation within the stated wall-clock budget.}
\label{tab:grpo-run-level-gsm-ranks}
\end{table*}


\begin{table*}[p]
\centering

\footnotesize
\setlength{\tabcolsep}{4pt}
\renewcommand{\arraystretch}{1.05}

\resizebox{\textwidth}{!}{%
\begin{tabular}{llcccc}
\toprule
\textbf{Experiment}
& \textbf{Method / rank}
& \textbf{LR}
& \textbf{SFT initial test}
& \textbf{Wall-clock}
& \textbf{Final test} \\
\midrule

Llama-3.1-8B / IFEval
& LoRA-GRPO, $r=8$
& $10^{-6}$
& 0.451
& 4 h
& $0.433 \pm 0.011$ \\

&
& $5\!\times\!10^{-6}$
& 0.451
& 4 h
& $0.446 \pm 0.009$ \\

&
& $10^{-5}$
& 0.451
& 4 h
& $0.476 \pm 0.002$ \\

&
& $5\!\times\!10^{-5}$
& 0.451
& 4 h
& $0.483 \pm 0.002$ \\

&
& $10^{-4}$
& 0.451
& 4 h
& $0.447 \pm 0.012$ \\

&
& $5\!\times\!10^{-4}$
& 0.451
& 4 h
& $0.458 \pm 0.008$ \\

&
& $10^{-3}$
& 0.451
& 4 h
& $0.421 \pm 0.015$ \\

&
& $5\!\times\!10^{-3}$
& 0.451
& 4 h
& $0.000 \pm 0.000$ \\

&
& $10^{-2}$
& 0.451
& 4 h
& $0.000 \pm 0.000$ \\

\bottomrule
\end{tabular}%
}
\caption{LoRA-GRPO learning-rate sweep results for Llama-3.1-8B on IFEval.
For each run, we report the test-set performance of the checkpoint from
the final validation evaluation within the stated wall-clock budget.}
\label{tab:grpo-run-level-ifeval}
\end{table*}


\begin{table*}[p]
\centering

\tiny
\setlength{\tabcolsep}{3pt}
\renewcommand{\arraystretch}{1.02}

\resizebox{\textwidth}{!}{%
\begin{tabular}{llcccc}
\toprule
\textbf{Experiment}
& \textbf{Method / rank}
& \textbf{LR}
& \textbf{SFT initial test}
& \textbf{Wall-clock}
& \textbf{Final test} \\
\midrule

Qwen2.5-7B / PRM800K$\rightarrow$MATH500
& SVD-GRPO, $r=2$
& $10^{-6}$
& 0.398
& 8 h
& $0.000 \pm 0.000$ \\

&
& $5\!\times\!10^{-6}$
& 0.398
& 8 h
& $0.000 \pm 0.000$ \\

&
& $10^{-5}$
& 0.398
& 8 h
& $0.000 \pm 0.000$ \\

&
& $5\!\times\!10^{-5}$
& 0.398
& 8 h
& $0.308 \pm 0.016$ \\

&
& $10^{-4}$
& 0.398
& 8 h
& $0.239 \pm 0.012$ \\

&
& $5\!\times\!10^{-4}$
& 0.398
& 8 h
& $0.413 \pm 0.005$ \\

&
& $10^{-3}$
& 0.398
& 8 h
& $0.411 \pm 0.004$ \\

&
& $5\!\times\!10^{-3}$
& 0.398
& 8 h
& $0.418 \pm 0.004$ \\

&
& $10^{-2}$
& 0.398
& 8 h
& $0.422 \pm 0.008$ \\

\addlinespace[4pt]

Qwen2.5-7B / PRM800K$\rightarrow$MATH500
& SVD-GRPO, $r=4$
& $10^{-6}$
& 0.405
& 8 h
& $0.000 \pm 0.000$ \\

&
& $5\!\times\!10^{-6}$
& 0.405
& 8 h
& $0.000 \pm 0.000$ \\

&
& $10^{-5}$
& 0.405
& 8 h
& $0.000 \pm 0.000$ \\

&
& $5\!\times\!10^{-5}$
& 0.405
& 8 h
& $0.000 \pm 0.000$ \\

&
& $10^{-4}$
& 0.405
& 8 h
& $0.000 \pm 0.000$ \\

&
& $5\!\times\!10^{-4}$
& 0.405
& 8 h
& $0.000 \pm 0.000$ \\

&
& $10^{-3}$
& 0.405
& 8 h
& $0.300 \pm 0.014$ \\

&
& $5\!\times\!10^{-3}$
& 0.405
& 8 h
& $0.278 \pm 0.018$ \\

&
& $10^{-2}$
& 0.405
& 8 h
& $0.322 \pm 0.012$ \\

\addlinespace[4pt]

Qwen2.5-7B / PRM800K$\rightarrow$MATH500
& SVD-GRPO, $r=8$
& $10^{-6}$
& 0.401
& 8 h
& $0.444 \pm 0.002$ \\

&
& $5\!\times\!10^{-6}$
& 0.401
& 8 h
& $0.406 \pm 0.004$ \\

&
& $10^{-5}$
& 0.401
& 8 h
& $0.235 \pm 0.018$ \\

&
& $5\!\times\!10^{-5}$
& 0.401
& 8 h
& $0.432 \pm 0.005$ \\

&
& $10^{-4}$
& 0.401
& 8 h
& $0.346 \pm 0.009$ \\

&
& $5\!\times\!10^{-4}$
& 0.401
& 8 h
& $0.427 \pm 0.002$ \\

&
& $10^{-3}$
& 0.401
& 8 h
& $0.452 \pm 0.002$ \\

&
& $5\!\times\!10^{-3}$
& 0.401
& 8 h
& $0.402 \pm 0.006$ \\

&
& $10^{-2}$
& 0.401
& 8 h
& $0.488 \pm 0.001$ \\

\addlinespace[4pt]

Qwen2.5-7B / PRM800K$\rightarrow$MATH500
& SVD-GRPO, $r=16$
& $10^{-6}$
& 0.422
& 8 h
& $0.472 \pm 0.002$ \\

&
& $5\!\times\!10^{-6}$
& 0.422
& 8 h
& $0.472 \pm 0.004$ \\

&
& $10^{-5}$
& 0.422
& 8 h
& $0.484 \pm 0.001$ \\

&
& $5\!\times\!10^{-5}$
& 0.422
& 8 h
& $0.492 \pm 0.003$ \\

&
& $10^{-4}$
& 0.422
& 8 h
& $0.480 \pm 0.002$ \\

&
& $5\!\times\!10^{-4}$
& 0.422
& 8 h
& $0.480 \pm 0.014$ \\

&
& $10^{-3}$
& 0.422
& 8 h
& $0.492 \pm 0.001$ \\

&
& $5\!\times\!10^{-3}$
& 0.422
& 8 h
& $0.484 \pm 0.006$ \\

&
& $10^{-2}$
& 0.422
& 8 h
& $0.510 \pm 0.002$ \\

\bottomrule
\end{tabular}%
}
\caption{SVD-GRPO learning-rate sweep results for Qwen2.5-7B on
PRM800K$\rightarrow$MATH500. Each LoRA rank is evaluated using all nine
learning rates. For each run, we report the test-set performance of the
checkpoint from the final validation evaluation within the stated
wall-clock budget.}
\label{tab:grpo-run-level-prm}
\end{table*}

\end{document}